\documentclass{article}

\PassOptionsToPackage{numbers,sort&compress}{natbib}

\usepackage[final]{neurips_2022}

\usepackage{amsmath,amsfonts,bm,bbm}

\def\eqref#1{equation~\ref{#1}}

\def\1{\bm{1}}

\DeclareMathAlphabet{\mathsfit}{\encodingdefault}{\sfdefault}{m}{sl}
\SetMathAlphabet{\mathsfit}{bold}{\encodingdefault}{\sfdefault}{bx}{n}

\usepackage[utf8]{inputenc} %
\usepackage[T1]{fontenc}    %
\usepackage[dvipsnames]{xcolor}         %
\definecolor{mydarkblue}{rgb}{0,0.08,0.45}
\usepackage[colorlinks,citecolor=mydarkblue,urlcolor=mydarkblue,linkcolor=mydarkblue]{hyperref}
\usepackage{url}            %
\usepackage{booktabs}       %
\usepackage{amsfonts}       %
\usepackage{nicefrac}       %
\usepackage{microtype}      %
\usepackage{paralist}
\usepackage{longtable}
\usepackage{layouts}

\usepackage{graphicx}
\usepackage[capitalize]{cleveref}
\usepackage[small]{caption}
\usepackage{subcaption}
\usepackage{algorithm}
\usepackage{algorithmic}

\usepackage{wrapfig}
\usepackage{tcolorbox}  %
\usepackage{placeins} %

\newcommand{\takeaway}[1]{%
\begin{tcolorbox}[colback=blue!6!white,leftrule=2.5mm,size=title]
\textbf{Takeaway}: #1
\end{tcolorbox}
\vspace{-0.1cm}%
}

\def\ifcomments{\iftrue} %

\usepackage{todonotes} %

\definecolor{factorred}{rgb}{0.8352941176470589, 0.3686274509803922, 0.0}
\definecolor{factorgreen}{rgb}{0.00784313725490196, 0.6196078431372549, 0.45098039215686275}
\definecolor{factorblue}{rgb}{0.00392156862745098, 0.45098039215686275, 0.6980392156862745}
\definecolor{factororange}{rgb}{0.8705882352941177, 0.5607843137254902, 0.0196078431372549}

\setcitestyle{square}

\title{Assaying Out-Of-Distribution\\Generalization in Transfer Learning}

\usepackage{makecell}

\author{%
\hspace{-1.15cm}\begin{tabular}{ccc}
    Florian Wenzel$^{*\, 1}$
    &
    Andrea Dittadi$^{\dagger\, 2}$
    &
    Peter Gehler$^1$
    \\[.25em]
    Carl-Johann Simon-Gabriel$^1$
    &
    Max Horn$^1$
    &
    Dominik Zietlow$^1$
   \\[.25em]
    David Kernert$^1$
    &
    Chris Russell$^1$ %
    &
    Thomas Brox$^1$ %
    \\[.25em]
    Bernt Schiele$^1$ %
    &
    Bernhard Sch\"olkopf$^1$ %
    &
    Francesco Locatello$^1$
\end{tabular}\\
\\[1em]
$^1$ AWS T\"ubingen
$^2$  Technical University of Denmark
}

\begin{document}

\maketitle
\renewcommand{\thefootnote}{$*$}
\footnotetext{Correspondence to: \texttt{flwenzel@amazon.de}.}
\renewcommand{\thefootnote}{$\dagger$}
\footnotetext{Part of this work was done during an internship at AWS T\"ubingen.}
\renewcommand{\thefootnote}{1}
\footnotetext{The code for the evaluation study is at \href{https://github.com/amazon-research/assaying-ood}{github.com/amazon-research/assaying-ood}. Author contributions are \hyperref[sec:contributions]{listed at the end of paper}.}
\renewcommand{\thefootnote}{\arabic{footnote}}

\begin{abstract}
  Since out-of-distribution generalization is a generally ill-posed problem, various proxy targets (e.g., calibration, adversarial robustness, algorithmic corruptions, invariance across shifts) were studied across different research programs resulting in different recommendations. While sharing the same aspirational goal, these approaches have never been tested under the same experimental conditions on real data. In this paper, we take a unified view of previous work, highlighting message discrepancies that we address empirically, and providing recommendations on how to measure the robustness of a model and how to improve it. To this end, we collect $172$ publicly available dataset pairs for training and out-of-distribution evaluation of accuracy, calibration error, adversarial attacks, environment invariance, and synthetic corruptions. We fine-tune over $31\mathrm{k}$ networks, from nine different architectures
  in the many- and few-shot setting. Our findings confirm that in- and out-of-distribution accuracies tend to increase jointly, but show that their relation is largely dataset-dependent, and in general more nuanced and more complex than posited by previous, smaller scale studies$^1$.
\end{abstract}

\section{Introduction}
\label{sec:introduction}
With deep learning enabling a variety of downstream applications~\cite{he2016deep,krizhevsky2012imagenet,kolesnikov2020big,dosovitskiy2021an}, failures of robustness leading to %
systematic~\cite{barocas2017fairness,eyuboglu2021domino,hardt2016equality} and catastrophic deployment errors~\cite{ranjan2019attacking,deng2020analysis,beede2020human} have become increasingly relevant. From early work on studying distribution shifts~\cite[e.g.,][]{NIPS2006_b1b0432c, 10.1007/s10994-009-5152-4} and the classical ``cow on the beach'' example (e.g.,~in~\cite{beery2018recognition}), several works have highlighted sometimes spectacular failures of machine learning when the test distribution differs from  training~\cite{1312.6199,barbu19,shetty2019not,imagenet-c,karahan2016image,michaelis2019benchmarking,roy2018effects,beede2020human,schott2021visual}.
This has motivated the study of different types of distribution shifts, ultimately branching the field into several sub-communities that, while sharing the same underlying objective, rely on different evaluation protocols and provide different recommendations to practitioners. %

\textbf{(1)}~The studies \cite{imagenet-c,michaelis2019benchmarking,karahan2016image,roy2018effects,azulay2019deep,barbu19,engstrom2017exploring} focused on algorithmically corrupting upstream pre-training datasets~\cite{ILSVRC15} to test generalization. %
Perhaps unsurprisingly, the choice of augmentations can significantly alter this notion of robustness~\cite{saikia2021improving,cubuk2018autoaugment,many-faces,hendrycks2019augmix,rusak2020simple}.  \textbf{(2)} As synthetic corruptions need not transfer to real world distribution shifts~\cite{many-faces}, new realistic datasets were collected to test upstream robustness~\cite{barbu19,recht2019imagenet,shankar2021image,many-faces,wang2019learning,hendrycks2021natural}. Here,
scale has been identified as a reliable ingredient~\cite{robustness_convnet,accuracy-line,xie2020self,recht2019imagenet,scaling-transformer}, despite other works~\cite{yue2020interventional} arguing that extensive upstream pre-training can harm downstream robustness. \textbf{(3)} Exhaustive comparisons attempted to disentangle intrinsic architectural robustness from specific training schedules~\cite{bhojanapalli2021understanding,naseer2021intriguing,bai2021transformers,mao2021towards,paul2021vision,chen2021vision}, addressing underspecification~\cite{d2020underspecification} with inductive biases. %
Orthogonally, several (less scalable) works advocated for leveraging the compositional (perhaps causal~\cite{scholkopf2021toward}) structure in the underlying data-generative process to introduce suitable inductive biases~\cite{ParKilRojSch18,goyal2020recurrent,locatello2020object,locatello2020weakly,dittadi2021generalization,rahaman2021dynamic,alias2021neural}. %
\textbf{(4)} Simultaneously, Bayesian approaches for uncertainty predictions have been proposed to improve  model calibration~\cite{nixon2019measuring, ABDAR2021243, conf/nips/MaddoxIGVW19, DBLP:journals/corr/abs-2002-06715,DBLP:journals/corr/abs-2205-00403, DBLP:conf/icml/WenzelRVSTMSSJN20, conf/iclr/ZhangLZCW20, fortuin2021} and  robustness on new distributions~\cite{ovadia2019, hyperensembles, multidomain-calibration}. Recent work, however, found that larger models were natively better calibrated~\cite{calibration-modern}.
\textbf{(5)} The adversarial training community developed an entire literature on different worst case local perturbations of  training data~\cite{1312.6199, madry-pgd-2018}, with $5000$+ papers written to date \citep{all-adversarial-papers} and a never ending cycle of new defenses and attacks~\cite{obfuscated-gradients,papernot2016technical,rauber2017foolbox,rauber2020foolboxnative,robust-bench}.
\textbf{(6)} Other niche approaches investigated carefully designed test sets~\cite{sagawa2019distributionally,liang2022metashift,lost-domain} and training protocols that promote invariance across several distributions~\cite{arjovsky2019invariant,sagawa2019distributionally,creager2021environment,parascandolo2021learning,krueger2021out}. Despite this progress, \emph{empirical risk minimization (ERM)} remains a strong contender~\cite{lost-domain}.
Overall, the significant community effort towards more robust machine learning models have resulted in  diverse proxy evaluation targets yielding different practical recommendations.

At the same time, the workflow of successful applications developed in the opposite direction ~\cite{kolesnikov2020big,dosovitskiy2021an,radford2021learning,ramesh2021zero,gato}. Instead of collecting large application-specific datasets, one trains generalist backbones on the greatest possible amount of data and then transfers the model using available domain-specific examples. Besides the test data likely being ``on manifold'', one is almost certainly guaranteed that there will be some sort of distribution shift at test time as the size of the fine-tuning dataset decreases.%

Focusing on classification of visual data, we evaluate the different key metrics from these communities in a unified manner and under the same experimental conditions to investigate the gaps in common practices. We restrict ourselves to the realistic situation where we have an ImageNet pre-trained model available and a new target distribution as downstream task. After the model has been fine-tuned, the test data may be OOD.
From  $36$ existing datasets, we extract $172$ \emph{in-distribution (ID)} and \emph{out-of-distribution (OOD)} dataset pairs, fine-tuning and evaluating over $31\mathrm{k}$ models to gain a broader insight in the sometimes contradicting statements on OOD robustness in previous research. %
We organize our study around two key questions: (1)~What are good proxy measures of OOD robustness when having access to a single dataset?
(2)~How do architecture choices and fine-tuning strategies affect robustness?
We plan to publish the code with the camera-ready version of the paper.

Our key contributions are \textbf{(1)}~We conduct a large systematic study of OOD robustness, evaluating the effect of architecture type, augmentation, fine-tuning strategies and few-shot learning. %
We investigate the interplay of robustness to corruptions, adversarial robustness, robustness to natural distribution shifts, calibration and other robustness metrics in a unified setting and under the same experimental conditions. \textbf{(2)}~We find that out-of-distribution generalization has many facets. Insights of previous papers---sometimes presented as general conclusions--- hold only on a subset of the tasks/datasets included in our study and hence actually only reflect a special case. \textbf{(3)}~In general, in-distribution classification error (accuracy) is the best predictor of OOD accuracy, but other secondary metrics can provide additional insights. \textbf{(4)}~With these results, we revisit previous studies and recommendations, reinterpreting their conclusions, resolving some contradictions, and suggesting critical areas for further research.

\section{Experimental setup}\label{sec:setup}
We follow the modern workflow of applications of computer vision to (long-tail) downstream tasks from existing pre-trained backbones. The model is transferred using a set of (potentially few) examples from a new distribution. At test time, we assume that the classes remain the same (closed-world setting), but that the distribution may otherwise change. We specifically focus on the effect of distribution shifts \emph{after} a model has been transferred to a new distribution (i.e., the \emph{downstream} implications) and  discuss the empirical differences and similarities compared to results concerning \emph{upstream} OOD robustness that were discussed in previous studies~\cite{barbu19,recht2019imagenet,shankar2021image,wang2019learning,hendrycks2021natural,robustness_convnet,many-faces}.

\emph{Experimental protocol and datasets:} We evaluate nine  state-of-the-art deep learning models with publicly available pre-trained weights for ImageNet1k / ILSVRC2012~\cite{ILSVRC15}. We consider $36$ datasets grouped into ten different \textit{tasks} sharing the same labels. Datasets of the same task represent a set of natural distribution shifts. For each task, we take a single training dataset to fine-tune the model and report evaluation metrics on both its ID test set and all the other OOD test sets.
We extract $172$ (ID, OOD) dataset pairs from the different domains of the ten tasks: DomainNet~\cite{peng2019moment}, PACS~\cite{li2017deeper}, SVIRO~\cite{DiasDaCruz2020SVIRO}, Terra Incognita~\cite{beery2018recognition} as well as the Caltech101~\cite{FeiFei2004LearningGV}, VLCS~\cite{conf/cvpr/TorralbaE11}, Sun09~\cite{choi_cvpr10}, VOC2007~\cite{pascal-voc-2007} and the Wilds datasets~\cite{wilds2021} (from which we extract two tasks).
In our experimental protocol we do not make any assumptions on the particular shift type and the considered tasks reflect multiple shift types (e.g., presumably a strong covariate shift in  DomainNet and a partial label shift in Camelyon17 of the Wilds benchmark). 
See \cref{app:datasets} for a detailed overview.
Models are fine-tuned on a single GPU using Adam~\cite{kingma2014adam} with a batch-size of $64$ and a constant learning rate.

\emph{Evaluation on ID, OOD and corrupted data:}
Some tasks, such as DomainNet, PACS and SVIRO come with different datasets/domains. For those, we report for each dataset the ID (test) performance and the OOD (test) performances on the other datasets in the task. For the datasets from the WILDS benchmark, we use the provided ID test and OOD test splits.
If a task consists of multiple OOD data we compute the metrics additionally on held-out OOD data. To do so, for each (ID, OOD) dataset pair, we average the performance on the remaining OOD datasets. This approach is sometimes called multi-domain evaluation~\cite[e.g.,][]{multidomain-calibration}.
Alongside  the provided OOD datasets, we evaluate the models on the corrupted ID test set. We apply $17$ types of corruptions from \cite{imagenet-c} each with $5$ severity levels. The corrupted version of the datasets can be viewed as a synthetic distribution shift and we investigate how informative they are of natural distribution shifts.

\emph{Models:} To ensure that our results are relevant for researchers and practitioners alike, we consider both widely deployed and recent top-performing methods: Resnet50d~\cite{he2019bag}, DenseNet~\cite{huang2017densely}, EfficientNetV2~\cite{tan2021efficientnetv2}, gMLP~\cite{NEURIPS2021_4cc05b35}, MLP-Mixer~\cite{NEURIPS2021_cba0a4ee}, ResMLP~\cite{touvron2021resmlp}, Vision Transformers\footnote{Trained on ImageNet21k and fine tuned on ImageNet1k.}~\cite{dosovitskiy2021an}, Deit~\cite{pmlr-v139-touvron21a}, Swin Transformer~\cite{liu2021swin}.
We list the exact model names in \cref{tab:model_names}. Our choice of models covers convolutional networks, transformer variants and mixers.
Weights for the pre-trained models were taken from the PyTorch Image Models repository~\cite{rw2019timm}.

\emph{Model hyperparameters and augmentation strategies:}
For each model we consider the learning rate and the number of fine-tuning epochs.
We first ran a large sweep over these two hyperparameters on a subset of the experiments and used it to pre-select a set of four parameter combinations that included the best performing models for each architecture.
Additionally, we study three different augmentation strategies: standard ImageNet augmentation (i.e., no additional augmentation), \emph{RandAugment}~\cite{NEURIPS2020_d85b63ef} and \emph{AugMix}~\cite{hendrycks2019augmix}. More details can be found in \cref{app:model_selection}.

\emph{Fine-tuning strategy and few-shot training:}
We investigate fine-tuning the full architecture and fine-tuning only the head.
Additionally, we consider three training paradigms: training on the full downstream dataset, and two few-shot settings: ``few-shot-$100$'' (a subset with $100$ examples per class, if available) and ``few-shot-$10$'' (with $10$ examples per class). In the few-shot settings, the images are randomly selected, and classes that have fewer images as the cap of $10$ or $100$, respectively, are not over-sampled.

\emph{Metrics:} We pick some of the most popular metrics that are used to measure progress towards robust machine learning.
We report six different metrics: \emph{classification error}, \emph{negative log-likelihood (NLL)}, \emph{demographic disparity} \cite{dwork2012fairness,locatello2019fairness} on inferred groups~\cite{creager2021environment} as a measure of invariance\footnote{As there is no ``measure of invariance'' for a single dataset, we rely on ~\cite{creager2021environment} that finds a partition of the data maximising the IRM~\cite{arjovsky2019invariant} penalty.}, the \emph{expected calibration error (ECE)}  \cite{guo2017calibration}, and adversarial classification error for two different $\ell_2$-attack sizes. The metrics are, where applicable, evaluated on ID, OOD, and corrupted test sets, (except adversarial error, which we did not evaluate on the corrupted test sets). See \cref{app:metrics} for more details.

\section{Additional related work}
As much of the related work was already mentioned in the introduction we highlight two main areas of closely related works: one regarding benchmarks for generalization to new distributions and one on the interplay between different evaluation metrics.

\textbf{Benchmarking robustness to OOD.} Closely to our setting,
\cite{vtab} benchmarked models in a few-shot learning setting but did not analyze the robustness of the fine-tuned models. In follow up work, \cite{robustness_convnet} related the results of \cite{vtab} to \textit{upstream} robustness but did not consider downstream distribution shifts.
\cite{lost-domain} analyzed a variety of domain generalization algorithm and found that none of them could beat a strong ERM baseline. While several of our datasets overlap with theirs, we consider the transfer learning setting as opposed to domain generalization.
Their insights may be in part explained by the fact that the regularization is either orthogonal to OOD accuracy or simply harms accuracy overall as in~\cite{zietlow2022leveling}.
\cite{fine-grained,schott2021visual} proposed a model for analyzing different fine-grained distribution shifts. Their work is limited to few datasets and model types and only cover accuracy evaluations.
\cite{broad_study_domain} studied the effect of the pre-training strategy on domain generalization and \cite{plex2022} studied extensions to large pre-trained models for improved reliability, whereas our work analyzes fine-tuning protocols and the robustness on downstream tasks.
\cite{many-faces} found that larger models and better augmentation techniques improve robustness but did not consider different model types, augmentation techniques or evaluation metrics.
Our work studies robustness in a larger scope than previous work, which focused on certain dimensions of our empirical investigations. None of the previous work studied the interplay of different robustness metrics.

\textbf{Studying the interplay of robustness metrics.}
There has been only limited work on analyzing informativeness of robustness metrics on OOD generalization.
\cite{taori-shifts} analyzed distribution shifts of ImageNet and found that corruption metrics do not imply robustness to natural shift.
Recently, \cite{accuracy-line}, based on previous studies by \cite{taori-shifts,shankar19,barbu19, pmlr-v119-shankar20c}, observed a clear linear relationship between ID accuracy and OOD accuracy
and hypothesized that this could be a general pattern in contradiction to~\cite{d2020underspecification}. \cite{agreement_on_the_line} extended this line of work to agreement between networks and \cite{andreassen21} found that large pre-trained models are above the linear trend in early stages of fine-tuning. However, our extended set of experiments show that a clear linear trend is only visible on some (ID, OOD) dataset pairings.
\cite{NEURIPS2021_ecf9902e} empirically investigated different generalization measures and found that measures relating to the Fisher information perform best.

\section{A broad look at out-of-distribution generalization}
\label{sec:a-broad-look-at-out_of_distribution-generalization}

In the following we explore the facets of out-of-distribution generalization, highlighting discrepancies to prior work and discuss their implications.
\subsection{The main latent factors that explain the empirical results}
\label{sec:factor-analysis}

\begin{wrapfigure}[16]{r}{0.5\textwidth}   %
    \centering
    {
    \captionsetup{aboveskip=4pt}
    \vspace{-11pt}
    \includegraphics[width=0.5\textwidth]{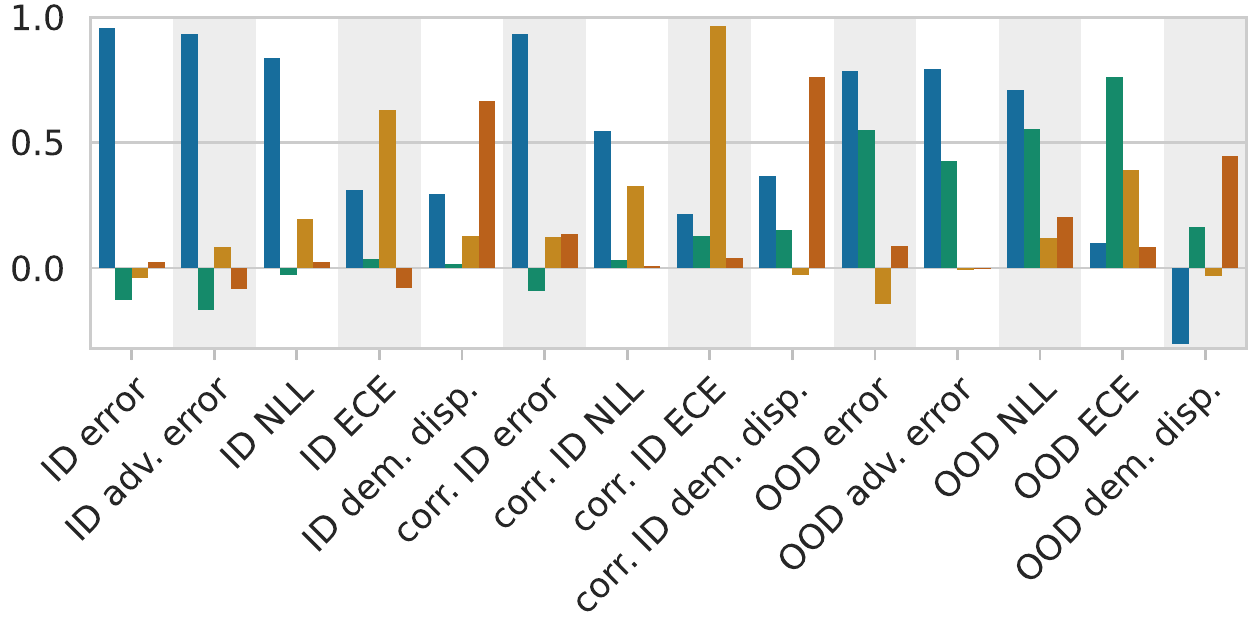}
    \caption{Factor loadings (contributions) of different metrics based on a factor analysis with 4 orthogonal factors (color-coded), highlighting similarities between the metrics. The factor \textcolor{factorblue}{\textbf{Blue}}: captures classification error, adversarial error, log-likelihood, and their corrupted variants. \textcolor{factorgreen}{\textbf{Green}}: only in OOD metrics. \textcolor{factororange}{\textbf{Yellow}}: expected calibration error. \textcolor{factorred}{\textbf{Red}}: demographic disparity.
    }\label{fig:factor-analysis}
    }
\end{wrapfigure}

To get a first overview of the relations between the different metrics and their generalization properties, we perform a factor analysis to discover the main orthogonal latent factors that explain the variance in the metrics evaluated on each ID dataset, its corrupted variant, and the metrics averaged over all compatible OOD datasets for each fine-tuned model. For details, see~\cref{app:factor-analysis}.

Based on the scree plot in \cref{app:factor-analysis}, we retain four factors.
Their contributions (loadings) to each metric are shown in \cref{fig:factor-analysis}.
Interestingly, each factor has a clear interpretation.
Factor 1 (blue) is very well aligned with ID classification error, log-likelihood and adversarial attacks.
Factor 2 (green) captures OOD-specific variance, since it is particularly pronounced in almost every out-of-distribution metric, and only there.
Factor 3 (orange) relates mainly to the expected calibration error and factor 4 (red) to demographic disparity.
The dominant presence of factor 1 (blue) in all classification error and log-likelihood metrics ID and OOD suggests that ID classification error can be a reasonably good predictor of OOD classification error, which we further discuss in \cref{sec:metrics_predict_ood}.
However, the presence of an OOD-specific factor also suggests that ID versus OOD accuracy (classification error) cannot always lie ``on a line'' \cite{accuracy-line}---we investigate this further in \cref{sec:acc-not-on-line}.
Another noteworthy point is that the corrupted metrics and adversarial classification errors have almost no OOD component and are generally very close to the corresponding ID metric.
Similarly, the loadings of the corrupted metrics are much closer to those of the ID metrics than to OOD metrics.
This suggests that the performance on artificially corrupted data may not predict the OOD performance significantly better than the bare ID metrics.
We further discuss this in \cref{sec:other_metrics_for_ood_predictions}.
Finally, the fact that demographic disparity and expected calibration error are each mainly captured by their own, specific factor suggests that, maybe surprisingly, those metrics are largely independent of the networks' classification error.
Further details are discussed in \cref{sec:other_metrics_for_ood_predictions}.

\takeaway{One latent factor suffices to capture accuracy and log-likelihood on ID, corrupted, and adversarial datasets. OOD behavior, calibration, and environment invariance are each captured by a separate factor. A separate factor for OOD metrics suggests that artificial and adversarial corruptions do not fully mimic real distribution shifts.}

\subsection{The many facets of out-of-distribution generalization}\label{sec:acc-not-on-line}

\begin{figure}[tb]
    \centering
    \includegraphics{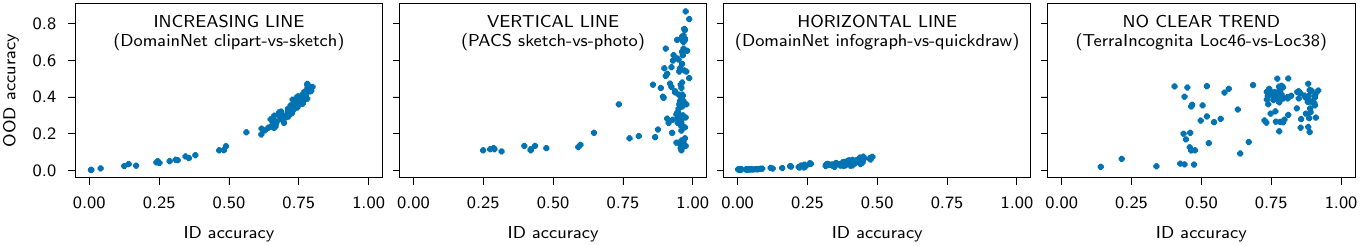}
    \caption{Typical scatter plot patterns observed in our data (see \cref{app:scatter_plots} for all plots). \emph{Increasing line}: ID and OOD accuracy show a clear functional dependency. In contrast to previous claims this is not the typical setting (only observed on a subset of datasets). \emph{Vertical line}: the same ID peformance leads to different OOD behavior (underspecification setting). \emph{Horizontal line close to zero accuracy}: no transfer of information from the ID to the OOD dataset. \emph{No clear trend}: random associations between ID and OOD accuracy (i.e., zero correlation).}
    \label{fig:scatter-plots-summary}
\end{figure}

Prior publications \cite{accuracy-line,taori-shifts,robustness_convnet} observed that OOD accuracy strongly linearly correlates with ID accuracy, or, in other words, that ID vs OOD accuracy nearly lie ``on a line''.
In contrast, we find that this is not a general trend when tested on more tasks.
\cref{fig:scatter-plots-summary} shows the four typical settings we observe. For some (ID, OOD) dataset pairs we observe a clear functional dependency as claimed by \cite{accuracy-line,taori-shifts} (increasing line).
For other dataset pairs we observe a clear underspecification problem~\cite{d2020underspecification}: very similar ID performances (in most cases close to $1$) lead to different OOD performances (vertical line). In this setting, ID accuracy is not a sufficient model selection criterion for obtaining robust models\footnote{One may be tempted to think of this as a saturation phenomenon, where the ID data is too easy to learn to distinguish the good networks from the bad ones.
In that case, however, the generalization properties should significantly depend on the architecture (and pre-training performance), so that models with best OOD performance should be the same on every dataset.
What we observe instead is that the order seems to be largely random in different dataset pairs.}.
In some settings, the models do not transfer information from the ID to the OOD data at all and, despite having different ID performance, all models have very poor OOD performance. Finally, we observe a fourth setting, where OOD accuracy is hardly correlated to ID accuracy.
Interestingly, we never see a decreasing trend, i.e., improved ID performance never systematically results into lower OOD performance.
Hence, despite the many shapes of ID and OOD dependency, it is still a good strategy to maximize the ID accuracy in order to maximize the OOD accuracy.

\textbf{Results can significantly change for different shift types.}
We highlighted how much ID to OOD generalization can change on different tasks/datasets.
This is further confirmed by the task-specific correlation matrices in \cref{app:correlation_matrix_per_task,app:correlation_matrix_by_shift}, which, more generally, show that there can be significant differences in various metrics between different tasks or shift types.
For example, comparing the \emph{terra-incognita} and \emph{wilds-fmow} specific correlation matrices, we see that for \emph{terra-incognita} calibration and demographic disparity have a strong \emph{positive} correlation with OOD accuracy, whereas for \emph{wilds-fmow} the correlation is strongly \emph{negative}. Similarly, multi-domain calibration as proposed by \cite{multidomain-calibration} only improves OOD robustness on some tasks, but has a negative effect on others (details in \cref{sec:metrics_predict_ood}).
\Cref{app:correlation_matrix_by_shift} shows that focusing on different shift types can also lead to contradicting findings.
For instance, for models that were trained on \emph{artificial} data (such as sketches, clipart, simulated environments) and evaluated on \emph{real} OOD data, corruption metrics are more predictive of OOD robustness than for models that were trained on \emph{real} data and tested on \emph{artificial} OOD data. Additionally, we discuss in \cref{app:task_difficulty} the dependence of the results on the task difficulty.

\takeaway{ID and OOD accuracy only show a linear trend on specific tasks. We observe three additional settings: underspecification (vertical line), no generalization (horizontal line), and random generalization (large point cloud). We did not observe any trade-off between accuracy and robustness, where more accurate models would overfit to ``spurious features'' that do not generalize. Robustness methods have to be tested in many different settings. Currently, there seems to be no single method that is superior in all OOD settings.}

\subsection{What are good proxies to measuring robustness to distribution shifts?}
\label{sec:metrics_predict_ood}

\begin{figure}
    \centering
    {
    \captionsetup{aboveskip=0pt}
        \includegraphics[width=\linewidth]{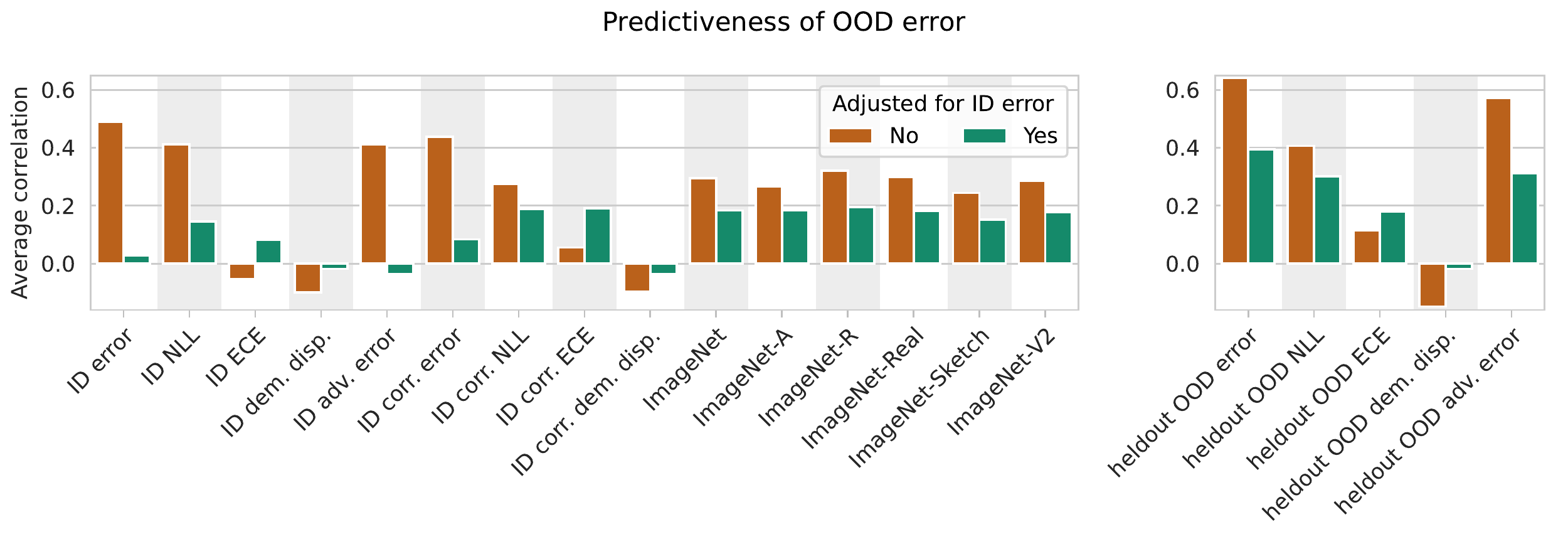}
        \caption{%
        \textbf{\textsc{left:}}
        \emph{What is a good proxy for classification error under natural distribution shifts?} We measure how well several popular robustness metrics on in-distribution (ID) data predict classification error on out-of-distribution (OOD) datasets.
       \textcolor{factorred}{\textbf{Red bars}}: The predictiveness score is computed based on Spearman's rank correlation coefficient between the robustness metric and OOD classification error. We find that, among all considered metrics, ID classification error is the strongest predictor of OOD robustness.
        \emph{What is the additional information content of the robustness metrics adjusted for ID classification error?} \textcolor{factorgreen}{\textbf{Green bars}}: We compute the adjusted predictiveness scores as outlined in \cref{sec:metrics_predict_ood}. When adjusted for ID classification error, all secondary metrics only provide limited information.
        \textbf{\textsc{right:}} \emph{How predictive are the metrics on additional held-out OOD data?} Evaluating accuracy on held-out OOD data (multi-domain evaluation) is the strongest predictor of OOD accuracy and provides significant additional information to ID accuracy (see adjusted scores).
        }
        \label{fig:barplot_corr_unified}
    }
\end{figure}

Can we predict the robustness of a model by using a proxy measure? In other words, how predictive is a certain \emph{metric $A$} (e.g., ID expected calibration error) of another \emph{metric $B$} (e.g., OOD classification error)?
To this end we compute the \emph{averaged correlation matrix} which reports the rank correlation of all metrics, averaged over all tasks. The matrix and details on the method are deferred to \cref{app:corr_matrices}.
We already saw --and the matrix confirms-- that accuracy is a strong predictor of OOD accuracy. This raises the question if other metrics add any additional information on OOD accuracy which is not already provided by ID accuracy.
To test this, we compute adjusted predictiveness scores as follows. For each dataset pair, we fit a linear regression to predict OOD accuracy from ID accuracy. We then report the averaged rank correlation coefficient between the obtained residuals and each metric.
This measure is similar to the \emph{effective robustness} proposed in~\cite{accuracy-line}.
Results are shown in \cref{fig:barplot_corr_unified} and discussed in the upcoming subsections.

\subsubsection{Overall classification error is the best general predictor of OOD robustness} %
\cref{fig:full_corr_matrix}, derived from the full averaged correlation matrix in \cref{app:corr_matrices}, shows that among all considered metrics, ID classification error is the strongest predictor of OOD classification error. This finding is in contrast to works that hypothesized that evaluating the classification error on corrupted data (e.g., ImageNet-C~\cite{imagenet-c}) or on adversarial perturbed data~\cite{adversarial-ood} provides additional information on how models perform under natural distribution shifts. Although these metrics show a high correlation with OOD classification error, we do not find that they add significant information when adjusting for ID classification error.
However, when having access to additional OOD datasets, the classification error on the held-out OOD datasets is even more powerful predictor of the robustness of the OOD dataset of interest, see \cref{fig:barplot_corr_unified} (right). We find that this is the most reliable model selection procedure of all considered metrics.

Our findings imply that if practitioners want to make the model more robust on OOD data, the main focus should be to improve the ID classification error. This is in accordance with previous work that found that models with high ID classification error tend to be more robust~\cite{accuracy-line, robustness_convnet}. We speculate that the risk of ``overfitting'' large pre-trained models to the downstream test set is minimal, and it seems to be not a good strategy to, e.g., reduce the capacity of the model in the hope of better OOD generalization~\cite{yue2020interventional}. %
Finally, we recommend that architectural innovations and training techniques can leverage scale but that robustness comparisons should always be adjusted for classification error.

\takeaway{Accuracy is the strongest ID predictor of OOD robustness and models that generalize well in distribution tend to also be more robust.
Evaluating accuracy on additional held-out OOD data is an even stronger predictor.
}

\subsubsection{What can we learn from other metrics beyond accuracy?}
\label{sec:other_metrics_for_ood_predictions}

The first interesting result is that calibration on ID data is \emph{not} predictive of OOD robustness or OOD log-likelihood (see \cref{fig:full_corr_matrix} in the appendix). Restricted to the ID regime, however, we observe a correlation between ID calibration and ID classification error, which is in accordance with \cite{calibration-modern}. This difference is explained by the fact that ID calibration is not predictive of OOD calibration without an OOD held-out set (see \cref{sec:correlations_metrics_transfer}). In contrast to the observations in~\cite{multidomain-calibration}, we see that a model that is well-calibrated on multiple domains (held-out OOD data) may not always have lower OOD classification error (e.g., negative correlation for domain-net but positive on office-home, see \cref{app:correlation_matrix_per_task}).
Interestingly, invariance measured with environment inference~\cite{creager2021environment} and demographic disparity~\cite{locatello2019fairness} is not predictive of OOD robustness but seems to be a good proxy for calibration of OOD data (see \cref{fig:full_corr_matrix}) which is consistent with our observations on multi-domain calibration\footnote{Given the decomposability of the log-score, the objectives of both approaches are related.} and may be useful for OOD detection.

ID log-likelihood and adversarial accuracy are both weak predictors of OOD robustness compared to ID accuracy, and when adjusted for ID accuracy they only add marginal to no information.
Since the correlation between ID adversarial classification error and OOD classification error is fully explained by ID accuracy (see \cref{fig:barplot_corr_unified}, left) suggests that adversarial distribution shifts do not characterize well natural distribution shifts.

\textbf{Synthetic corruptions}
We apply the synthetic corruptions proposed by \cite{imagenet-c} to all datasets. %
First, we find that classification error and log-likelihood evaluated on the corrupted data are strongly correlated to OOD classification error (see \cref{fig:barplot_corr_unified}, left). However, we find that the information provided by the corrupted metrics is significantly reduced when adjusted for ID accuracy.
With the partial exception of corrupted calibration being more informative of OOD calibration than ID calibration (see \cref{sec:correlations_metrics_transfer}). In summary, evaluation on corrupted data does not seem to bring the same benefits as using real held-out OOD data (see \cref{fig:barplot_corr_unified}, right).
Interestingly, we find that adversarial classification error is highly correlated to the classification error under synthetic corruptions (see \cref{fig:full_corr_matrix}). Therefore, if the practitioner cares about shifts defined by artificial corruptions, studying the adversarial robustness on ID data will be informative.

\textbf{Robustness to upstream dataset shifts}
In our study all models are pre-trained on ImageNet (upstream dataset) and then fine-tuned on downstream data. In this section, we explore if upstream robustness propagates downstream. %
First, we notice in \cref{fig:barplot_corr_unified} (left) that the original performance on ImageNet is linked to OOD classification error in accordance to previous studies~\cite{robustness_convnet}. When we adjust for ID classification error, the clean ImageNet performance is among the strongest predictors for OOD classification error. %
Second, we find that robustness on ImageNet shifts does not give much additional information to the downstream robustness compared to clean performance. The performance on ImageNet shifts is almost perfectly correlated with the ID performance (in this setting accuracy is perfectly ``on the line'', c.f.~\cref{sec:acc-not-on-line}), but this relationship does not translate to our diverse set of downstream shifts. %

\takeaway{Other metrics can add marginal additional information for OOD robustness. Calibration appears to be predictive of ID accuracy but does not transfer to new distributions and adversarial robustness appears not to reflect robustness to natural distribution shifts.
Corruptions are only marginally useful for measuring robustness to natural distribution shifts and should not be used as a substitute to real held-out OOD data.
ImageNet upstream performance provides information on downstream robustness. However, robustness to commonly used shifts of ImageNet does not imply downstream robustness more than the clean upstream accuracy.
}

\begin{wrapfigure}[16]{r}{0.5\textwidth}   %
    \centering {
    \captionsetup{aboveskip=4pt}
    \vspace{-11pt}
    \includegraphics[width=0.5\textwidth]{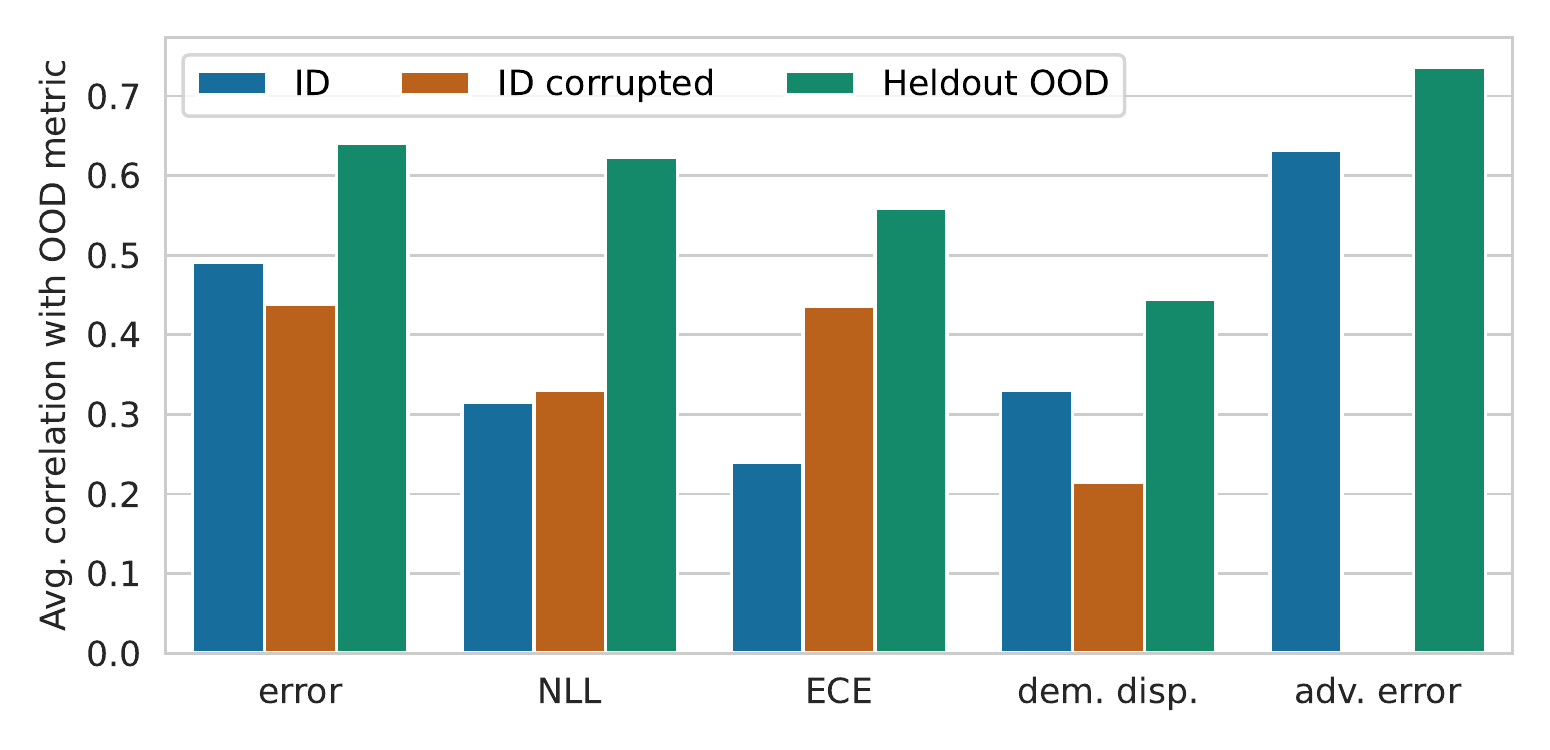}
    \caption{For each source metric on the x-axis we display the averaged correlation with the same metric evaluated on OOD data (target). The different colors indicate on which data domain the source metric was evaluated; either on ID, ID corrupted, or held-OOD data. Note that in our study we do not evaluate adversarial classification error on corrupted data.
    }
    \label{fig:correlations_metrics_transfer}
    }
\end{wrapfigure}

\subsection{On the transfer of metrics from ID to OOD data} \label{sec:correlations_metrics_transfer}
The main focus of \cref{sec:metrics_predict_ood} was to analyze how informative the different metrics are of OOD classification error. In a more general setting, we now explore how well a metric evaluated on ID data predicts their score on OOD data. \cref{fig:correlations_metrics_transfer} shows the averaged correlation coefficient of each metric---evaluated either on ID, corrupted or held-out data---with the same metric evaluated on OOD data.
First, we find that all ID metrics transfer moderately well to OOD data (blue bars). For adversarial attacks the transfer is highest. This suggests that the models respond similarly to adversarial attacks on ID data and on OOD data.
On the other hand, ID calibration transfers worst among all metrics, i.e., a model that is well calibrated on ID data, is not necessarily well calibrated on OOD data. This points to an important problem, since in many production systems models are only calibrated on ID data.
Second, we observe that the evaluation on corrupted data does not add significant information to the evaluation on ID data (blue vs.\ red bars) for most metrics. Interestingly, we observe one exception; for calibration the evaluation on corruptions is significantly more informative.
Third, when having access to additional held-out data, the evaluation on this data is the strongest predictor for the OOD behavior for all metrics (green bars).

\takeaway{Among all metrics adversarial robustness transfers best from ID to OOD data, which suggests that models respond similarly to adversarial attacks on ID and OOD data. Calibration transfers worst, which means that models that are well calibrated on ID data are not necessarily well calibrated on OOD data.}

\section{The effect of the training strategy on out-of-distribution robustness}
\label{sec:effect-of-augmentations-fine_tuning-stragety-and-few_shot-learning}

We now investigate the influence of the training strategy and model architecture on OOD robustness for practitioners.
Although we will observe clear trends, \textbf{they should be taken with care}, since each model was \emph{pre-trained} with its own training procedure (with different optimizers, learning rate schedules, augmentations, sometimes even datasets, etc.), which is likely to confound downstream results even after using a unified fine-tuning procedure. This is a general problem since different architectures usually require a specific pre-training procedure. Most practitioners usually undergo the same pipeline, starting from a network with publicly available pre-trained weights.

\subsection{The effect of augmentations, fine-tuning strategy and few-shot learning}

\begin{figure}
\begin{minipage}[t]{0.5\textwidth}
\vspace*{0pt}
    \includegraphics[width=0.95\linewidth]{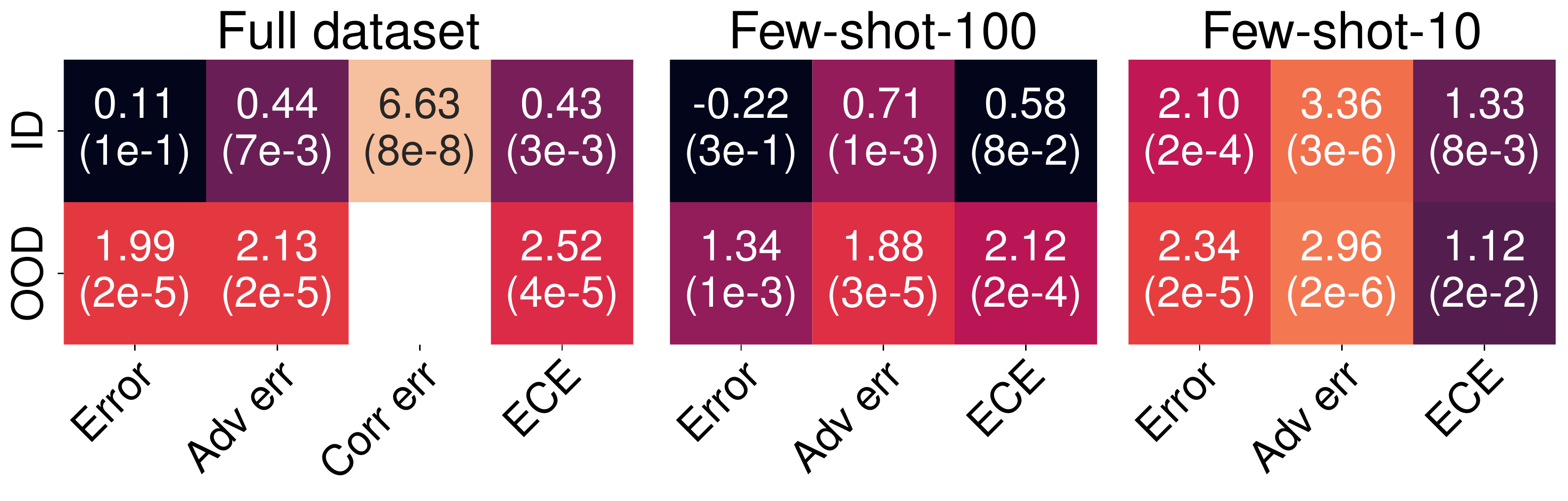}
    \caption{Performance gap (difference) between models trained with and without augmentations together with a p-value in parenthesis to assess its significance. Black fields indicate a p-value above the 0.05 significance threshold (i.e., non-significant); the other values are significant. Overall, augmentations help increasing the model's accuracy and its robustness to all kinds of distribution shifts (artificial and adversarial corruptions, OOD generalization), more so when data is scarce (few-shot settings).
    }
    \label{fig:effect_of_augmentations}
\end{minipage}
\hfill
\begin{minipage}[t]{0.47\textwidth}
\vspace*{0pt}
    \centering
    {\tiny%
    \begin{longtable}{lccc}
\toprule
        Model &                            ID Error &                           OOD Error &                          OOD-ID Gap \\
\midrule
\endfirsthead

\toprule
        Model &                            ID Error &                           OOD Error &                          OOD-ID Gap \\
\midrule
\endhead
\midrule
\multicolumn{4}{r}{{Continued on next page}} \\
\midrule
\endfoot

\bottomrule
\endlastfoot
         Deit & \textbf{0.101} \textcolor{gray}{$\pm$ 0.005} & \textbf{0.364} \textcolor{gray}{$\pm$ 0.008} & 0.263 \textcolor{gray}{$\pm$ 0.006} \\
         Swin & 0.111 \textcolor{gray}{$\pm$ 0.005} & 0.371 \textcolor{gray}{$\pm$ 0.008} & \textbf{0.260} \textcolor{gray}{$\pm$ 0.006} \\
        ViT-B & 0.124 \textcolor{gray}{$\pm$ 0.005} & 0.384 \textcolor{gray}{$\pm$ 0.008} & 0.259 \textcolor{gray}{$\pm$ 0.006} \\
     ResNet50 & 0.124 \textcolor{gray}{$\pm$ 0.005} & 0.406 \textcolor{gray}{$\pm$ 0.008} & 0.283 \textcolor{gray}{$\pm$ 0.006} \\
EfficientNet2 & 0.129 \textcolor{gray}{$\pm$ 0.005} & 0.407 \textcolor{gray}{$\pm$ 0.008} & 0.277 \textcolor{gray}{$\pm$ 0.006} \\
         GMLP & 0.140 \textcolor{gray}{$\pm$ 0.006} & 0.413 \textcolor{gray}{$\pm$ 0.008} & 0.273 \textcolor{gray}{$\pm$ 0.006} \\
       ResMLP & 0.134 \textcolor{gray}{$\pm$ 0.005} & 0.413 \textcolor{gray}{$\pm$ 0.008} & 0.279 \textcolor{gray}{$\pm$ 0.006} \\
        Mixer & 0.142 \textcolor{gray}{$\pm$ 0.006} & 0.425 \textcolor{gray}{$\pm$ 0.008} & 0.282 \textcolor{gray}{$\pm$ 0.006} \\
  DenseNet169 & 0.145 \textcolor{gray}{$\pm$ 0.005} & 0.443 \textcolor{gray}{$\pm$ 0.008} & 0.298 \textcolor{gray}{$\pm$ 0.006} \\
\end{longtable}

    \captionof{table}{Average classification error of model architectures with the standard error of this average in grey.
    To simulate a typical transfer learning workflow, we selected the best performing augmentations based on ID validation data for each fine tuning domain.}
    \label{tab:id_ood_gap}
    }
\end{minipage}
\end{figure}

To evaluate the effect of augmentations during fine-tuning, we average the performance of networks trained with RandAugment~\citep{NEURIPS2020_d85b63ef} and AugMix~\citep{hendrycks2019augmix} and compare it to fine-tuning without augmentations. \cref{fig:effect_of_augmentations} shows the performance gap between models trained with and without augmentations together with the p-value of a one-sided Wilcoxon signed-rank test that assesses whether the model trained without augmentations is better than the other one.
Overall, augmentations appear to increase accuracy across all corruption types (natural, corrupted and adversarial data), particularly on OOD data.
This suggests that augmentations not only improve accuracy in distribution, but also increase the model's robustness under certain shifts.
The effect is more pronounced when data is scarce (few-shot setting), although exceptions exist (accuracy in ``few-shot-$100$''). We discuss additional results in \cref{appendix:augmentation}.

Previous studies have shown that the fine-tuning strategy significantly affects the robustness \cite{fine-grained, DBLP:journals/corr/abs-2106-15831, kumar2022finetuning}. In our study we investigate two popular fine-tuning methods: (1) fine-tuning the full architecture and (2) fine-tuning the head only, while keeping the rest of the architecture frozen. We discuss the results in \cref{appendix:fine-tuning} and find that fine-tuning the full architecture is better for most of the considered tasks when having access to the full datasets. However, in the low data regime (few-shot-10 setting), fine-tuning the head only is beneficial on 40\% of the tasks.

\takeaway{Augmentations can improve accuracy and robustness to all kinds of distribution shifts (artificial and adversarial corruptions, OOD generalization), especially when data is scarce. While fine-tuning the full architecture is beneficial when having access to the full dataset, fine-tuning the head only can lead to higher robustness in the low data regime.

}

\subsection{The effect of the model architecture}

With many pre-trained backbones available in libraries like~\citep{rw2019timm} that often achieve very similar results on ImageNet, it is not obvious whether the architecture choice matters.
\cref{tab:id_ood_gap} shows the average ID and OOD classification errors of each model. %
Interestingly, we observe that while the Vision Transformer ViT-B was trained on more data it performs worse than Swin- and Deit-Transformers both on ID and OOD data (both approx.~$3\,\%$ higher error than Deit). This indicates that the extensions made to vision transformers improve generalization performance in the transfer learning  and fine-tuning scenario, while additionally requiring less data.
Further, we notice that the model with lowest average OOD classification error, does \emph{not} show the lowest performance gap, i.e., the performance on ID data and OOD data are not necessarily more closely aligned when performance on ID and OOD accuracy increases.

\takeaway{In the light of previous work that argued that domain generalization methods only have a marginal effect on OOD robustness~\cite{lost-domain}, we encourage more research on robust architectures, as our results indicate that the architecture can indeed make a difference.}

\section{Conclusions}
\label{sec:conclusion}
In this paper, we thoroughly investigated out-of-distribution generalization and the interplay of several secondary metrics in the transfer learning setting.
We focused on understanding sometimes contradicting empirical evidence from previous studies and on reconciling the results with anecdotal evidence from common practice in computer vision.
We fine-tuned and evaluated over $31\mathrm{k}$ models across several popular architectures on $172$ (ID, OOD)-dataset pairs and found the following.
\textbf{(1)}~The risk of overfitting on the transfer distribution appears small: models that perform better in distribution tend to perform better OOD. All other proxy metrics convey only limited information on OOD performance after adjusting for ID accuracy.
\textbf{(2)}~Out-of-distribution generalization is a multi-faceted concept that cannot be reduced to a problem of ``underspecification''~\cite{d2020underspecification} or to simple linear relations between ID and OOD accuracies~\cite{accuracy-line}.
However, we did not observe any trade-off between accuracy and robustness, as is commonly assumed in the domain generalization literature~\cite{arjovsky2019invariant,sagawa2019distributionally,creager2021environment,parascandolo2021learning,krueger2021out}.
While such trade-offs may exist, we posit that they may not be very common in non-adversarially chosen test sets.
\textbf{(3)}~While calibration appears to transfer poorly to new distributions,
adversarial examples and synthetic corruptions transfer well to OOD data but seem ill-suited to mimic natural distribution shifts.
\textbf{(4)}~Held-out OOD validation sets can be good proxies for OOD generalization.
As such, they should be a key focus of any practitioner who worries about distribution shifts at test time.

In light of these results, we suggest three critical areas for further research.
\textbf{(1)}~Creating synthetic interventional distributions is an appealing alternative to hand-crafted augmentations and corruptions to both evaluate and improve robustness. High-fidelity generative models could be used to identify specific axes of variation that a model is not robust to. While this has been studied in the context of fairness with labelled sensitive attributes  \cite[e.g.,][]{zietlow2022leveling}, discovering such factors of variation remains an unsolved task that relates to disentanglement~\cite{locatello2019challenging} and causal representation learning~\cite{scholkopf2021toward}. \textbf{(2)}~While fine-grained studies of OOD performance can shed light into specific generalization properties of neural networks, they should be interpreted with care. In particular, conclusions from adversarially constructed test sets should not be generalized to broader settings. Instead, they may be useful to compile model cards~\cite{mitchell2019model} that contain specific strengths and weaknesses of a model, e.g., in terms of robustness to certain transformations, since we saw that these properties can transfer to new distributions. \textbf{(3)}~More work is needed to understand whether inductive bias in the architecture is a meaningful tool to tackle generic distribution shifts. While we did observe some architecture-specific differences in performance, the many confounding factors during pre-training make it difficult to draw any definitive conclusion on this matter. Experimental protocols that specifically investigate the intrinsic robustness of architectures and its relation to ID accuracy are still required.

\section*{Contributions}
\label{sec:contributions}
\textbf{Florian, Andrea, Peter, Carl-Johann, Max, Dominik, David and Francesco} contributed to the codebase.

\textbf{Max, David and Peter} designed and implemented the first version of the code.

\textbf{Andrea} initiated the first robustness experiments.

\textbf{Florian} conducted the main experiments and prepared the results for further analysis.

\textbf{Florian, Andrea, Carl-Johann, Max, Dominik, Chris and Francesco} contributed to the analysis and the interpretation of the results.

\textbf{Florian and Andrea} conducted the correlation analysis of robustness metrics.

\textbf{Carl-Johann and Chris} conducted the factor analysis.

\textbf{Carl-Johann} prepared the factor loadings and scatter plots, and analyzed the ``many facets of out-of-distribution generalization''.

\textbf{Max} analyzed the in-distribution vs. out-of-distribution performance gap.

\textbf{Dominik} analyzed the effect of augmentation and fine-tuning strategies on robustness. 

\textbf{Francesco} proposed and advised the project.

\textbf{Thomas, Bernt and Bernhard} provided additional valuable insights and regular feedback.

\textbf{All authors} contributed to the writing of the paper.

\textbf{Florian} led the project.

\newpage

\bibliographystyle{plainnat}
\bibliography{refs}

\newpage
\section*{Checklist}

\begin{enumerate}

\item For all authors...
\begin{enumerate}
  \item Do the main claims made in the abstract and introduction accurately reflect the paper's contributions and scope?
    \answerYes{See \cref{sec:introduction,sec:conclusion} and particularly the take-away messages in  \cref{sec:a-broad-look-at-out_of_distribution-generalization,sec:effect-of-augmentations-fine_tuning-stragety-and-few_shot-learning}}
  \item Did you describe the limitations of your work?
    \answerYes{See \cref{sec:effect-of-augmentations-fine_tuning-stragety-and-few_shot-learning,sec:app:limitations}}
  \item Did you discuss any potential negative societal impacts of your work?
    \answerYes{See \cref{sec:app:societal-impact}}
  \item Have you read the ethics review guidelines and ensured that your paper conforms to them?
    \answerYes{}
\end{enumerate}

\item If you are including theoretical results...
\begin{enumerate}
  \item Did you state the full set of assumptions of all theoretical results?
    \answerNA{}
        \item Did you include complete proofs of all theoretical results?
    \answerNA{}
\end{enumerate}

\item If you ran experiments...
\begin{enumerate}
  \item Did you include the code, data, and instructions needed to reproduce the main experimental results (either in the supplemental material or as a URL)?
    \answerNo{See \cref{sec:introduction}, we will release code with the camera-ready version.}
  \item Did you specify all the training details (e.g., data splits, hyperparameters, how they were chosen)?
    \answerYes{See \cref{sec:setup,app:model_selection}}
        \item Did you report error bars (e.g., with respect to the random seed after running experiments multiple times)?
    \answerNo{For all the numbers/data points we report, we are averaging over multiple experiments (in most cases more than 100). However, we do not average over restarts of single configurations since restarts (multiple seeds) would have further increased the anyways high computational effort.}
        \item Did you include the total amount of compute and the type of resources used (e.g., type of GPUs, internal cluster, or cloud provider)?
    \answerYes{See \cref{app:compute_overview}, $\sim17$ GPU years on Nvidia T4 GPUs (cloud hosted)}
\end{enumerate}

\item If you are using existing assets (e.g., code, data, models) or curating/releasing new assets...
\begin{enumerate}
  \item If your work uses existing assets, did you cite the creators?
    \answerYes{See \textit{Experimental protocol and datasets} and \textit{Models} in \cref{sec:setup}}
  \item Did you mention the license of the assets?
    \answerNo{}
  \item Did you include any new assets either in the supplemental material or as a URL?
    \answerNo{}
  \item Did you discuss whether and how consent was obtained from people whose data you're using/curating?
    \answerNA{}
  \item Did you discuss whether the data you are using/curating contains personally identifiable information or offensive content?
    \answerNA{}
\end{enumerate}

\item If you used crowdsourcing or conducted research with human subjects...
\begin{enumerate}
  \item Did you include the full text of instructions given to participants and screenshots, if applicable?
    \answerNA{}
  \item Did you describe any potential participant risks, with links to Institutional Review Board (IRB) approvals, if applicable?
    \answerNA{}
  \item Did you include the estimated hourly wage paid to participants and the total amount spent on participant compensation?
    \answerNA{}
\end{enumerate}

\end{enumerate}

\newpage
\appendix

\renewcommand\thefigure{S\arabic{figure}}
\renewcommand\thetable{S\arabic{table}}
\renewcommand\theequation{S\arabic{equation}}
\renewcommand{\thesection}{\Alph{section}}
\renewcommand{\thealgorithm}{S\arabic{algorithm}}
\setcounter{figure}{0}
\setcounter{table}{0}
\setcounter{equation}{0}

\section{Details on the averaged correlation matrices}
\label{app:corr_matrices}

\subsection{Method details}\label{app:corr_matrices:method}
Our goal is to measure how predictive a \emph{metric $A$} is of another \emph{metric $B$}. We measure the predictiveness based on Spearman's rank correlation coefficient between all combinations of metrics on ID, OOD, corrupted and held-out OOD data.
One issue with this measure however is that, while our empirical study covers multiple tasks with multiple (ID, OOD) dataset pairs, a correlation coefficient can only be computed for a specific dataset pair: pooling the data together from multiple dataset pairs (experiments) would lead to skewed results.
For instance, for one dataset pair the ID and OOD datasets could be intrinsically hard, leading to low metric values, and for another dataset pair, the datasets could be intrinsically easy, leading to high values.
Pooling the data of those two dataset pairs would lead to the impression of a clear (linear) trend which would not be necessarily present when considering the dataset pairs individually.
Hence, trends should only be considered at the individual dataset level.
Other studies circumvent this problem by reporting the correlation coefficients for each dataset pair individually (e.g., see \cite[Table~1]{accuracy-line}), but since we have a much larger study with $172$ dataset pairs, such a table would not be informative.
Therefore, we report the \emph{averaged correlation coefficient} of the $172$ dataset pairs.
We note that \emph{this score cannot be interpreted as a statistical correlation} (due to the averaging) \emph{but still serves as a meaningful measure of predictiveness}.
For example, if we think in terms of Pearson (rather than Spearman) correlations, then an average (Pearson) correlation of 1 would mean that, on every dataset pair, we observe a linear dependency, but the parameters of this dependency (slope and intercept) could be different for every dataset pair.
In particular, it would not mean that the pooled data lies on a single straight line.

The averaged correlation matrix is displayed in \cref{fig:full_corr_matrix} and computed as follows.
For a given (ID, OOD) dataset pair, a metric $A$, and a metric $B$, we first compute the Spearman's rank correlation coefficient~\cite{spearman04} of the metrics on that dataset pair. To this end, we pool the data of all model architectures, augmentation and fine-tuning methods and compute the correlation of the corresponding metric scores. Second, to obtain an aggregated predictiveness score over all dataset pairs, we computed the \emph{weighted average} of the correlation coefficients for all dataset pairs in each task. Since some tasks entail more dataset pairs then others, we compute a weighted average, such that each task has the same contribution to the final average predictiveness score. In summary, the averaged correlation coefficient between metric $A$ and metric $B$ is
\begin{align*}
    \textrm{score}(A,B) = \sum_{i\in I} w_i \textrm{corr}\left(A|\textrm{ID}_i; B|\textrm{OOD}_i\right),
\end{align*}
where $A|\textrm{ID}_i$ are the metric scores of metric $A$ evaluated on the dataset $\textrm{ID}_i$ and  $B|\textrm{OOD}_i$ are the metric scores of metric $B$ evaluated on the dataset $\textrm{OOD}_i$. $I$ is the index set of all dataset pairs. The weight $w_i=\nicefrac{1}{|T_i|}$ normalizes by the number of dataset pairs in task $T_i$. See \cref{app:datasets} on how the tasks are defined. In addition to the (ID, OOD) dataset pairs, we also compute the correlation of metrics on all other combinations of ID, OOD, and held-out OOD domains.

\paragraph{Remark: Pearson correlation coefficients.}
Instead of computing rank correlation coefficients between metrics, we also tried using Pearson (linear) correlation coefficients. Qualitatively, this led to similar results as using rank correlations (\cref{fig:full_corr_matrix}). The order of the predictiveness of metrics stayed the same (e.g., ID classification error still had the highest correlation coefficient with OOD classification error). However, since some trends are not linear (see \cref{app:scatter_plots}) we found that rank correlation coefficients lead to more robust results.

\paragraph{Adjusted predictiveness scores.}
As briefly described in \cref{sec:metrics_predict_ood}, we assess the usefulness of a metric to explain OOD accuracy by the additional information they provide on top of what is already explained by ID accuracy. The adjustment is similarly done as in \cite{taori-shifts}. For each (ID, OOD) dataset pair, we first compute a linear regression between ID and OOD classification error. The residuals of that regression is the variance in the data that is not explained by ID classification error. Second, we compute the correlation coefficient of the metric of interest with the residuals, i.e., we compute how informative the metric is of the variance that is not already explained by ID accuracy. Again, we repeat this procedure for all (ID, OOD) dataset pairs and compute the weighted average as explained in \cref{sec:metrics_predict_ood}. The adjusted predictiveness scores are displayed in \cref{fig:barplot_corr_unified} (green bars).

The adjusted predictiveness scores give a better sense of the usefulness of the metrics. For instance, inspecting \cref{fig:full_corr_matrix}, the averaged correlation between ID corrupted classification error and OOD classification error is $0.44$, which is only $0.05$ points lower than the score for ID classification error. Only looking at this number, we would be tempted to conclude that the ID corrupted classification error is actually a good predictor of robustness and, hence, might be worth the additional compute burden. However, if we compute the adjusted predictiveness score we obtain a score of $0.01$, see \cref{fig:barplot_corr_unified} (left, green bar). Hence, ID corrupted classification error provides no information that is not already covered by the standard ID classification error. This is also reflected by the fact that ID classification error and ID corrupted classification error are highly correlated (with an averaged correlation coefficient of $0.68$, see \cref{fig:full_corr_matrix}).

\subsection{The full correlation matrix}

\begin{figure}[tb]
    \centering
    \includegraphics[width=0.95\textwidth]{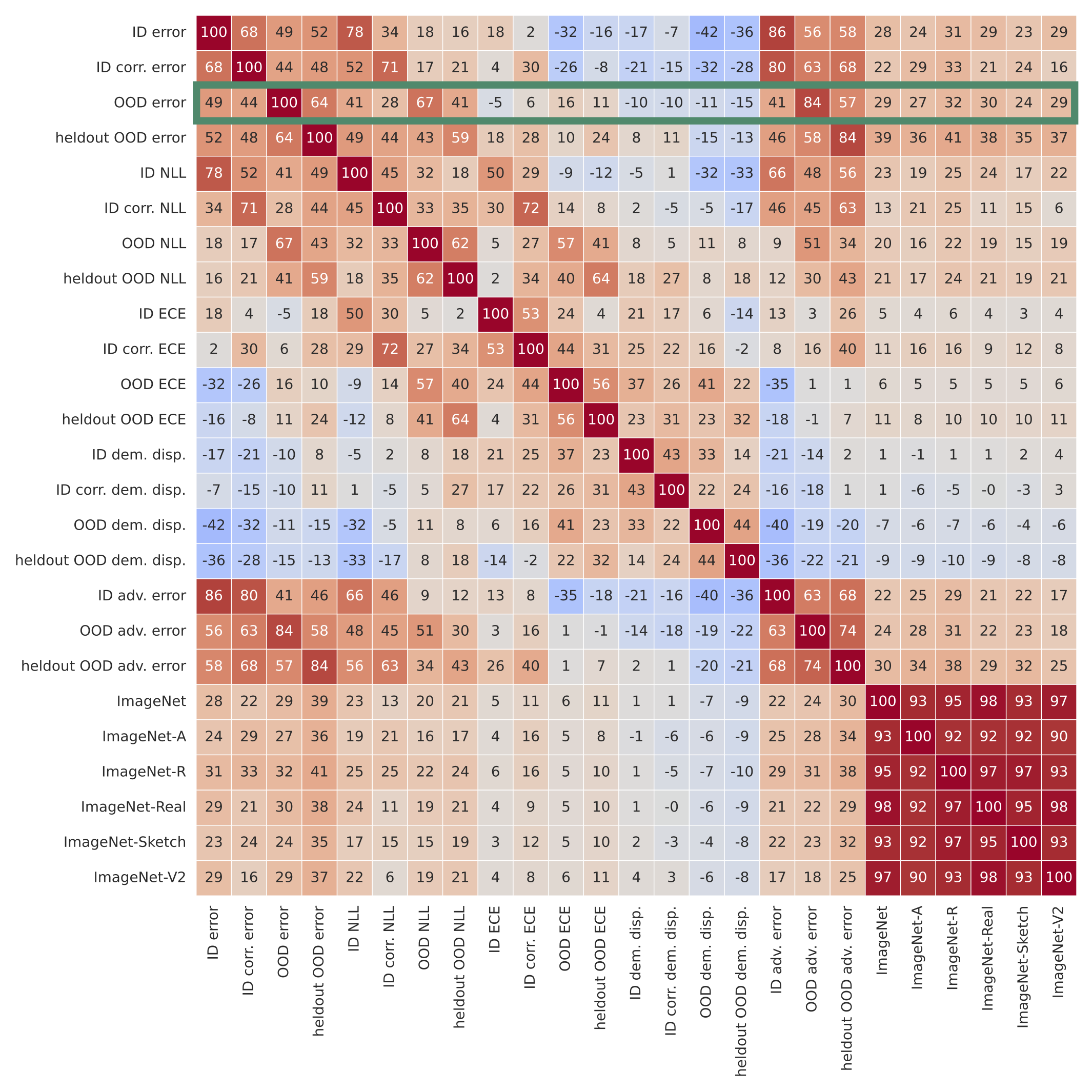}
    \caption{Rank correlation matrix averaged over all tasks. The matrix shows the predictiveness of a metric (either evaluated on ID data, ID corrupted data, OOD, held-out OOD data or ImageNet) of another metric. Positive values indicate a positive link between the metrics, i.e. better results for one metric tend to lead to better results for the other metric. Negative values indicate a negative link, i.e. better results for one metric tend to lead to worse results for the other metric.  All averaged correlation coefficients are multiplied by a factor of $100$ for better readability. As an example, the highlighted \textcolor{factorgreen}{\textbf{green rectangle}} shows how well metrics predict OOD classification error, this information is shown separately as a bar plot in~\cref{fig:barplot_corr_unified} (red bars) in the main text.}
    \label{fig:full_corr_matrix}
\end{figure}

We display the full averaged correlation matrix in \cref{fig:full_corr_matrix}. It shows the pair-wise averaged correlation coefficient between all combination of metrics and is computed as described in \cref{app:corr_matrices:method}. This matrix is the main source for our evaluations. For instance, \cref{fig:barplot_corr_unified} highlights one part of this matrix, focusing on the correlation of metrics with OOD classification error. The relevant row in the matrix is marked by a green rectangle.

\subsection{Analyzing each task separately}
\label{app:correlation_matrix_per_task}
\begin{figure}
    \centering
    \includegraphics[width=0.49\textwidth]{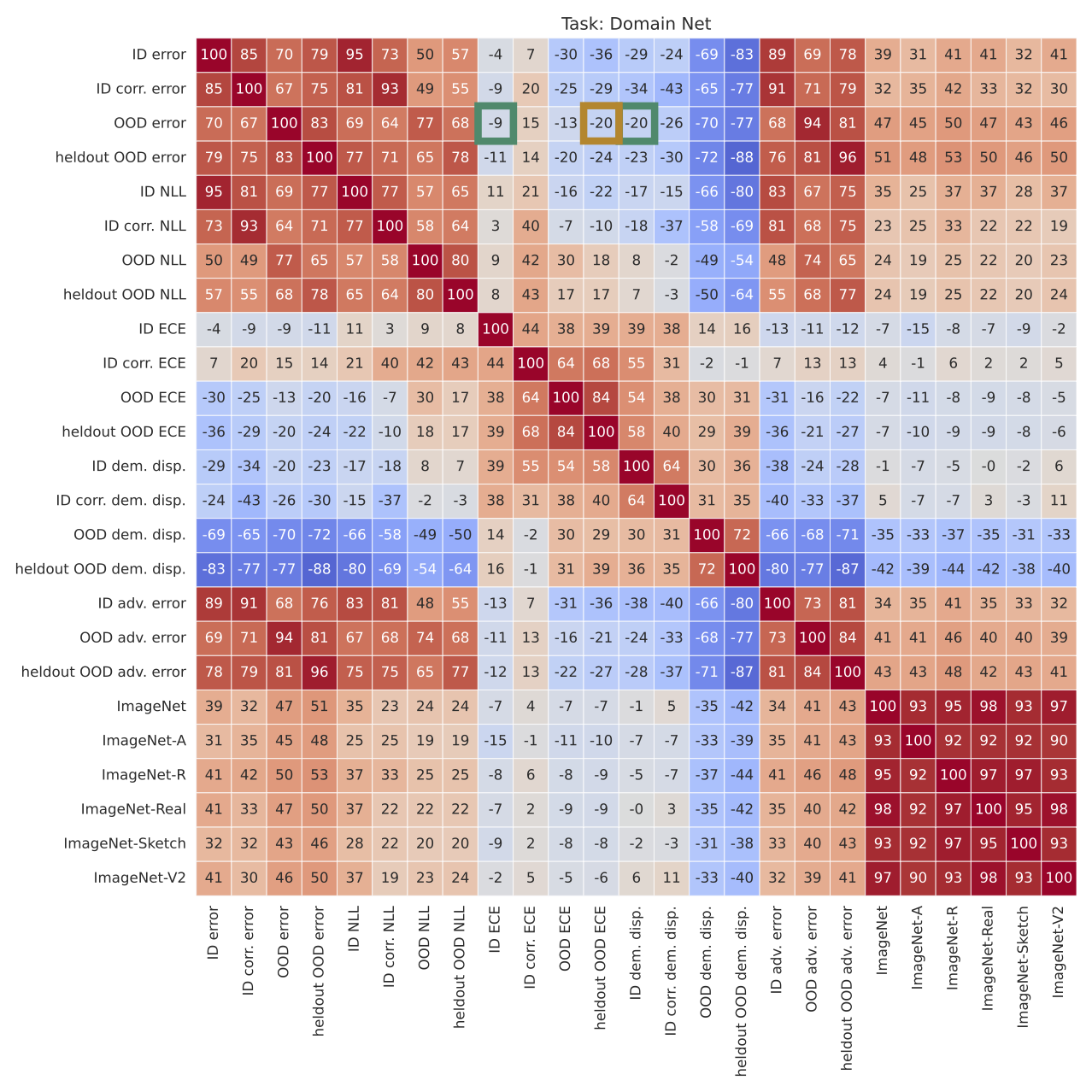}
    \hfill
    \includegraphics[width=0.49\textwidth]{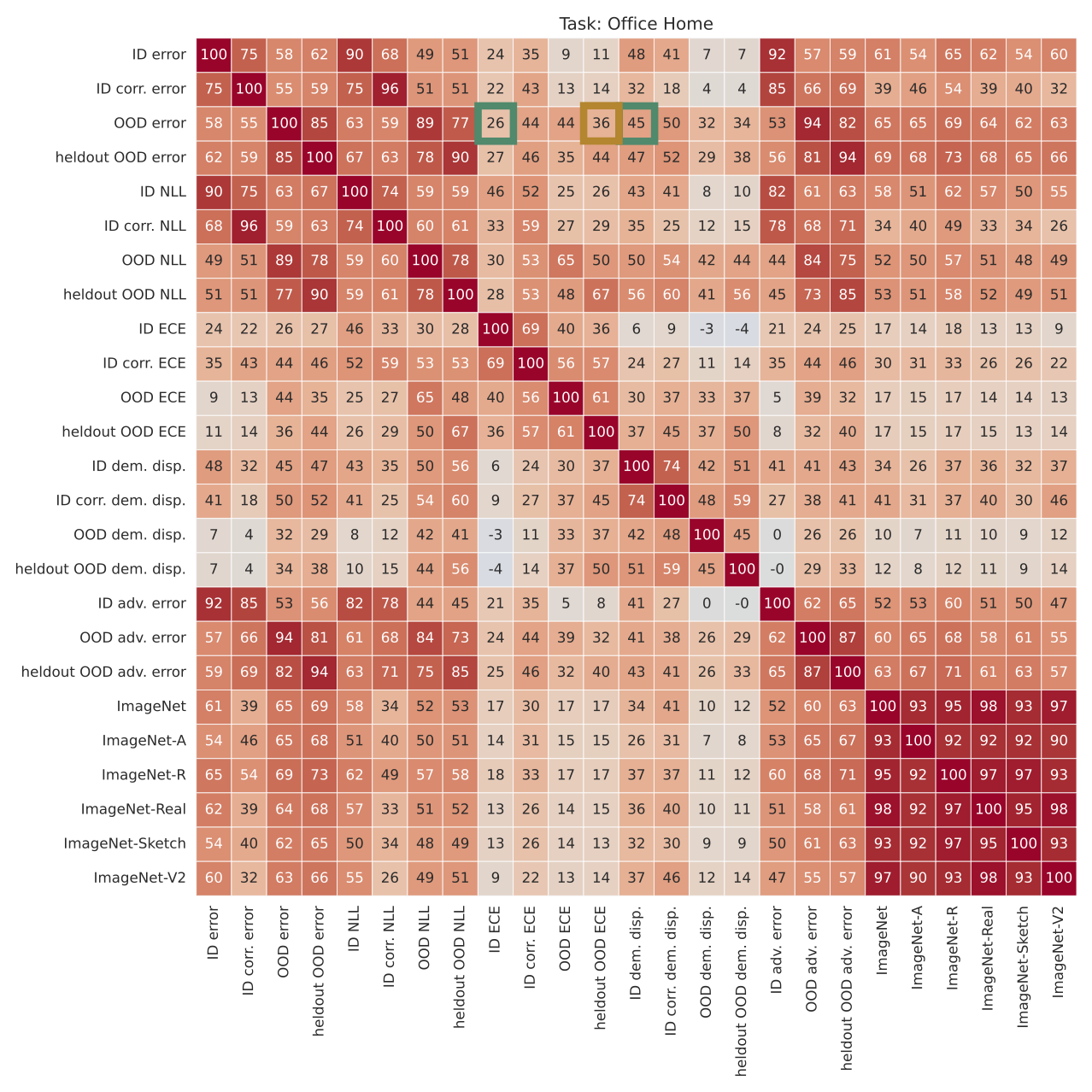}

    \includegraphics[width=0.49\textwidth]{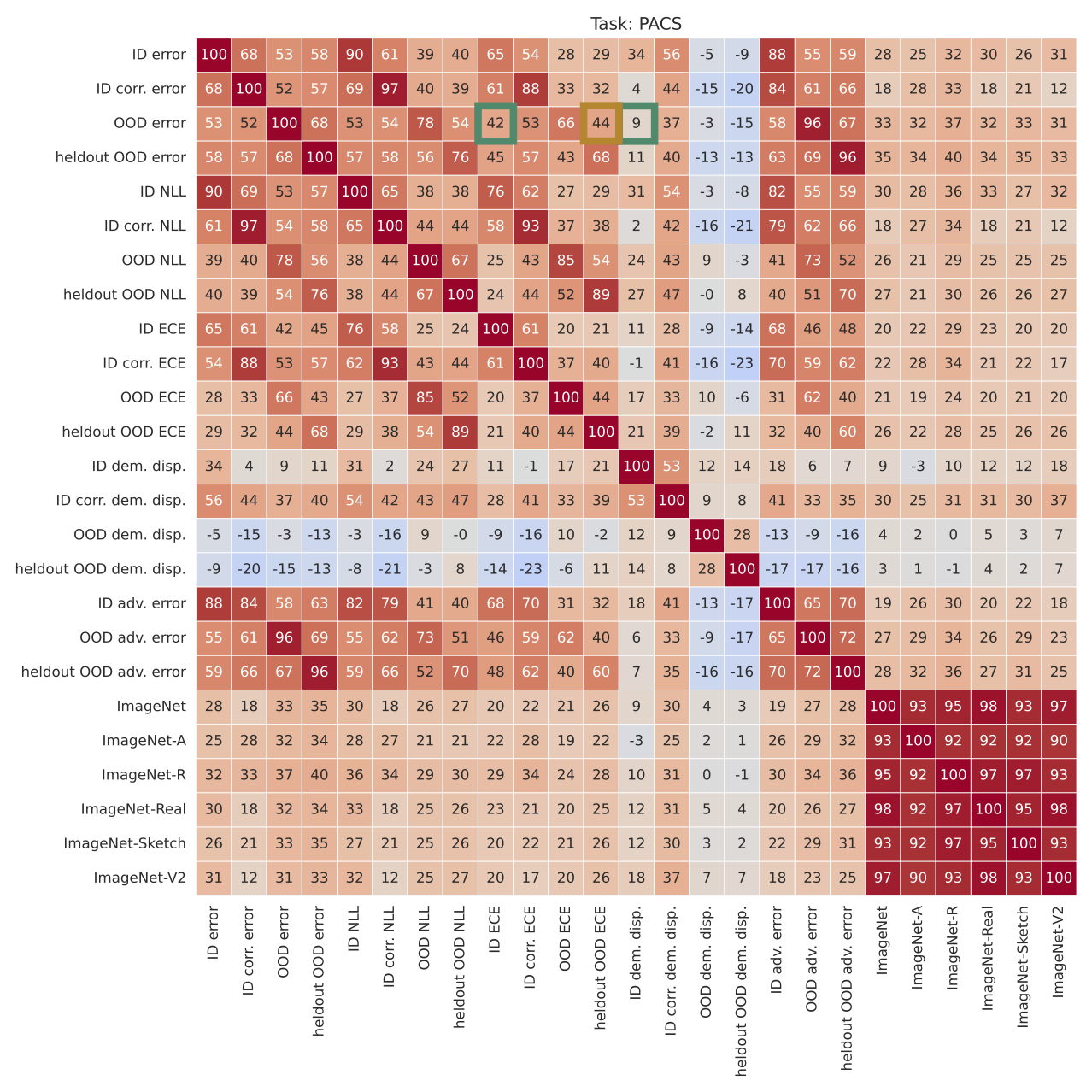}
    \hfill
    \includegraphics[width=0.49\textwidth]{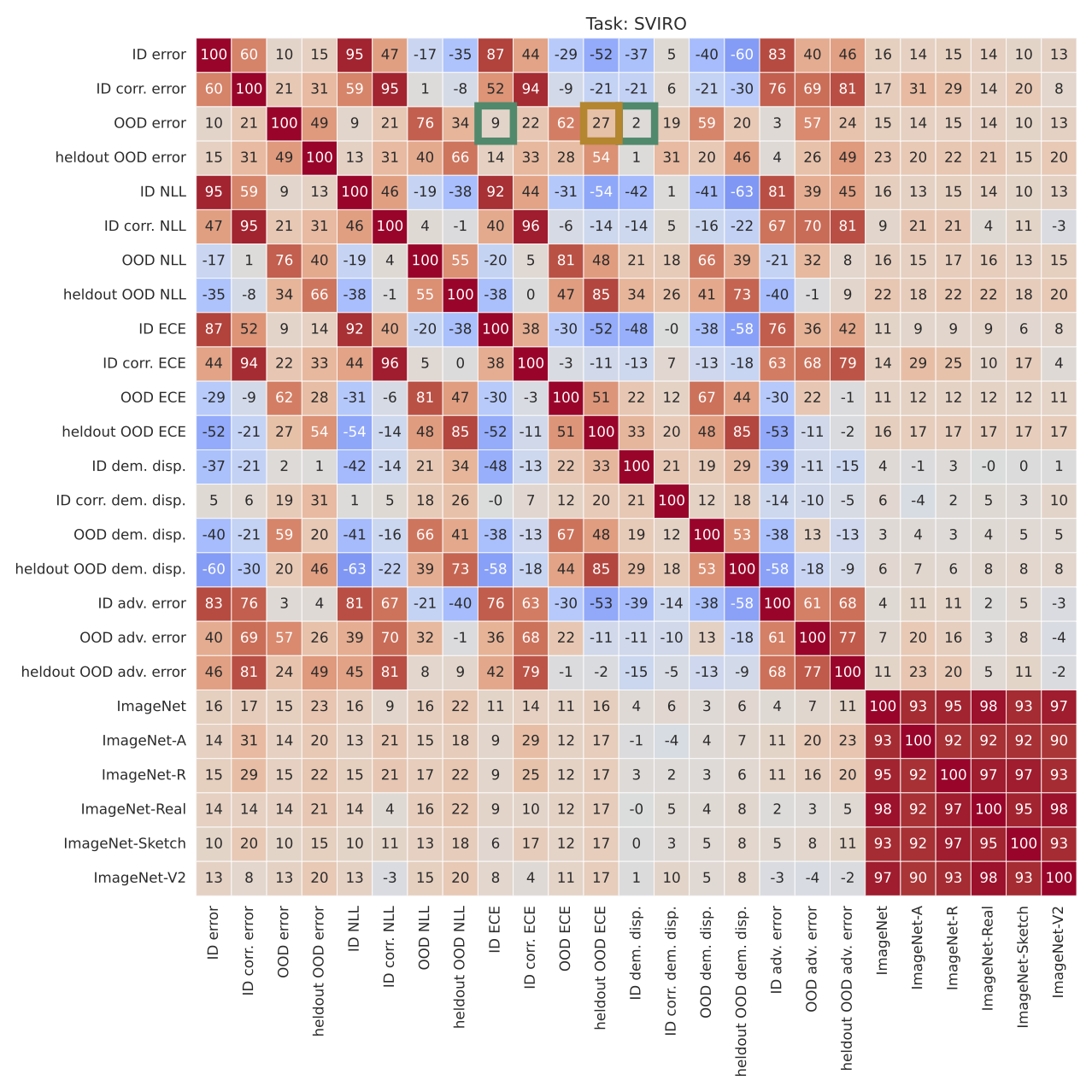}
    \caption{Averaged correlation matrices for each task separately. For each task we only consider the corresponding datasets and compute the correlation coefficients separately. We highlight differences between tasks for ID demographic disparity, ID calibration with \textcolor{factorgreen}{\textbf{green squares}} and differences for held-out ECE with \textcolor{factororange}{\textbf{yellow squares}}.}
    \label{fig:corr_matrix_by_task_1}
\end{figure}

\begin{figure}
    \centering
    \includegraphics[width=0.49\textwidth]{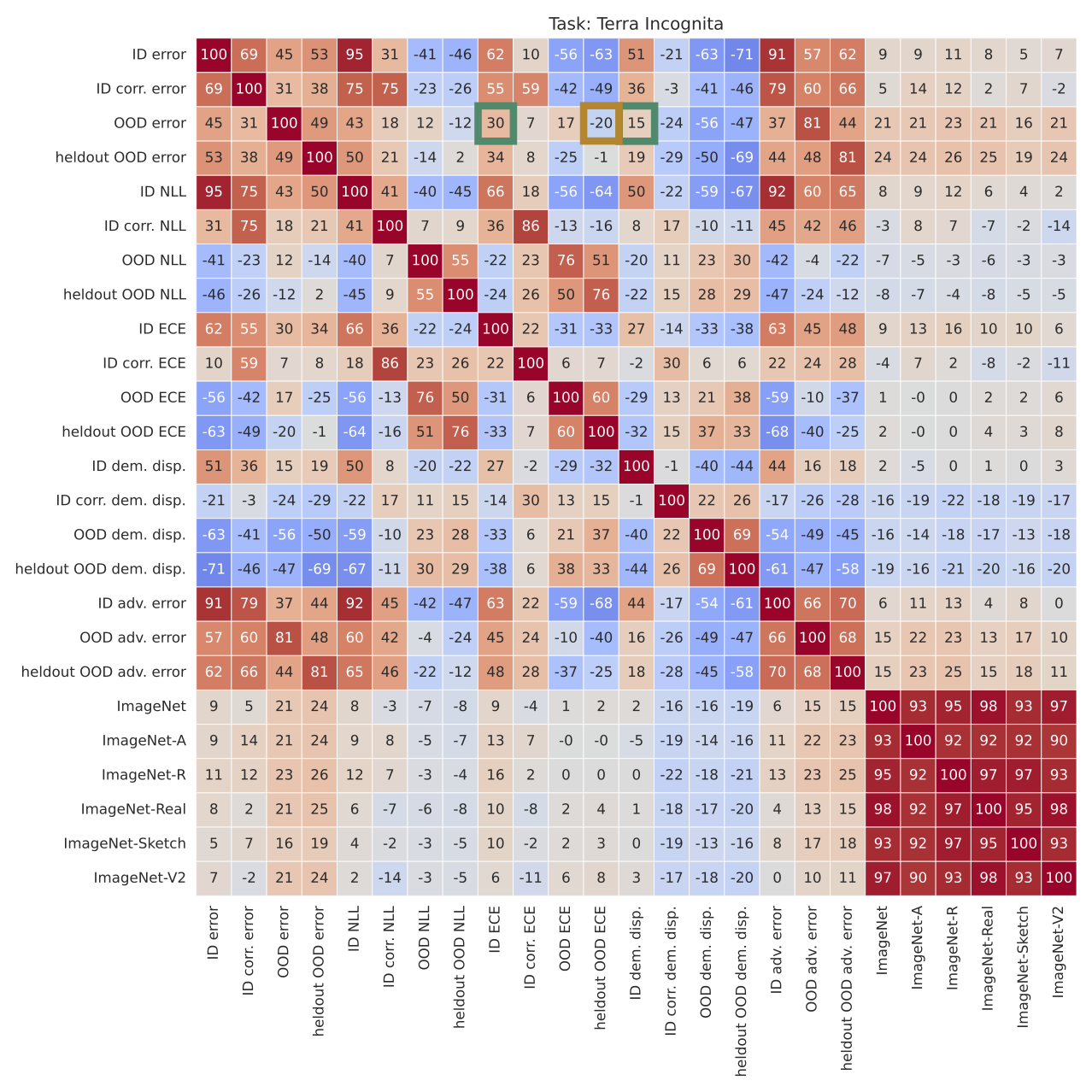}
    \hfill
    \includegraphics[width=0.49\textwidth]{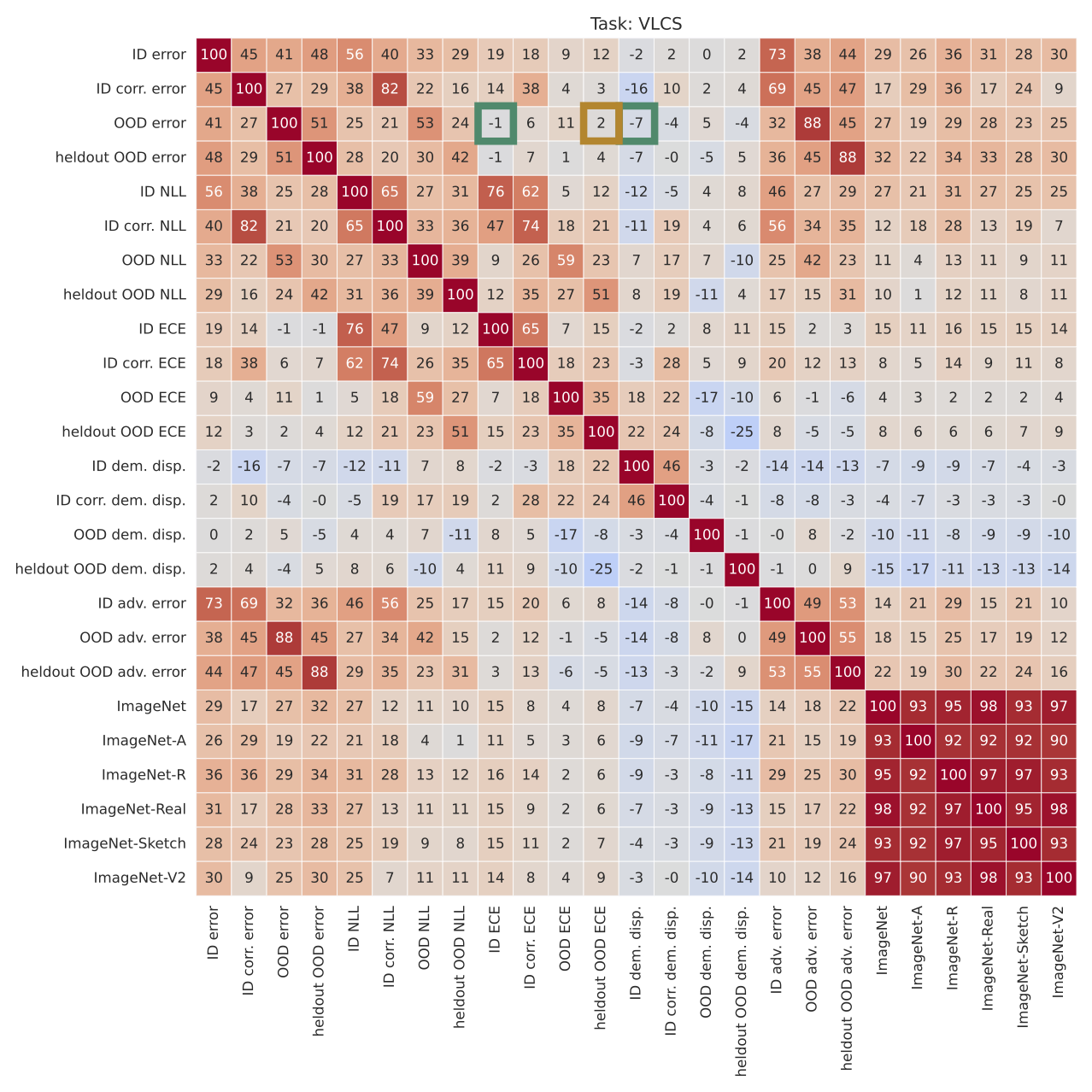}

    \includegraphics[width=0.49\textwidth]{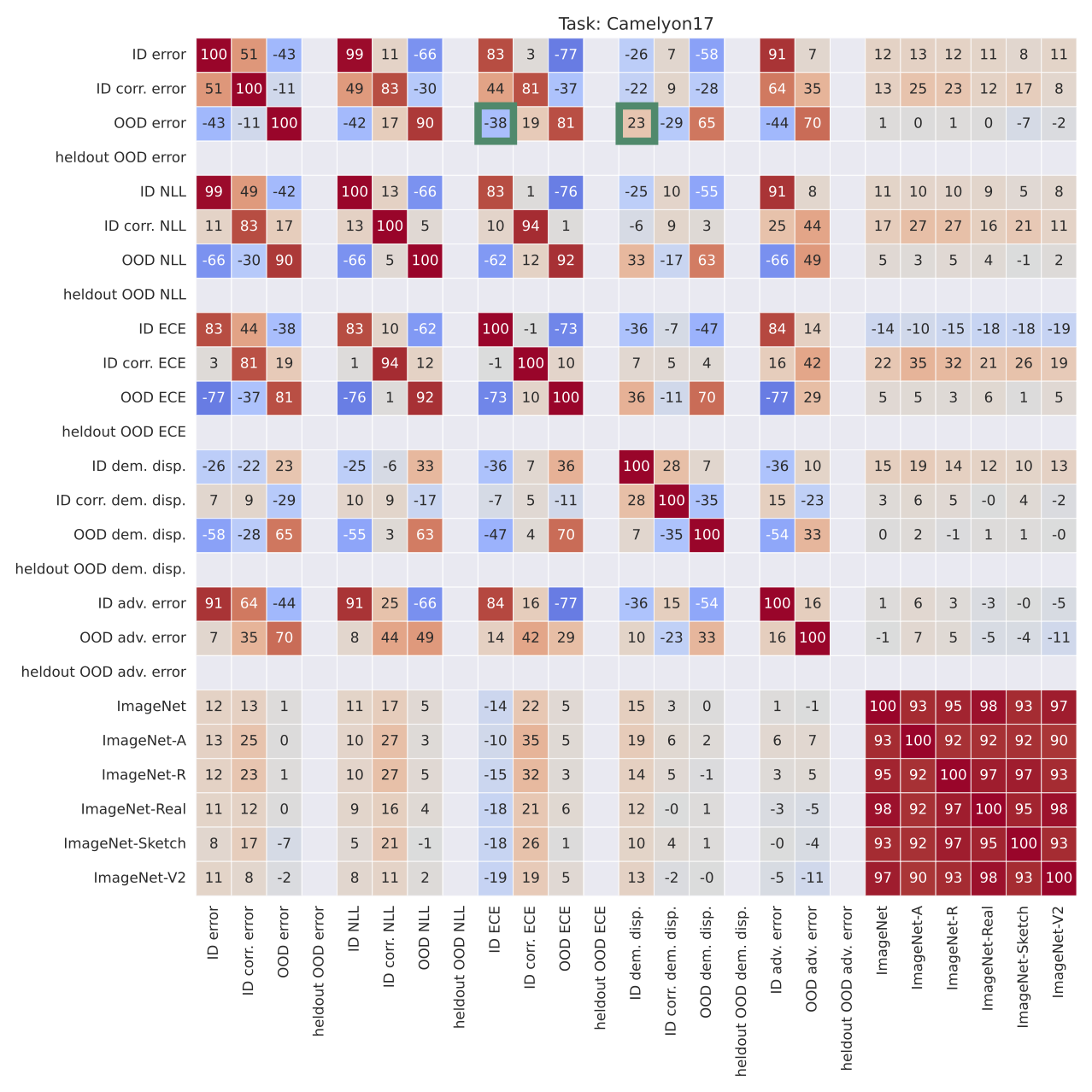}
    \hfill
    \includegraphics[width=0.49\textwidth]{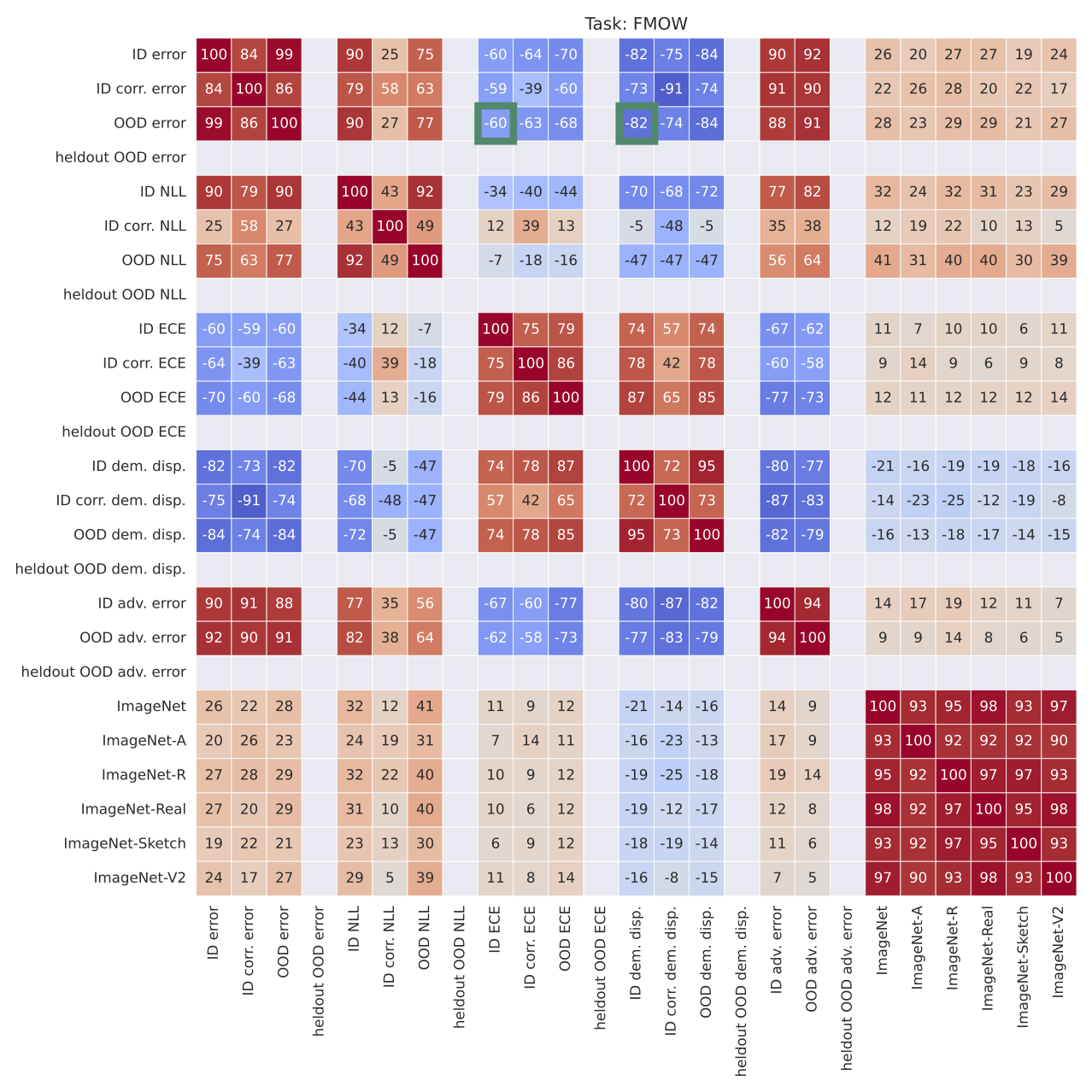}

    \includegraphics[width=0.49\textwidth]{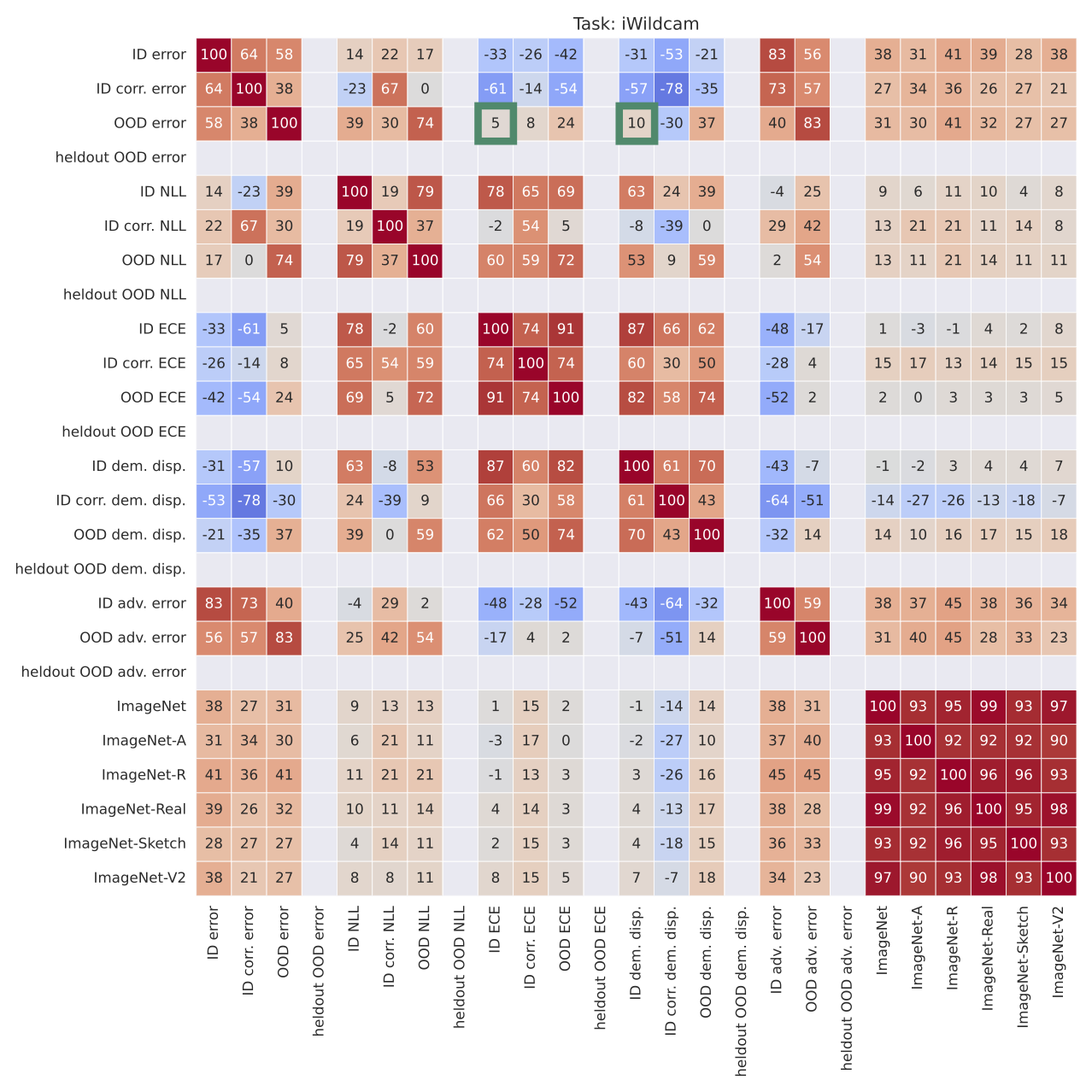}
    \hfill
    \includegraphics[width=0.49\textwidth]{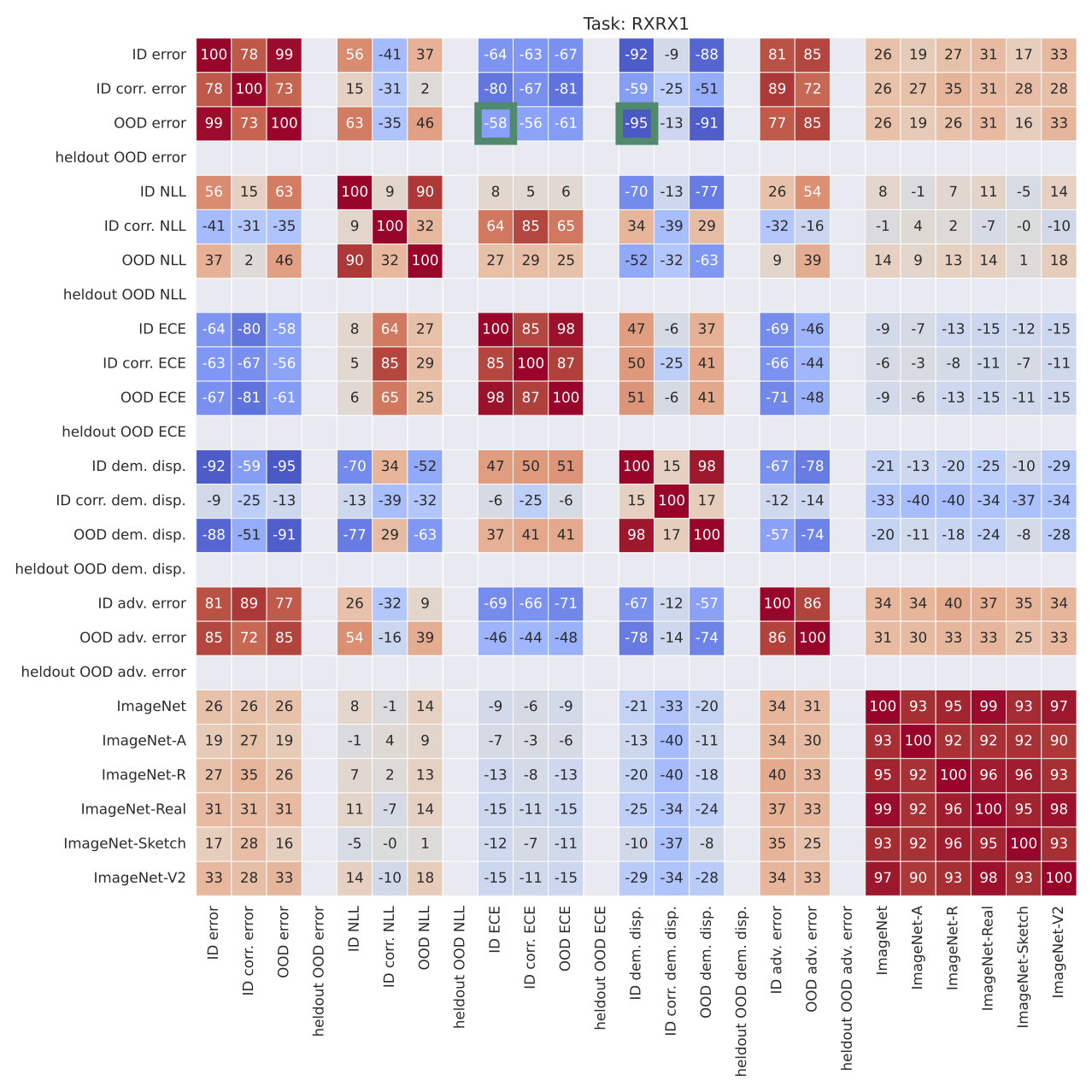}
    \caption{Averaged correlation matrices for each task separately (continued). We highlight differences between tasks for ID demographic disparity, ID calibration with \textcolor{factorgreen}{\textbf{green squares}} and differences for held-out ECE with \textcolor{factororange}{\textbf{yellow squares}}. Note that for the tasks \emph{Camelyon17}, \emph{FMOW}, \emph{iWildcam} and \emph{RXRX1}, no held-out OOD data is available (i.e., only one OOD split is provided).}
    \label{fig:corr_matrix_by_task_2}
\end{figure}

In \cref{sec:acc-not-on-line} in the main text we argue that out-of-distribution generalization has many facets. Some patterns observed in prior work actually only hold on a subsets of the tasks included in our study. To investigate the stability of observations across different tasks we compute the averaged correlation matrices for each task separately. For each task, we only include the datasets that belong to that tasks (see \cref{app:datasets}) and show the matrices in \cref{fig:corr_matrix_by_task_1,fig:corr_matrix_by_task_2}.

Some observations are stable across all tasks (e.g., classification error is always the strongest predictors on ID and held-out data, respectively). For other metrics we observe large variability across tasks. For instance for some tasks, demographic disparity has a strong positive link to OOD accuracy, for others it is reversed (i.e., here lower demographic disparity leads to a higher classification error). This also holds for calibration error. We marked the according cells in the task-specific correlation matrices with a green square. We also observe mixed positive and negative links for multi-domain calibration (held-out OOD ECE, marked by yellow squares in the plots). This is in contrast to the claims by~\cite{multidomain-calibration} that multi-domain calibration is generally a good indicator for OOD robustness since it actually only holds on a subset of the tasks.

\subsection{Analyzing each shift type separately}
\label{app:correlation_matrix_by_shift}
In this section we investigate how results change for different shift types. We define the shift type based on ID dataset regime and the OOD dataset regime as follows. First, we label each dataset by one of the domain types \emph{artificial} and \emph{real}, see \cref{tab:task_domain}. For instances, the label \emph{artificial} refers to datasets of images composed from sketches, clipart or simulated environments.
Second, for an (ID, OOD) dataset pair the shift is derived by the ID (source) domain type and the OOD (target) domain type. E.g., the dataset pair: (OfficeHome-ClipArt, OfficeHome-RealWorld) is labeled by the shift type \emph{artificial $\rightarrow$ real}.

\begin{figure}
    \centering
    \includegraphics[width=0.49\textwidth]{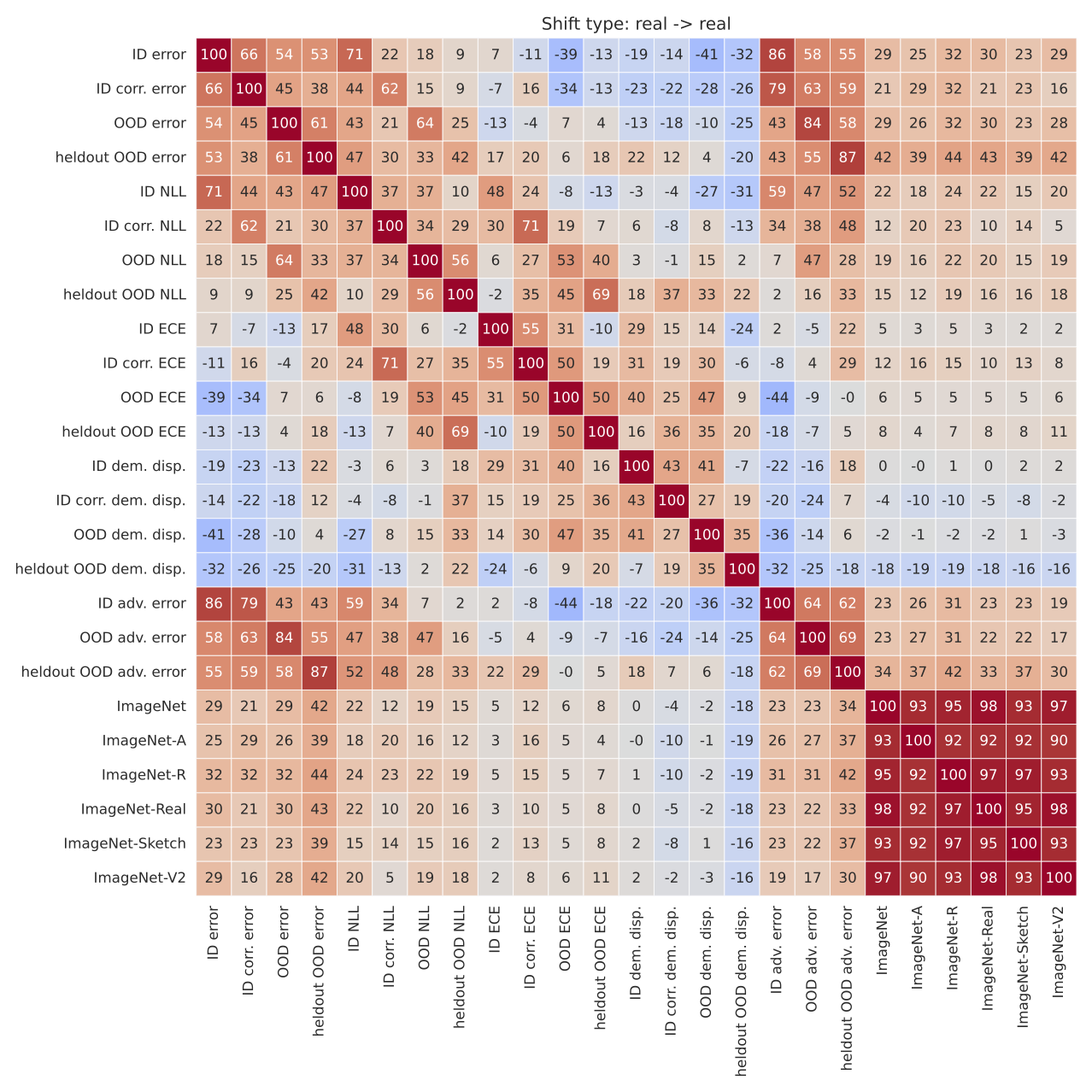}
    \hfill
    \includegraphics[width=0.49\textwidth]{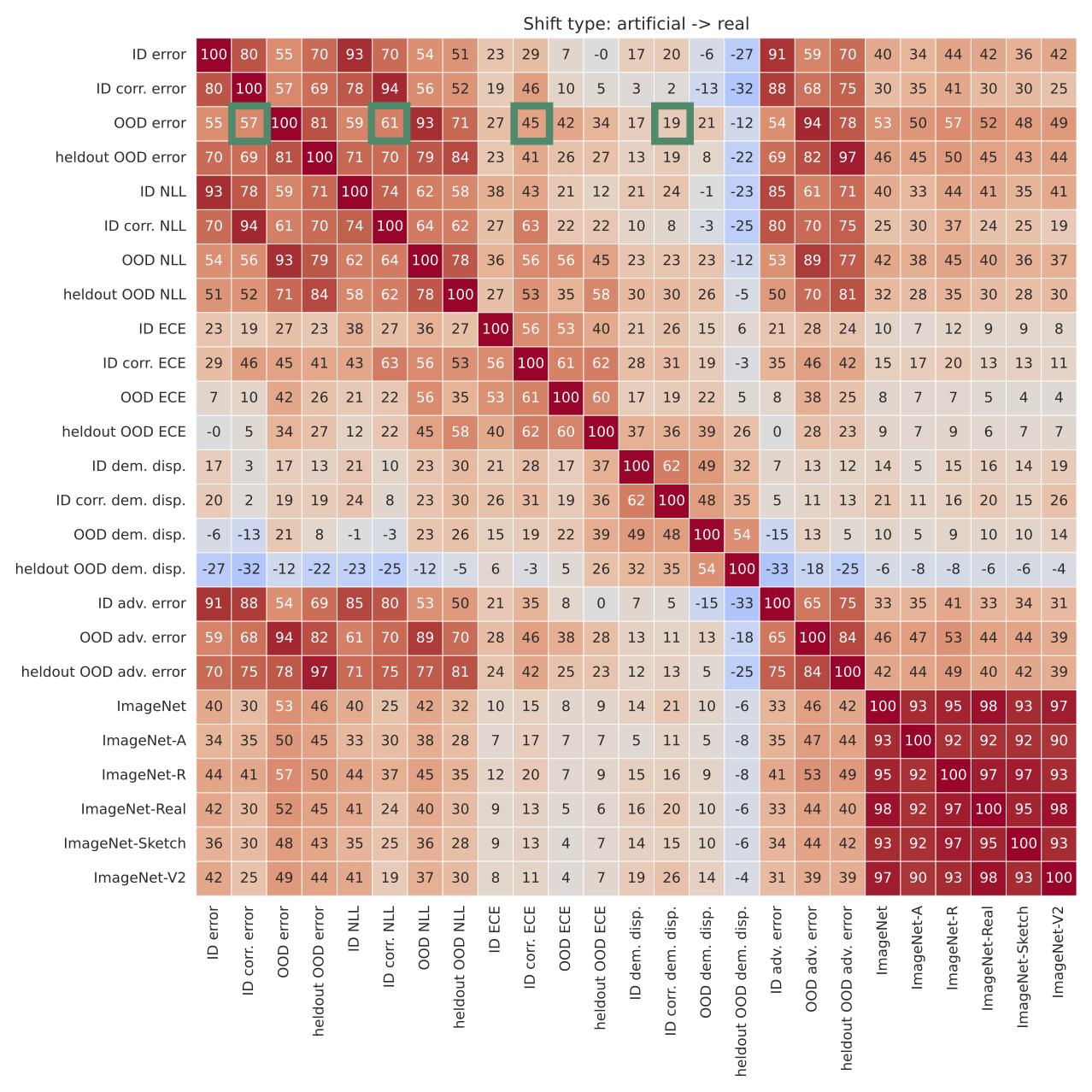}

    \includegraphics[width=0.49\textwidth]{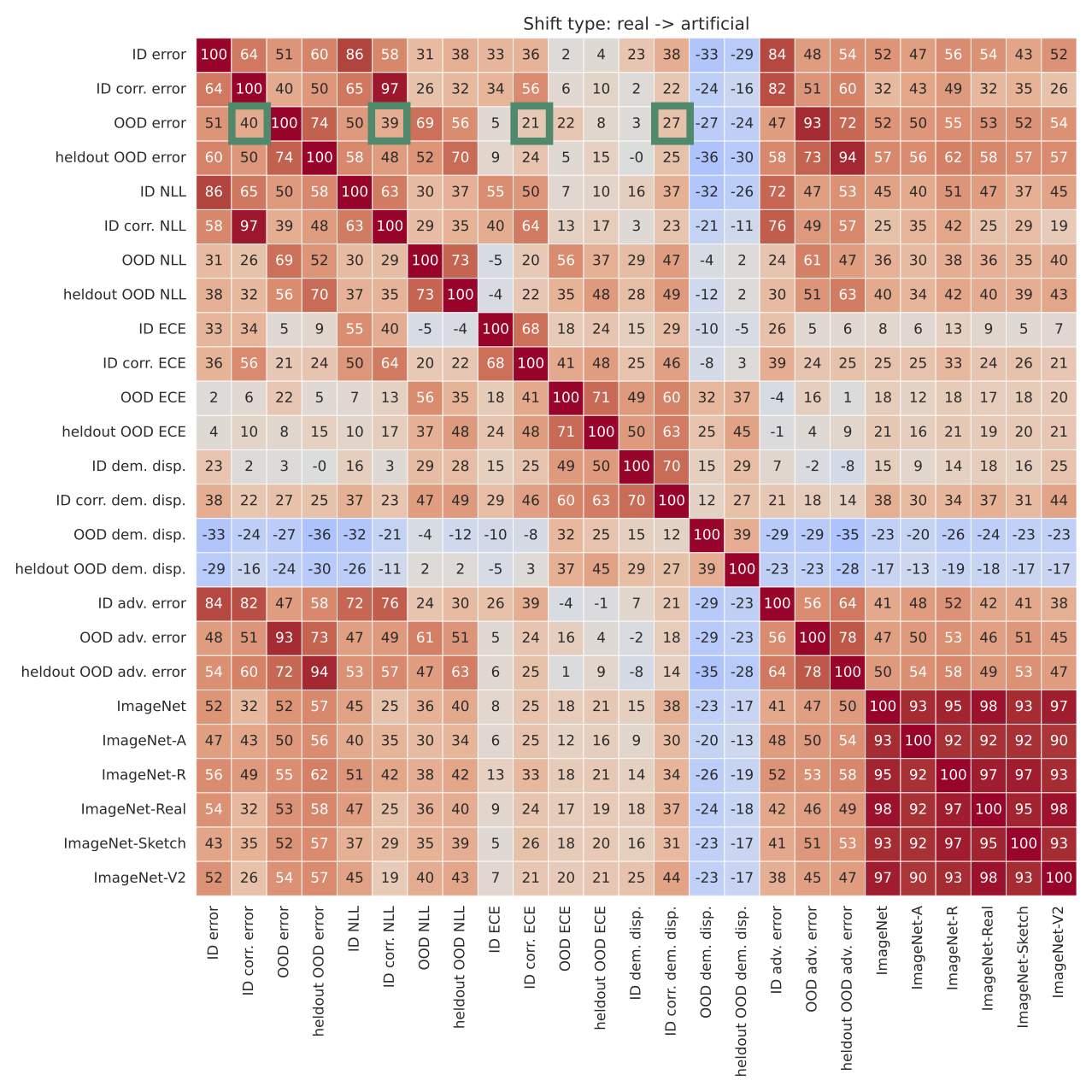}
    \hfill
    \includegraphics[width=0.49\textwidth]{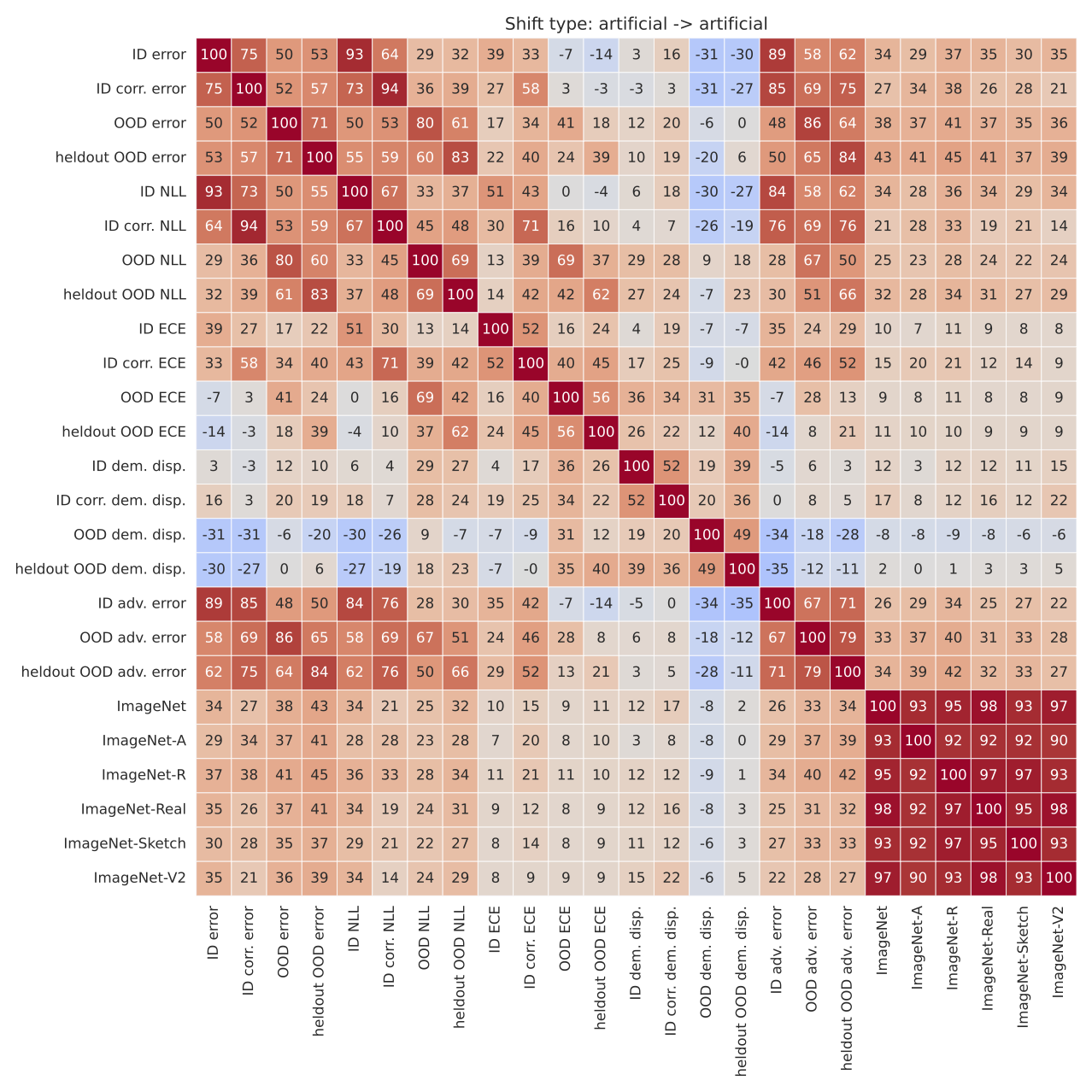}
    \caption{Averaged correlation matrices for each shift type separately. The shift types are defined based on the source and target domain, see~\cref{tab:task_domain}. Using \textcolor{factorgreen}{\textbf{green squares}} we highlight differences of the corruption metrics predicting OOD accuracy for the shift types \emph{real $\rightarrow$ artificial} and \emph{artificial $\rightarrow$ real}.}
    \label{fig:corr_matrix_by_shift}
\end{figure}

Similarly, as in \cref{app:correlation_matrix_per_task}, we compute the averaged correlation matrix for the subset of dataset pairs corresponding to each shift type separately. The matrices are displayed in the figures~\cref{fig:corr_matrix_by_shift}.
For instance, for the shift type \emph{artificial $\rightarrow$ real}, corruption metrics are more predictive of OOD accuracy then for the reversed shift type \emph{real $\rightarrow$ artificial}, see the highlighted green cells in the plots.


\FloatBarrier

\section{Factor analysis: details and background}
\label{app:factor-analysis}
\begin{figure}[tb]
    \centering
    \includegraphics[width=0.5\textwidth]{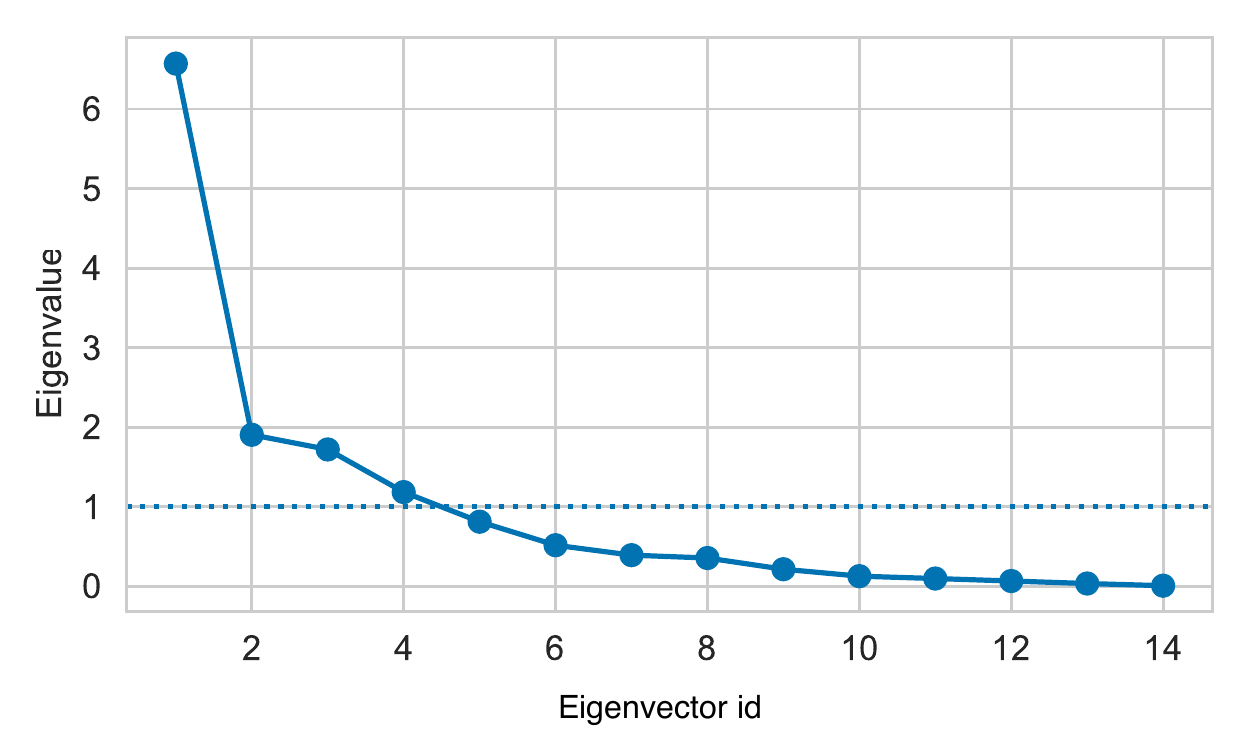}
    \caption{\emph{Scree plot}: Eigenvalues of the standardized data from the factor analysis, sorted by decreasing order of magnitude.
    As explained in \cref{sec:factor-analysis}, we decided to take the first $4$ factors, since only the first $4$ eigenvalues are above the usual threshold value of $1$.}
    \label{fig:scree-plot}
\end{figure}

This section gives a short introduction to factor analysis, provides details about the data and the pre-processing used for the factor analysis of \cref{sec:factor-analysis} and presents the data's scree plot in \cref{fig:scree-plot}.

\paragraph{Background.}
Factor analysis is a well-established tool from statistics used to analyze tabular data, such as data arising from a poll, where the columns would be the questions and the rows the answers of each individual.
It aims at finding a minimal amount of latent variables called \emph{factors} that suffice to explain the data and reflect the main dependencies between columns.
In particular, given $k$ factors $f_1, f_2, \ldots, f_k$, each column $c_i$ gets approximated as $c_i = \lambda_{i1} f_1 + \cdots \lambda_{ik} f_k$ where the $\lambda_{ij}$s are called factor \emph{loadings}.
Because columns get standardized prior to the analysis and factors are scaled to have unit norm, the loadings lie between $-1$ and $1$.
While the factors are a priori abstract variables with no pre-defined interpretation, one of the main challenges and goals of factor analysis is to understand what variability they capture by analyzing their loadings.
Note that the first $k$ factors span the subspace defined by the first $k$ eigenvectors of the data matrix.
However, the factors themselves are only defined up to a rotation, since rotating the factors and the loadings accordingly, the $c_i$s remain unchanged.
Some rotations lead to factors whose loadings are easier to interpret than others.
Therefore, the typical workflow of a factor analysis is as follows.
(1)~Decompose the data into eigenvectors and use the eigenvalues to choose the number $k$ of factors to keep.
The simplest rule of thumb is to keep all eigenvectors with eigenvalues $\geq$ 1.
(2)~Try out different rotations of the $k$ first eigenvectors, to make the loadings as easy as possible to explain or interpret.
To do so, one typically just tests several standard rotation methods such as the \emph{varimax}, \emph{quartimax}, or \emph{equamax} methods (which typically try to promote some form of sparse loadings).
(3)~Interpret the obtained factors and conclude on the relations between the columns that they show.

\paragraph{Data preprocessing.}
We organize our data in a table with rows being fine-tuned networks and columns the metrics.
Specifically, each network is defined by its model architecture, the adaptation dataset used for fine-tuning and the training parameters (augmentation strategy, learning rate, number of training epochs, and fine-tuning method), leading in theory to a total of $7776$ networks ($9$ architectures $\times$ $36$ datasets $\times$ $3 \cdot 2 \cdot 2 \cdot 2$ parameters).
Note that we considered only those networks finetuned on full datasets, ignoring the few-shot data.
For each network, we considered a total of $16$ metrics: the $6$ base metrics (cf.\ \cref{sec:setup}) computed respectively on the held-out test data from the adaptation domain (the in-distribution metrics), on its corrupted variant (except adversarial accuracy, which we did not compute on corrupted data), and, for each metric, its \emph{average} over all out-of-distribution datasets from the adaptation dataset's task.
Concerning the log-likelihood metric, most values lied between $0$ and $5$, but some outliers could take values up to $1\mathrm{e}{10}$.
We therefore mapped all log-likelihood values through the following function:
$f(x) = 10 ( 1 - \exp(-\nicefrac{x}{10}) )$.
Doing so ensures that all typical log-likelihood values remain nearly unchanged up to a rescaling factor ($f$ is almost linear in $0$), while bigger values saturate at $10$.
We used python's \texttt{factor\_analyzer} package for the analysis.

\paragraph{Factor analysis and factor loading plot.}
The scree plot in \cref{fig:scree-plot} shows the eigenvalues of the data in decreasing order.
We decided to retain $4$ factors, since only $4$ eigenvalues were $\geq 1$.
We then used the \emph{equamax} rotation method, but the results do not change significantly with other standard methods.
The factor loadings for each metric are shown in \cref{fig:factor-analysis} in the main text and discussed there (see \cref{sec:factor-analysis}).

\FloatBarrier
\section{Detailed analysis on the effect of the fine-tuning and augmentation strategies}
In the following we provide additional studies on the effect of the augmentation and fine-tuning method on the ID and OOD performance. The main results are summarized in the main text in \cref{sec:effect-of-augmentations-fine_tuning-stragety-and-few_shot-learning}.

\subsection{The effect of the fine-tuning strategy}
\label{appendix:fine-tuning}

\begin{figure}[tb]
    \centering
    \includegraphics[width=0.99\textwidth]{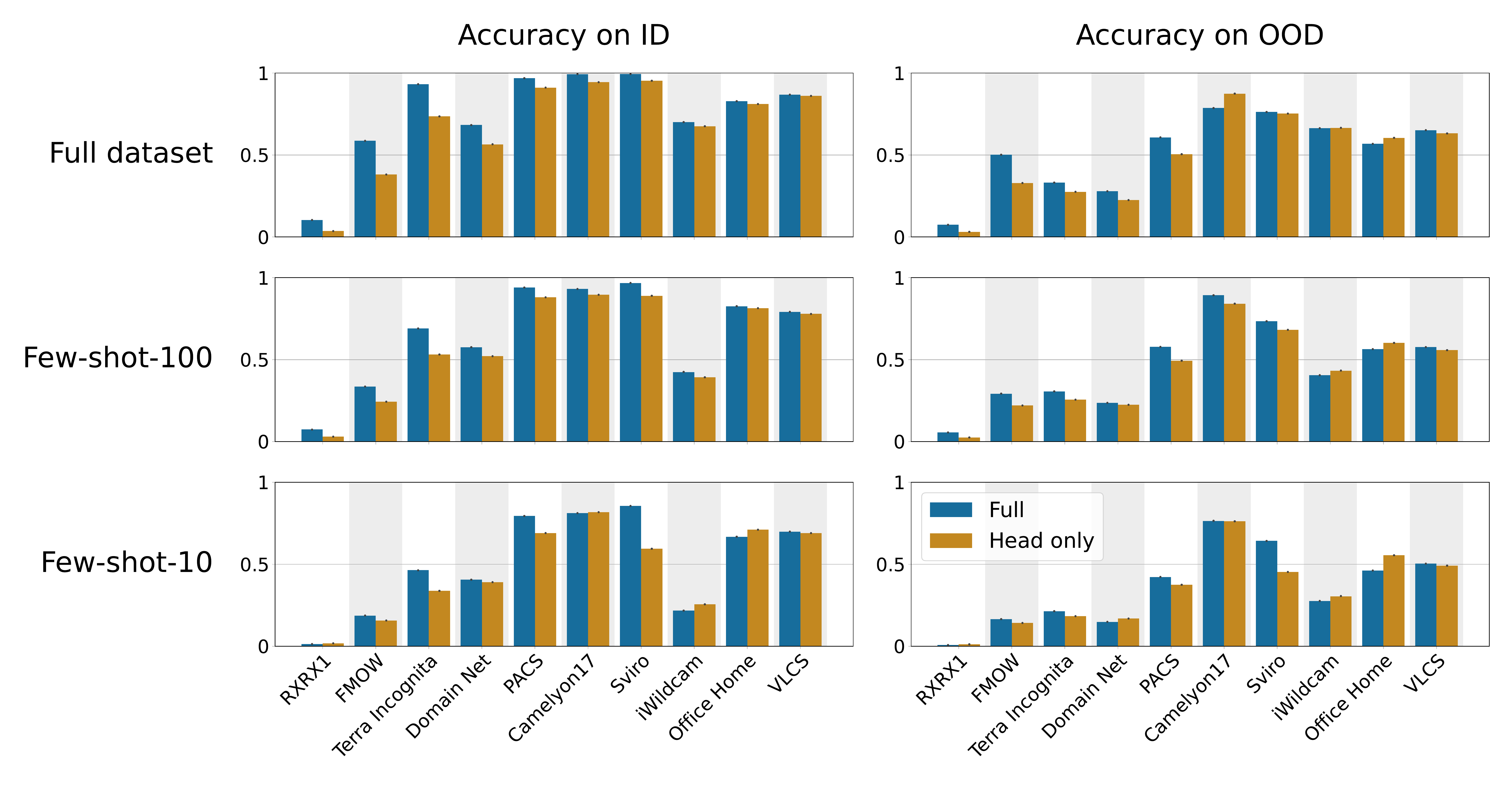}
    \label{fig:finetuning_barplots1}
    \caption{Each bar plot shows the average ID (left) and OOD accuracy (right) obtained by fine-tuning the full architecture or the head only. We compare the results in the setting of training on the full dataset and the few-shot settings (columns). In all plots, the black bars (which are very small and hence barely visible) indicate the standard error.}
\end{figure}

\begin{figure}[tb]
    \centering
    \includegraphics[width=0.99\textwidth]{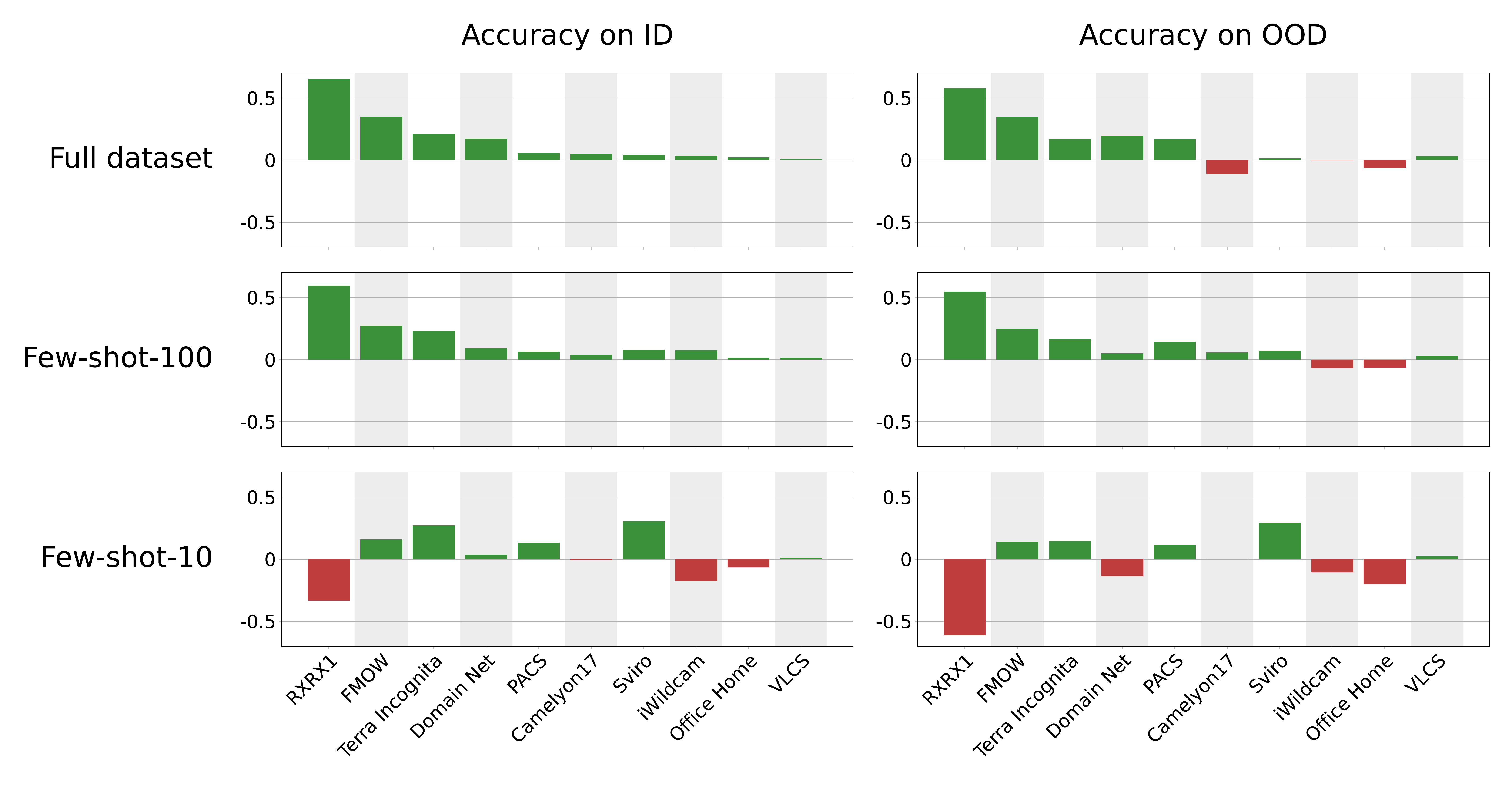}
    \label{fig:finetuning_barplots2}
    \caption{Each bar plot shows the normalized accuracy gap between fine-tuning the full architecture or the head only. We report the average ID (left) and OOD (right) accuracy in setting of training on the full dataset and the few-shot settings (columns). In the low data regime fine-tuning the head only can be beneficial, especially for OOD accuracy.}
\end{figure}

\cref{fig:finetuning_barplots1} shows the average ID and OOD accuracy for the two considered fine-tuning strategies: fine-tuning the full architecture and the linear probe classifier (fine-tuning the head only). To make the difference of both strategies more clear we show the normalized accuracy gap of both fine-tuning approaches in \cref{fig:finetuning_barplots2}. The gap is computed by
\begin{align}
    \textrm{gap} = \frac{\textrm{acc}_\textrm{full} - \textrm{acc}_\textrm{head only}}{\textrm{acc}_\textrm{full}},
\end{align}
where the accuracy terms are averaged over all datasets within a task.
We find that fine-tuning the full architecture is usually superior when using the full fine-tuning dataset. Interestingly, when having access to less data (especially in the few-shot-10 setting), we observe that the linear probe classifier can be better, especially when evaluating on OOD data. This may confirm the idea that changing the last layer only leads to a higher inductive bias by the pre-training data. This is particularly beneficial in low-data regimes when generalizing to OOD data. This insights are in line with previous work (e.g., \cite{lit2021}, Fig. 4).

\subsection{The effect of the augmentation strategy}
\label{appendix:augmentation}

In addition to \cref{fig:effect_of_augmentations} in the main text, we report the average ID and OOD accuracy for each augmentation method for each tasks separately. We do not observe a significant difference between the tasks.

\begin{figure}[tb]
    \centering
    \includegraphics[width=0.99\textwidth]{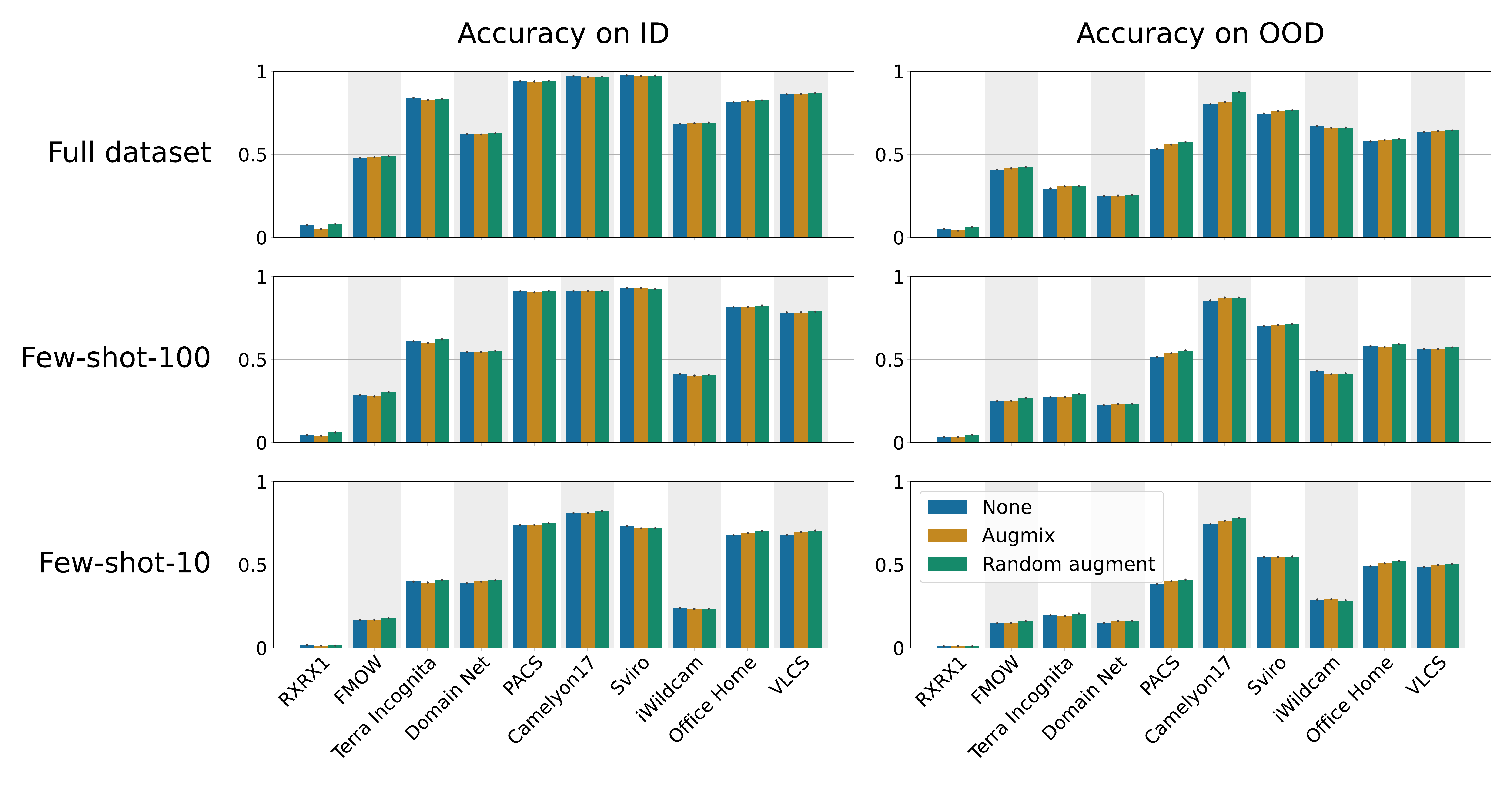}
    \label{fig:augmentation_barplots}
    \caption{Each bar plot shows the average ID (left) and OOD accuracy (right) obtained by using the different augmentation strategies. We compare the results in the setting of training on the full dataset and the few-shot settings (columns). In all plots, the black bars (which are very small and hence barely visible) indicate the standard error.}
\end{figure}

\FloatBarrier
\section{Detailed scatter plots for the relationship of ID vs.\ OOD accuracy}\label{app:scatter_plots}
\Cref{fig:scatter-plots-summary} in the main text shows four prototypical patterns of ID versus OOD accuracy plots, taken from four exemplary dataset pairs out of the $172$ dataset pairs considered in this study.
This section now shows all $172$ ID-vs-OOD-accuracy scatter plots in \cref{fig:scatter-plots}, grouped by task.
Each column corresponds to one adaptation (i.e., fine-tuning) dataset.
Interestingly, the exemplary scatter plot patterns described in \cref{fig:scatter-plots-summary} (functional relationship, vertical line/underspecification, no generalization/ horizontal line, and random generalization/point-cloud) appear to be essentially task related.
For example, in OfficeHome and DomainNet, almost all (ID, OOD) dataset pairs exhibit a clear functional relationship.
In Terra Incognita, generalization never works (horizontal line) or is near-random (unstructured point cloud).
The SVIRO dataset pairs almost systematically fall into the underspecification/vertical line category, and PACS and VLCS exhibit either a relatively clear functional relationships or an underspecification pattern (vertical line).

For better readability, the plots do not include all degrees of freedom of our study.
Specifically, we plot only those networks that were trained on the full adaptation dataset (i.e., we do not include the few-shot settings) and focus on the fine-tuning strategy where all weights of the network get fine-tuned (in contrast to fine-tuning the head only).

\begin{figure}
    \centering
    \captionsetup{belowskip=20pt}
    \includegraphics[width=.5\linewidth]{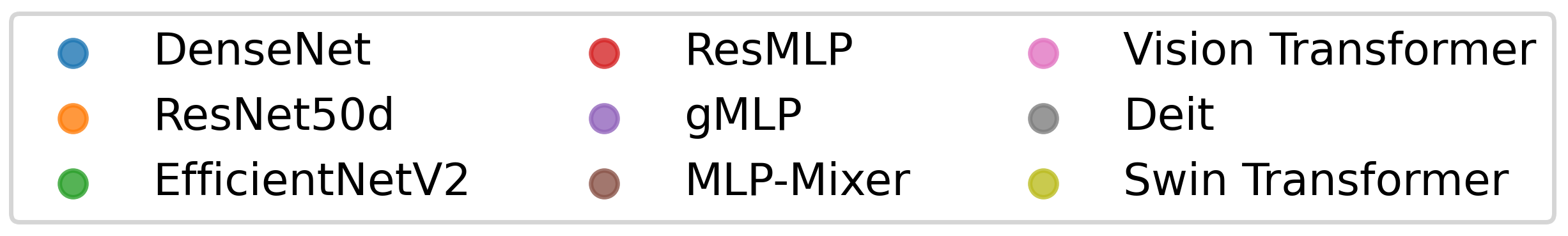}\\
    \vspace{1em}
    \begin{subfigure}[t]{0.49\textwidth}
        \centering
        \includegraphics[width=\linewidth]{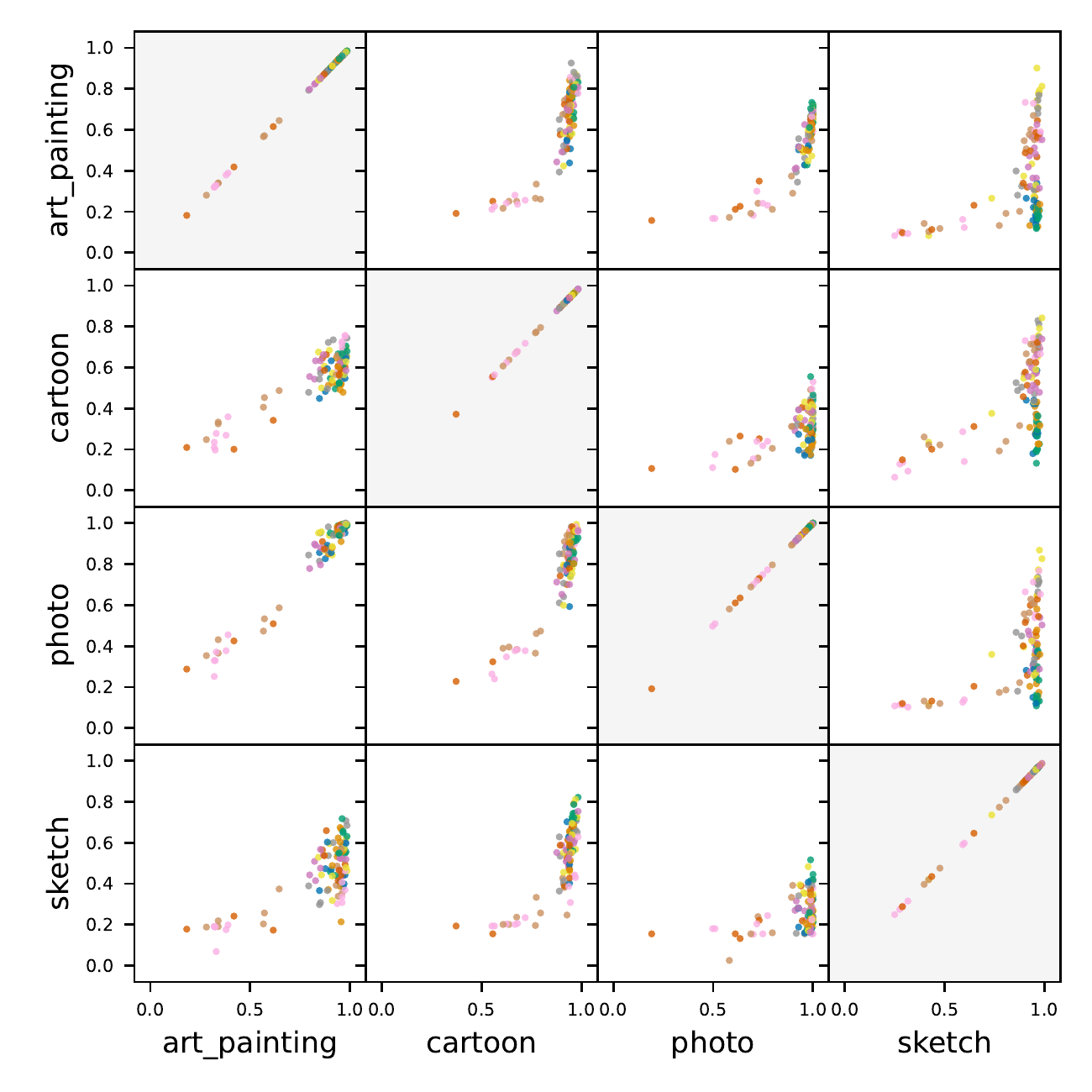}
        \caption{\textbf{PACS}: We observe either a clear functional relationships or an ``underspecified'' regime (vertical line).}
    \end{subfigure}
    \hfill
    \begin{subfigure}[t]{0.49\textwidth}
        \centering
        \includegraphics[width=\linewidth]{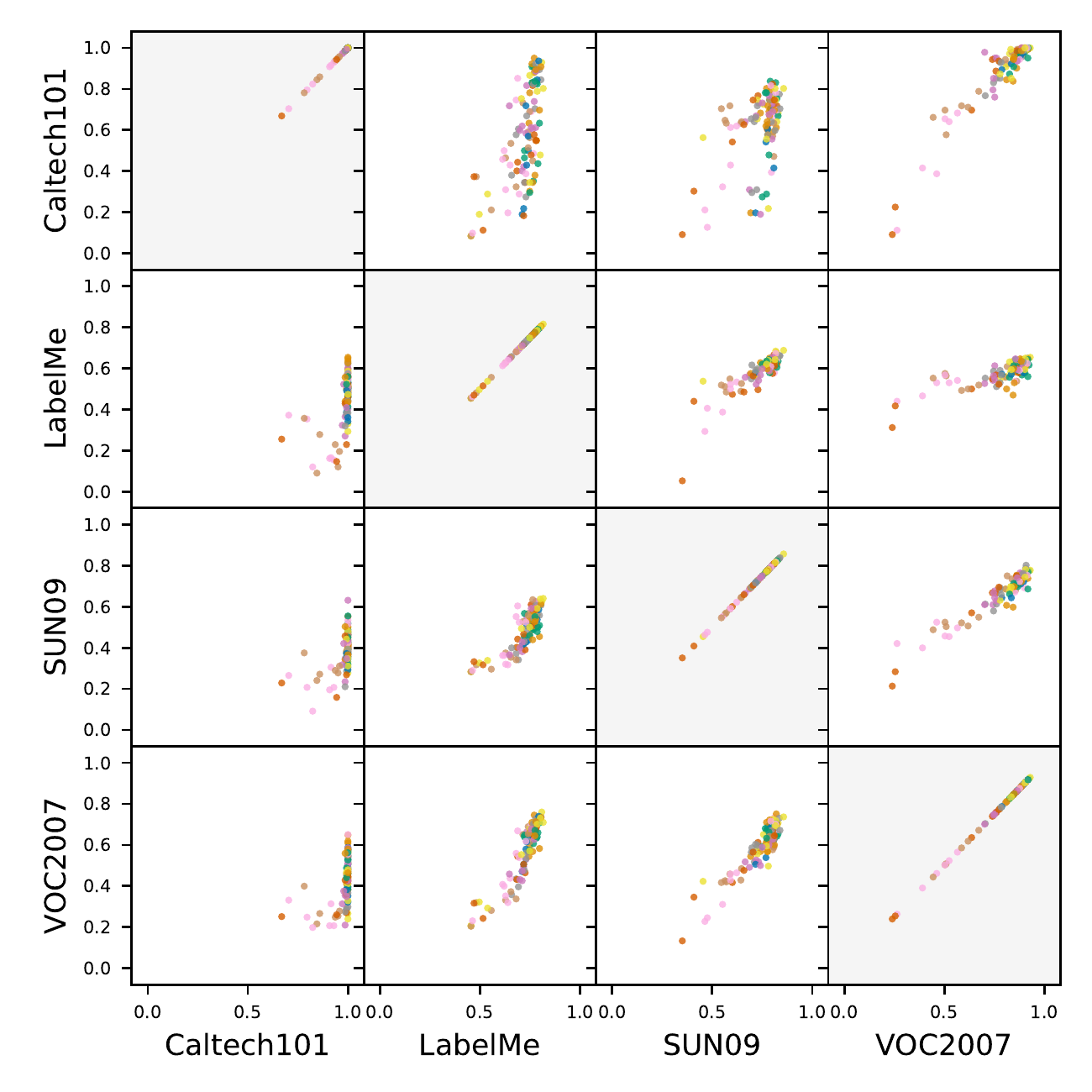}
        \caption{\textbf{VLCS}: We observe similar patterns as for PACS.}
    \end{subfigure}

    \begin{subfigure}[t]{0.49\textwidth}
        \centering
        \includegraphics[width=\linewidth]{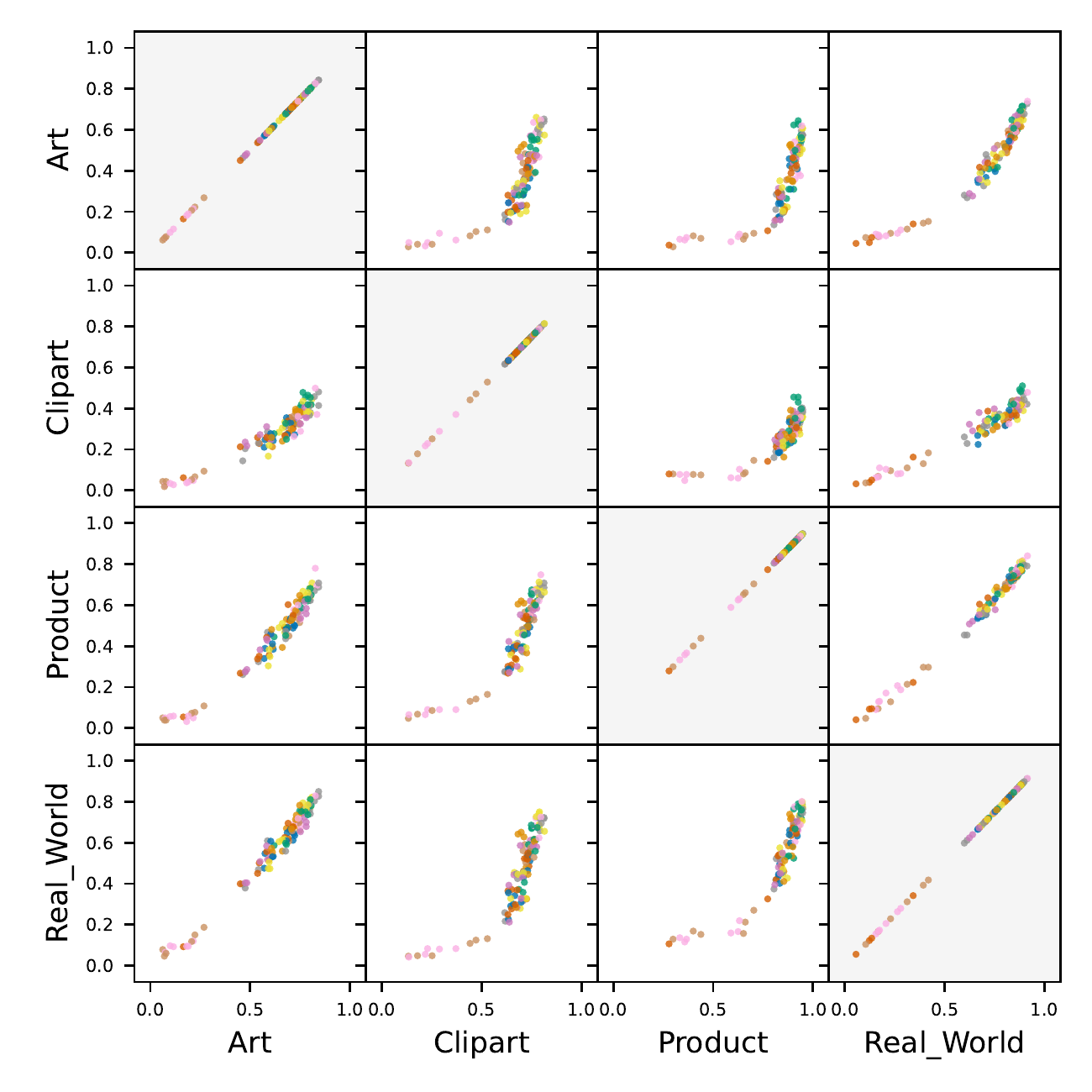}
        \caption{\textbf{Office Home}: For all dataset pairs we observe clear functional relationships.}
    \end{subfigure}
    \hfill
    \begin{subfigure}[t]{0.49\textwidth}
        \centering
        \includegraphics[width=\linewidth]{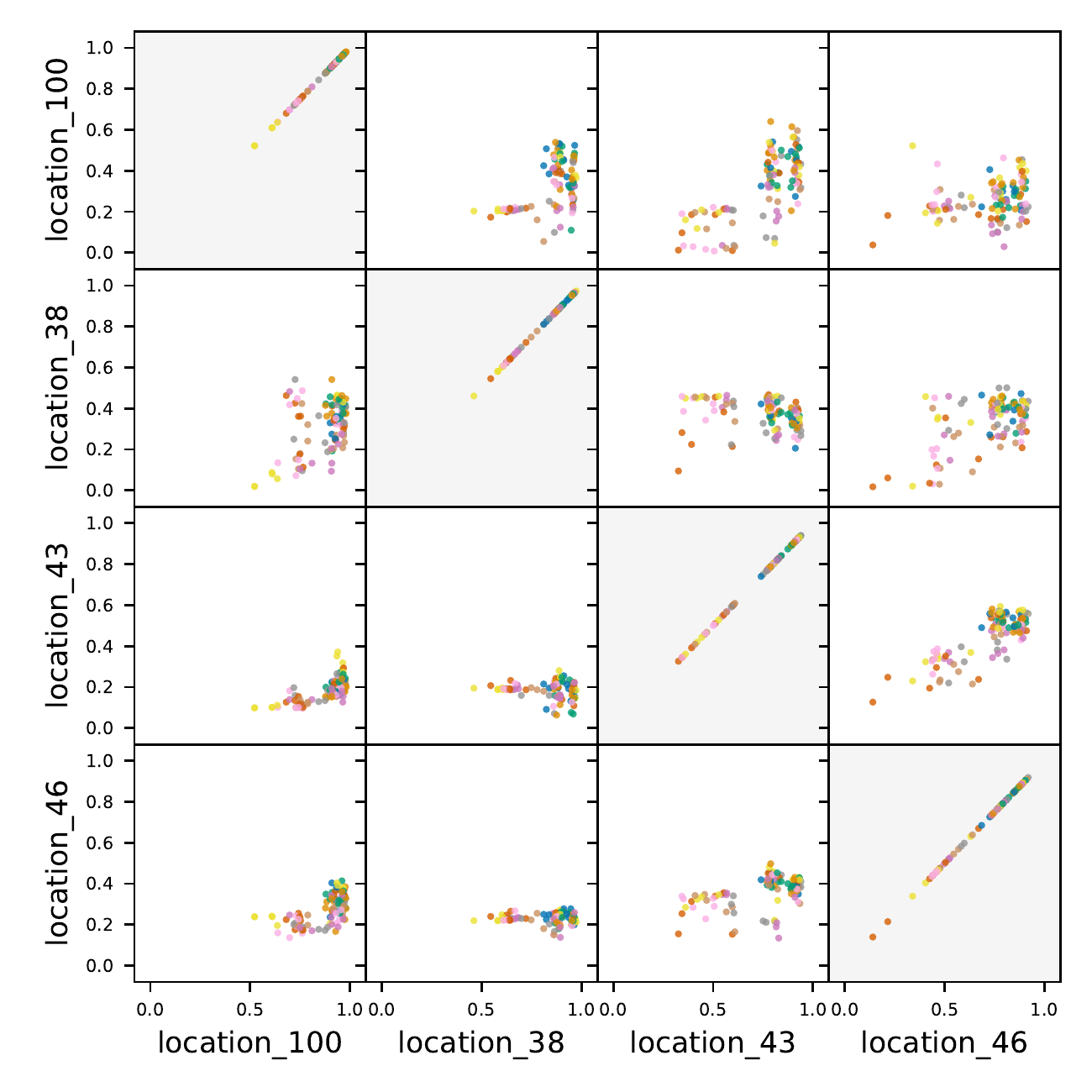}
        \caption{\textbf{Terra Incognita}: Either no transfer (horizontal line) or random associations (scattered point cloud).}
    \end{subfigure}
    \caption{ID versus OOD accuracy for various tasks and dataset pairs.
    Every point represents one fine-tuned network.
    The domains on the x-axis are the ones used for fine-tuning and measuring the ID test accuracy.
    The datasets on the y-axis are the OOD datasets.
    We plot only networks that were trained on the full dataset (no few-shot datasets) and with all their weights (no head-only fine-tuning).
    All points in a same column with the same x-axis value represent the same fine-tuned network.
    Interestingly, the different patterns described in \cref{fig:scatter-plots-summary} (clear functional relationship, underspecification / vertical line, no generalization / horizontal line, random generalization / point-cloud) appear to be essentially task dependent.
    The reader may want to focus on the regions with higher point densities, since the networks outside those denser regions typically correspond to a suboptimal combination of hyperparameters that did not allow to reach convergence.}
    \label{fig:scatter-plots}
\end{figure}
\begin{figure}\ContinuedFloat
    \captionsetup{belowskip=15pt}
    \centering
    \begin{subfigure}{0.95\textwidth}
        \centering
        \includegraphics[width=.5\linewidth]{figures/scatter_plots/legend_only.png}\\
        \vspace{1em}
        \includegraphics[width=.23\linewidth]{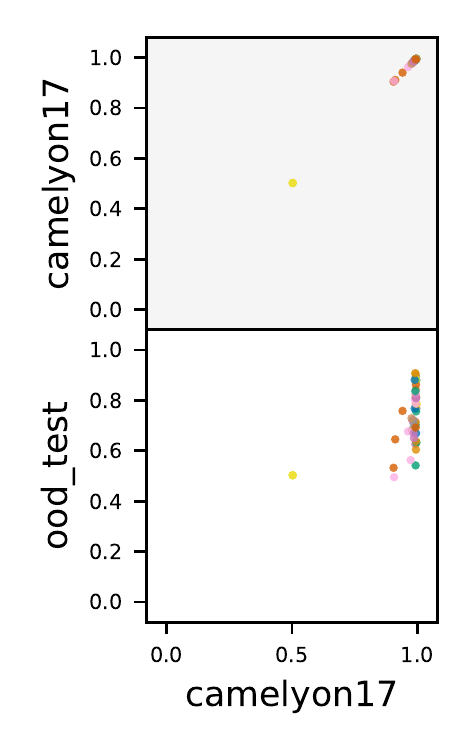}
        \includegraphics[width=.23\linewidth]{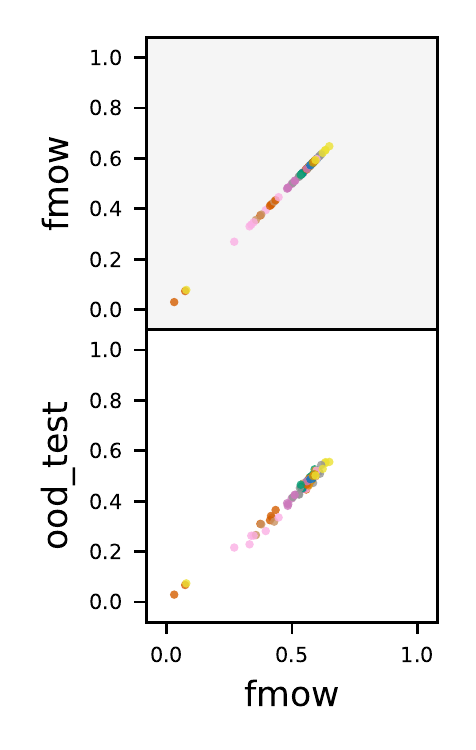}
        \includegraphics[width=.23\linewidth]{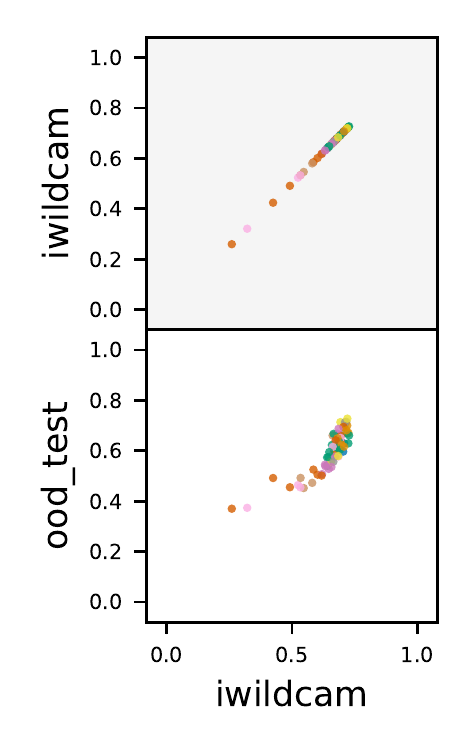}
        \includegraphics[width=.23\linewidth]{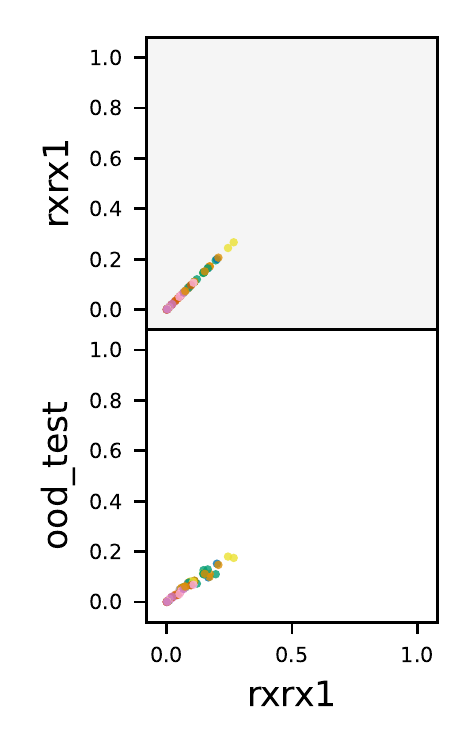}
        \caption{\textbf{WILDS Camelyon17, FMoW, iWildCam, RXRX1}: We observe underspecification (Camelyon17) or clear functional relationships (others).}
    \end{subfigure}

    \begin{subfigure}{0.9\textwidth}
        \centering
        \includegraphics[width=\linewidth]{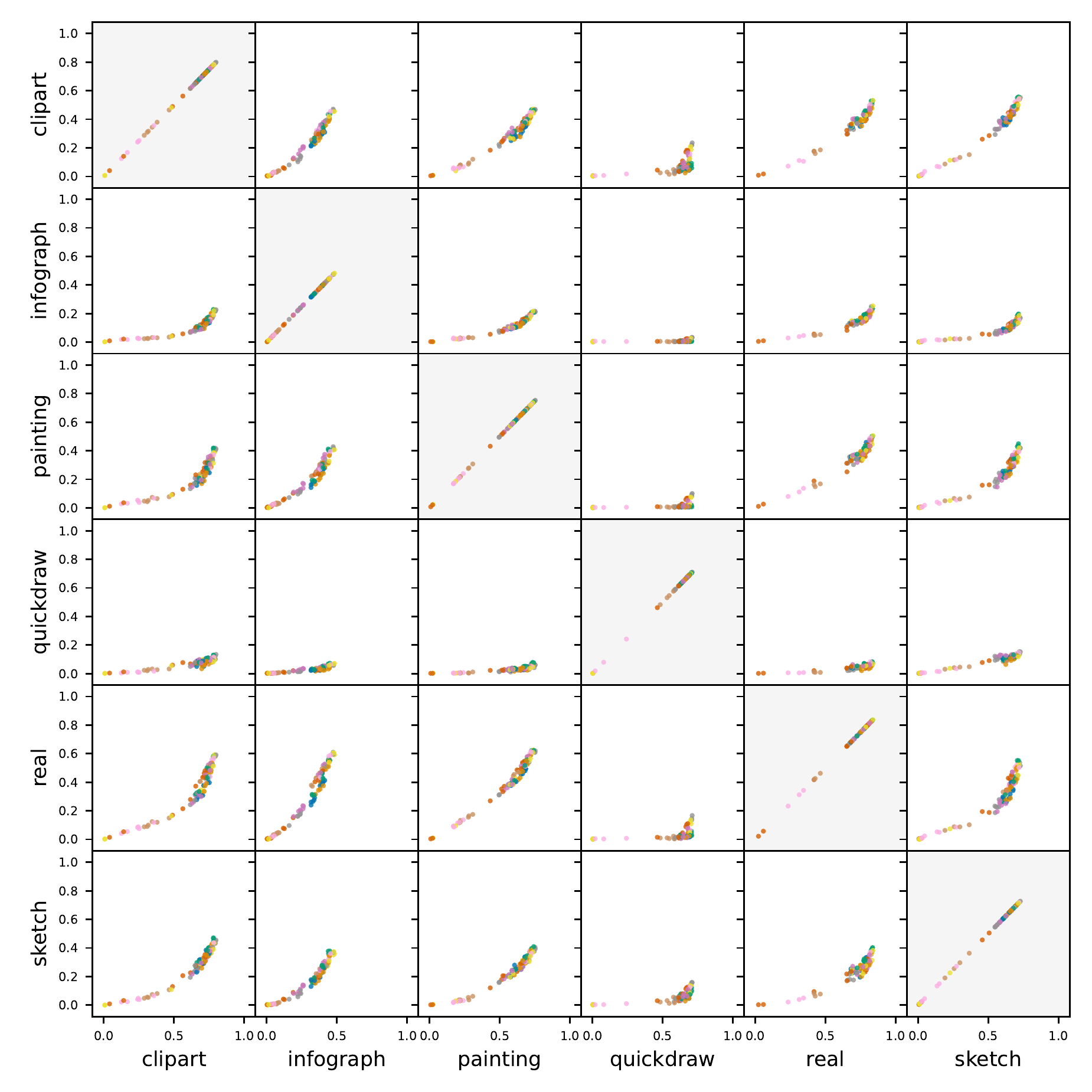}
        \caption{\textbf{Domain Net}: For all dataset pairs we observe clear functional relationships.}
    \end{subfigure}
    \caption{(Continued) ID vs OOD accuracy for various tasks and dataset pairs. For a longer caption, see previous page.}
\end{figure}
\begin{figure}\ContinuedFloat
    \captionsetup{belowskip=20pt}
    \centering
    \begin{subfigure}{\textwidth}
        \centering
        \includegraphics[width=.5\linewidth]{figures/scatter_plots/legend_only.png}\\
        \vspace{1em}
        \includegraphics[width=\linewidth]{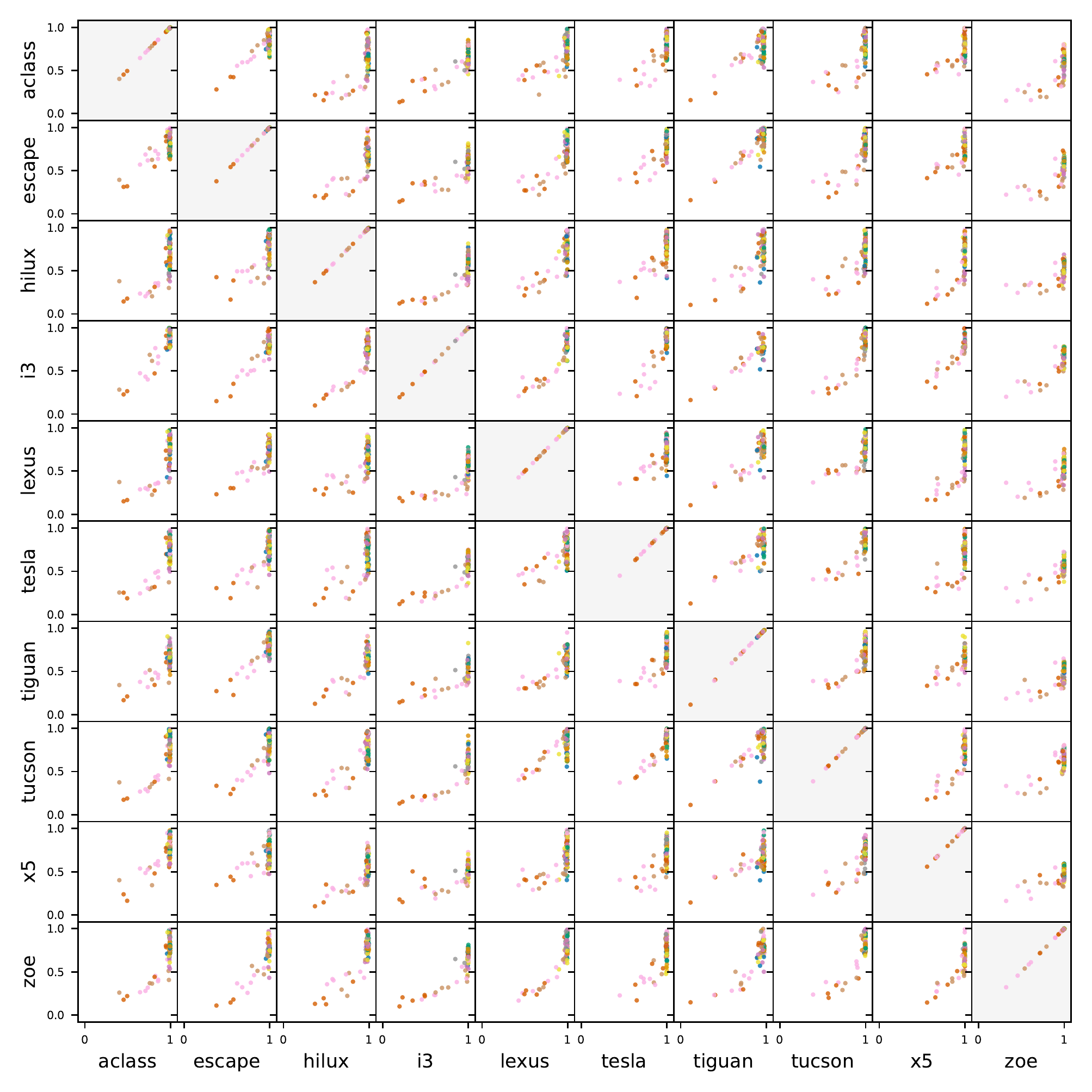}
        \caption{\textbf{SVIRO}: almost always underspecified regime (vertical line).}
    \end{subfigure}
    \caption{(Continued) ID vs OOD accuracy for various tasks and dataset pairs. For a longer caption, see previous page.}
\end{figure}

\FloatBarrier

\section{Details on the tasks and datasets}\label{app:datasets}
We use 36 datasets collected from different benchmarks: DomainNet~\cite{peng2019moment}, PACS~\cite{li2017deeper}, SVIRO~\cite{DiasDaCruz2020SVIRO}, Terra Incognita~\cite{beery2018recognition} as well as the Caltech101~\cite{FeiFei2004LearningGV}, VLCS~\cite{conf/cvpr/TorralbaE11}, Sun09~\cite{choi_cvpr10}, VOC2007~\cite{pascal-voc-2007} and the Wilds datasets~\cite{wilds2021}. We group the domains (datasets) into ten different tasks, see \cref{tab:task_domain} list. For all tasks besides the wilds tasks, we consider all possible (ID, OOD) dataset pairs, i.e., we fine-tune on one dataset in the task (ID dataset) and evaluate the model on all the others (which are considered to be the OOD datasets). For the Wilds datasets we use the predefined ID and OOD splits.

\begin{table}[htb]
    \centering
    \caption{Overview over tasks and their associated domains (datasets), along with a domain type tag.}
    \label{tab:task_domain}
    \begin{tabular}{llc}

    \toprule
    \textbf{Task} & \textbf{Domain} & \textbf{Domain type}\\ \midrule
PACS & art\_painting & Artificial\\
 & cartoon & Artificial\\
 & photo & Real\\
 & sketch & Artificial\\
\midrule
VLCS & Caltech101 & Real\\
 & LabelMe & Real\\
 & SUN09 & Real\\
 & VOC2007 & Real\\
\midrule
domain\_net & clipart & Artificial\\
 & infograph & Artificial\\
 & painting & Artificial\\
 & quickdraw & Artificial\\
 & real & Real\\
 & sketch & Artificial\\
\midrule
office\_home & Art & Artificial\\
 & Clipart & Artificial\\
 & Product & Real\\
 & Real\_World & Real\\
\midrule
sviro & aclass & Artificial\\
 & escape & Artificial\\
 & hilux & Artificial\\
 & i3 & Artificial\\
 & lexus & Artificial\\
 & tesla & Artificial\\
 & tiguan & Artificial\\
 & tucson & Artificial\\
 & x5 & Artificial\\
 & zoe & Artificial\\
\midrule
terra\_incognita & location\_100 & Real\\
 & location\_38 & Real\\
 & location\_43 & Real\\
 & location\_46 & Real\\
\midrule
wilds-camelyon17 & camelyon17-id & Real\\
 & camelyon17-ood & Real\\
\midrule
wilds-fmow & fmow-id & Real\\
 & fmow-ood & Real\\
\midrule
wilds-iwildcam & iwildcam-id & Real\\
 & iwildcam-ood & Real\\
\midrule
wilds-rxrx1 & rxrx1-id & Real\\
 & rxrx1-ood & Real\\
    \bottomrule
    \end{tabular}

\end{table}

\subsection{Comparison of the task difficulty}
\label{app:task_difficulty}
\begin{figure}
    \centering
    \includegraphics[width=0.8\textwidth]{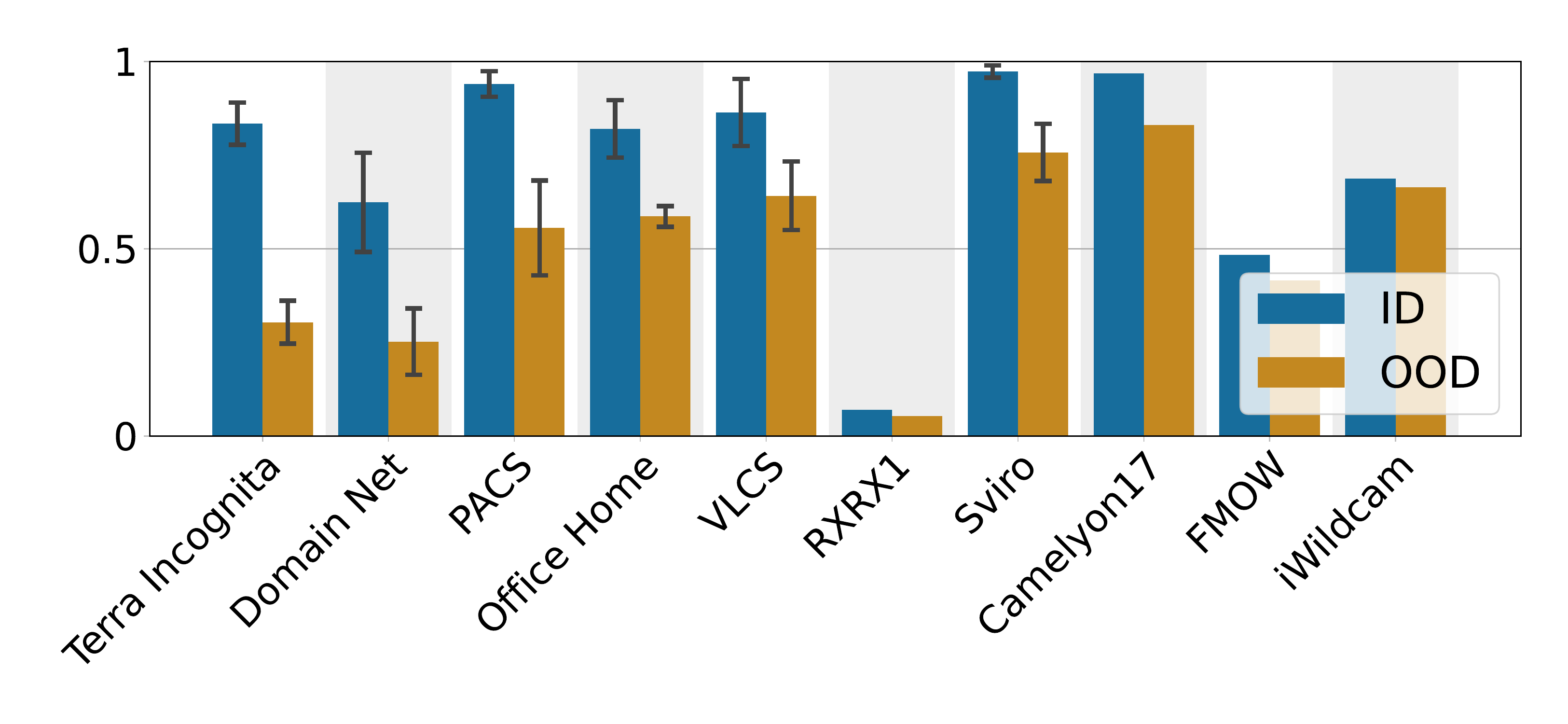}
    \includegraphics[width=0.8\textwidth]{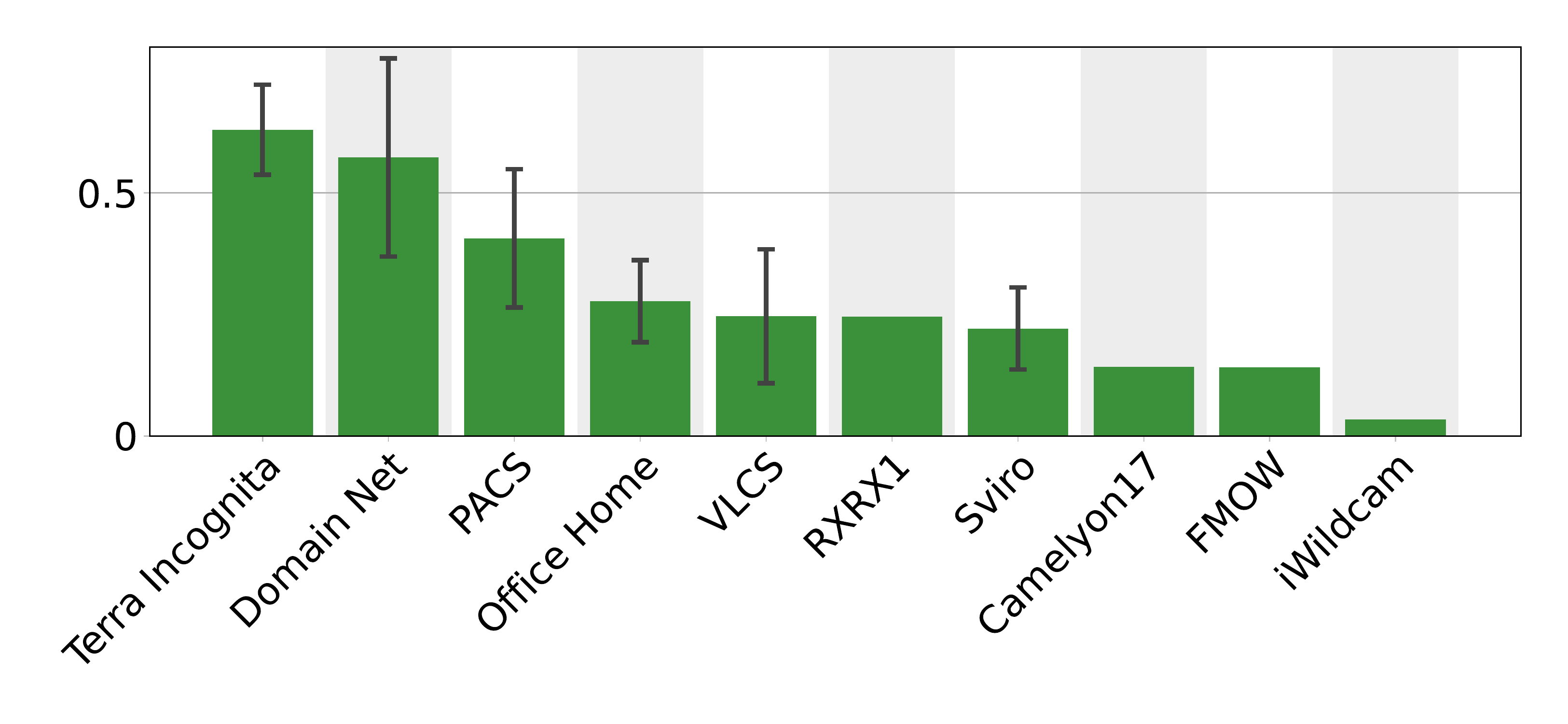}
    \caption{\textbf{\textsc{top:}} Mean accuracy on ID data and OOD data for each task (averaged over all datasets in the task, architectures, augmentation and fine-tuning strategies, c.f. \cref{tab:task_domain}). \textbf{\textsc{bottom:}} Normalized mean ID vs. OOD accuracy gap for each task, see \cref{eq:normalized_id_ood_gap}. This indicates the average difficulty of the domain generalization. In all plots, the black bars indicate the standard deviation (not the standard error as in the other plots) across the datasets within the task if the tasks contains multiple dataset pairs. }
    \label{fig:task_difficulty}
\end{figure}

In the following we discuss the difficulty of the different tasks. In our context, the difficulty of a task might be defined by how hard it is to transfer from the ID dataset to the OOD dataset. In other words, a task is more difficult if the distribution of the OOD data is more distant from the distribution of the ID data. It is not obvious how to measure this distance and multiple approaches have been proposed~\cite[e.g.,][]{10.1007/s10994-009-5152-4, zhang2021quantifying}. Here we choose a more direct measure and compare the ID and OOD performance for each task averaged over all dataset pairs in the task, and all models and training hyperparameters.
\cref{fig:task_difficulty} shows the average accuracy obtained on all datasets within each task. In the top plot we compare the average ID accuracy and OOD accuracy. In the bottom plot we display the normalized ID vs. OOD accuracy gap computed by
\begin{align}
\label{eq:normalized_id_ood_gap}
    \textrm{gap} = \frac{\textrm{acc}_\textrm{ID} - \textrm{acc}_\textrm{OOD}}{\textrm{acc}_\textrm{ID}},
\end{align}
where the accuracy terms are averaged over all datasets within a task. The average ID vs. OOD accuracy gap can serve as a proxy measure of the difficulty of a task. It indicates how hard the OOD prediction task is in average for a model that was trained on the ID data.

\begin{figure}
    \centering
    \includegraphics[width=0.6\textwidth]{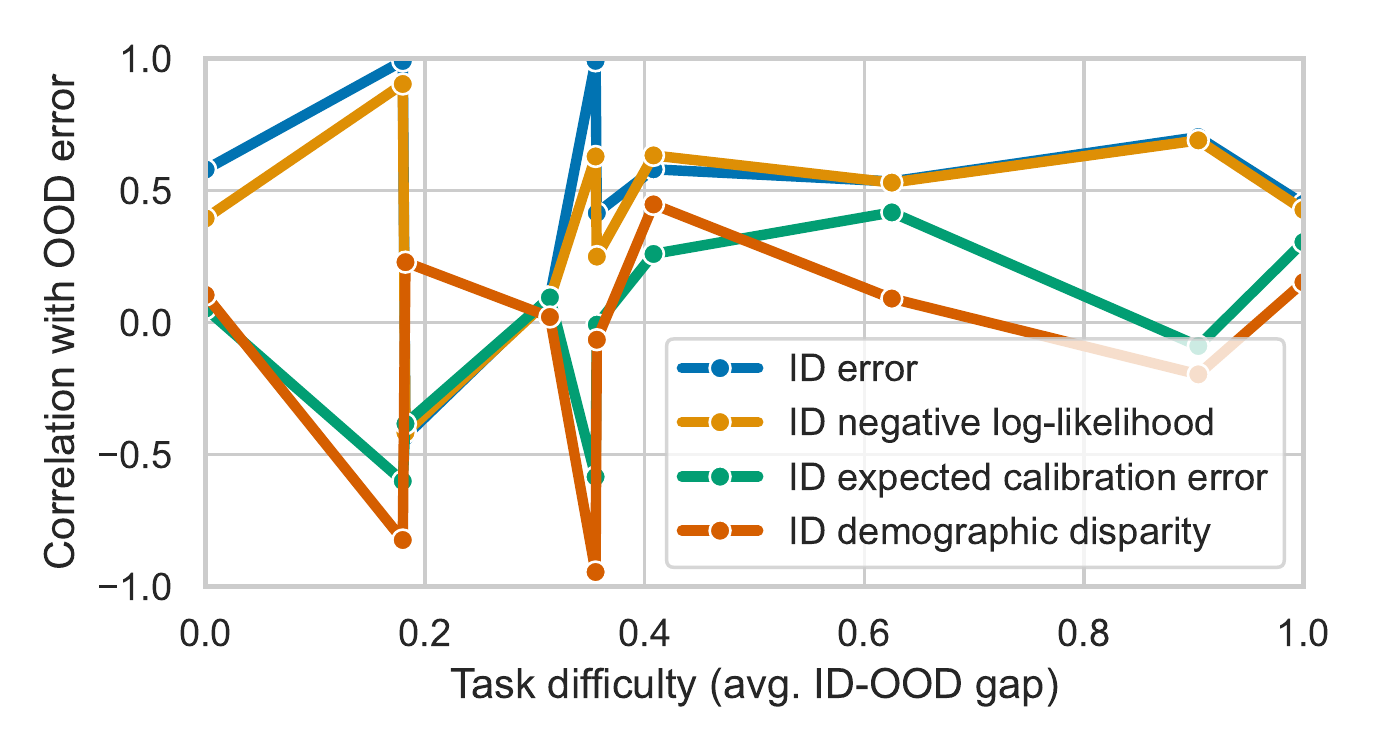}
    \caption{Each line shows the correlation coefficients between an ID metric and the OOD prediction error as a function of task difficulty. The task difficulty was computed by the normalized ID vs. OOD gap as shown in \cref{fig:task_difficulty} and was linearly mapped to values between 0 and 1. No clear pattern is visible and the correlation coefficients seem to be rather independent of the task difficulty.}
    \label{fig:task_difficulty_correlation_coeff}
\end{figure}

Next we discuss how the predictiveness of the different metrics depend on the task difficulty. \cref{fig:task_difficulty_correlation_coeff} shows the correlation coefficients (between the ID metrics and OOD prediction error) as a function of task difficulty. We observe that ID error is the best predictor of the ID error for most difficulty levels. Besides this observation we find that the results are noisy and no clear pattern can be observed. Surprisingly, the correlation coefficients seem to be rather independent of the task difficulty.

\FloatBarrier

\section{Detailed comparison of model performances}\label{app:model_comparison}
We compare the in-distribution (ID) and out-of-distribution (OOD) performance for each model on all tasks.
Here we follow the scenario where the hyperparameters are selected on an held-out ID validation split (this mimics the usual hyperparameter selection approach used in practice).
We then compute the ID performance on the ID test set and the mean OOD performance on all other (OOD) domains of a task.
Finally, we report the mean ID and OOD accuracy within each task and the gap between ID and OOD performance in \cref{tab:model_comparison}.
The numbers in gray represent the standard error of these means.
This table represents a more details view on the performance of each individual model compared to \cref{tab:id_ood_gap} in the main text, where the performance is averaged over tasks.

{
    \scriptsize
    \begin{longtable}{llccc}
\caption{Average ID and OOD accuracy of each model on each task over all domains in the task.} \label{tab:model_comparison} \\
\toprule
        Model &           Domain &                            ID Error &                           OOD Error &                           OOD-ID Gap \\
\midrule
\endfirsthead

\toprule
        Model &           Domain &                            ID Error &                           OOD Error &                           OOD-ID Gap \\
\midrule
\endhead
\midrule
\multicolumn{5}{r}{{Continued on next page}} \\
\midrule
\endfoot

\bottomrule
\endlastfoot
\midrule
         Deit &             PACS & 0.033 \textcolor{gray}{$\pm$ 0.003} & 0.333 \textcolor{gray}{$\pm$ 0.022} &  0.299 \textcolor{gray}{$\pm$ 0.023} \\
              &             VLCS & 0.118 \textcolor{gray}{$\pm$ 0.010} & 0.328 \textcolor{gray}{$\pm$ 0.017} &  0.210 \textcolor{gray}{$\pm$ 0.022} \\
              &      domain\_net & 0.316 \textcolor{gray}{$\pm$ 0.009} & 0.703 \textcolor{gray}{$\pm$ 0.014} &  0.387 \textcolor{gray}{$\pm$ 0.017} \\
              &     office\_home & 0.131 \textcolor{gray}{$\pm$ 0.008} & 0.341 \textcolor{gray}{$\pm$ 0.017} &  0.209 \textcolor{gray}{$\pm$ 0.020} \\
              &            sviro & 0.014 \textcolor{gray}{$\pm$ 0.001} & 0.219 \textcolor{gray}{$\pm$ 0.005} &  0.205 \textcolor{gray}{$\pm$ 0.005} \\
              & terra\_incognita & 0.127 \textcolor{gray}{$\pm$ 0.011} & 0.661 \textcolor{gray}{$\pm$ 0.013} &  0.533 \textcolor{gray}{$\pm$ 0.020} \\
              & wilds-camelyon17 & 0.022 \textcolor{gray}{$\pm$ 0.007} & 0.156 \textcolor{gray}{$\pm$ 0.036} &  0.134 \textcolor{gray}{$\pm$ 0.040} \\
              &       wilds-fmow & 0.468 \textcolor{gray}{$\pm$ 0.036} & 0.543 \textcolor{gray}{$\pm$ 0.030} &  0.074 \textcolor{gray}{$\pm$ 0.007} \\
              &   wilds-iwildcam & 0.275 \textcolor{gray}{$\pm$ 0.007} & 0.289 \textcolor{gray}{$\pm$ 0.009} &  0.014 \textcolor{gray}{$\pm$ 0.006} \\
              &      wilds-rxrx1 & 0.925 \textcolor{gray}{$\pm$ 0.005} & 0.939 \textcolor{gray}{$\pm$ 0.004} &  0.014 \textcolor{gray}{$\pm$ 0.002} \\
\midrule
  DenseNet169 &             PACS & 0.061 \textcolor{gray}{$\pm$ 0.006} & 0.523 \textcolor{gray}{$\pm$ 0.031} &  0.463 \textcolor{gray}{$\pm$ 0.031} \\
              &             VLCS & 0.137 \textcolor{gray}{$\pm$ 0.011} & 0.365 \textcolor{gray}{$\pm$ 0.018} &  0.228 \textcolor{gray}{$\pm$ 0.024} \\
              &      domain\_net & 0.387 \textcolor{gray}{$\pm$ 0.011} & 0.788 \textcolor{gray}{$\pm$ 0.012} &  0.401 \textcolor{gray}{$\pm$ 0.015} \\
              &     office\_home & 0.209 \textcolor{gray}{$\pm$ 0.010} & 0.496 \textcolor{gray}{$\pm$ 0.017} &  0.287 \textcolor{gray}{$\pm$ 0.019} \\
              &            sviro & 0.047 \textcolor{gray}{$\pm$ 0.003} & 0.274 \textcolor{gray}{$\pm$ 0.005} &  0.227 \textcolor{gray}{$\pm$ 0.006} \\
              & terra\_incognita & 0.186 \textcolor{gray}{$\pm$ 0.015} & 0.725 \textcolor{gray}{$\pm$ 0.015} &  0.539 \textcolor{gray}{$\pm$ 0.019} \\
              & wilds-camelyon17 & 0.036 \textcolor{gray}{$\pm$ 0.012} & 0.169 \textcolor{gray}{$\pm$ 0.038} &  0.133 \textcolor{gray}{$\pm$ 0.047} \\
              &       wilds-fmow & 0.526 \textcolor{gray}{$\pm$ 0.046} & 0.593 \textcolor{gray}{$\pm$ 0.037} &  0.067 \textcolor{gray}{$\pm$ 0.009} \\
              &   wilds-iwildcam & 0.319 \textcolor{gray}{$\pm$ 0.010} & 0.364 \textcolor{gray}{$\pm$ 0.012} &  0.046 \textcolor{gray}{$\pm$ 0.021} \\
              &      wilds-rxrx1 & 0.911 \textcolor{gray}{$\pm$ 0.039} & 0.937 \textcolor{gray}{$\pm$ 0.027} &  0.026 \textcolor{gray}{$\pm$ 0.014} \\
\midrule
EfficientNet2 &             PACS & 0.053 \textcolor{gray}{$\pm$ 0.005} & 0.469 \textcolor{gray}{$\pm$ 0.032} &  0.416 \textcolor{gray}{$\pm$ 0.032} \\
              &             VLCS & 0.140 \textcolor{gray}{$\pm$ 0.012} & 0.374 \textcolor{gray}{$\pm$ 0.018} &  0.234 \textcolor{gray}{$\pm$ 0.023} \\
              &      domain\_net & 0.369 \textcolor{gray}{$\pm$ 0.012} & 0.734 \textcolor{gray}{$\pm$ 0.014} &  0.365 \textcolor{gray}{$\pm$ 0.017} \\
              &     office\_home & 0.172 \textcolor{gray}{$\pm$ 0.009} & 0.403 \textcolor{gray}{$\pm$ 0.015} &  0.230 \textcolor{gray}{$\pm$ 0.017} \\
              &            sviro & 0.031 \textcolor{gray}{$\pm$ 0.002} & 0.249 \textcolor{gray}{$\pm$ 0.005} &  0.218 \textcolor{gray}{$\pm$ 0.005} \\
              & terra\_incognita & 0.187 \textcolor{gray}{$\pm$ 0.016} & 0.703 \textcolor{gray}{$\pm$ 0.013} &  0.516 \textcolor{gray}{$\pm$ 0.017} \\
              & wilds-camelyon17 & 0.039 \textcolor{gray}{$\pm$ 0.015} & 0.194 \textcolor{gray}{$\pm$ 0.039} &  0.155 \textcolor{gray}{$\pm$ 0.049} \\
              &       wilds-fmow & 0.530 \textcolor{gray}{$\pm$ 0.053} & 0.588 \textcolor{gray}{$\pm$ 0.047} &  0.059 \textcolor{gray}{$\pm$ 0.006} \\
              &   wilds-iwildcam & 0.325 \textcolor{gray}{$\pm$ 0.022} & 0.383 \textcolor{gray}{$\pm$ 0.017} &  0.058 \textcolor{gray}{$\pm$ 0.009} \\
              &      wilds-rxrx1 & 0.901 \textcolor{gray}{$\pm$ 0.032} & 0.927 \textcolor{gray}{$\pm$ 0.022} &  0.025 \textcolor{gray}{$\pm$ 0.013} \\
\midrule
         GMLP &             PACS & 0.070 \textcolor{gray}{$\pm$ 0.007} & 0.434 \textcolor{gray}{$\pm$ 0.026} &  0.363 \textcolor{gray}{$\pm$ 0.027} \\
              &             VLCS & 0.136 \textcolor{gray}{$\pm$ 0.011} & 0.361 \textcolor{gray}{$\pm$ 0.016} &  0.225 \textcolor{gray}{$\pm$ 0.022} \\
              &      domain\_net & 0.404 \textcolor{gray}{$\pm$ 0.012} & 0.753 \textcolor{gray}{$\pm$ 0.013} &  0.348 \textcolor{gray}{$\pm$ 0.017} \\
              &     office\_home & 0.195 \textcolor{gray}{$\pm$ 0.009} & 0.409 \textcolor{gray}{$\pm$ 0.016} &  0.214 \textcolor{gray}{$\pm$ 0.020} \\
              &            sviro & 0.034 \textcolor{gray}{$\pm$ 0.002} & 0.260 \textcolor{gray}{$\pm$ 0.006} &  0.226 \textcolor{gray}{$\pm$ 0.006} \\
              & terra\_incognita & 0.180 \textcolor{gray}{$\pm$ 0.016} & 0.708 \textcolor{gray}{$\pm$ 0.010} &  0.528 \textcolor{gray}{$\pm$ 0.021} \\
              & wilds-camelyon17 & 0.041 \textcolor{gray}{$\pm$ 0.016} & 0.214 \textcolor{gray}{$\pm$ 0.032} &  0.173 \textcolor{gray}{$\pm$ 0.042} \\
              &       wilds-fmow & 0.556 \textcolor{gray}{$\pm$ 0.056} & 0.624 \textcolor{gray}{$\pm$ 0.046} &  0.069 \textcolor{gray}{$\pm$ 0.011} \\
              &   wilds-iwildcam & 0.338 \textcolor{gray}{$\pm$ 0.009} & 0.329 \textcolor{gray}{$\pm$ 0.006} & -0.008 \textcolor{gray}{$\pm$ 0.011} \\
              &      wilds-rxrx1 & 0.972 \textcolor{gray}{$\pm$ 0.008} & 0.977 \textcolor{gray}{$\pm$ 0.006} &  0.005 \textcolor{gray}{$\pm$ 0.002} \\
\midrule
        Mixer &             PACS & 0.091 \textcolor{gray}{$\pm$ 0.009} & 0.469 \textcolor{gray}{$\pm$ 0.028} &  0.378 \textcolor{gray}{$\pm$ 0.029} \\
              &             VLCS & 0.150 \textcolor{gray}{$\pm$ 0.011} & 0.386 \textcolor{gray}{$\pm$ 0.017} &  0.236 \textcolor{gray}{$\pm$ 0.022} \\
              &      domain\_net & 0.433 \textcolor{gray}{$\pm$ 0.014} & 0.785 \textcolor{gray}{$\pm$ 0.012} &  0.352 \textcolor{gray}{$\pm$ 0.016} \\
              &     office\_home & 0.215 \textcolor{gray}{$\pm$ 0.011} & 0.472 \textcolor{gray}{$\pm$ 0.016} &  0.257 \textcolor{gray}{$\pm$ 0.020} \\
              &            sviro & 0.023 \textcolor{gray}{$\pm$ 0.001} & 0.251 \textcolor{gray}{$\pm$ 0.005} &  0.228 \textcolor{gray}{$\pm$ 0.005} \\
              & terra\_incognita & 0.172 \textcolor{gray}{$\pm$ 0.015} & 0.737 \textcolor{gray}{$\pm$ 0.013} &  0.565 \textcolor{gray}{$\pm$ 0.018} \\
              & wilds-camelyon17 & 0.025 \textcolor{gray}{$\pm$ 0.008} & 0.154 \textcolor{gray}{$\pm$ 0.029} &  0.129 \textcolor{gray}{$\pm$ 0.036} \\
              &       wilds-fmow & 0.534 \textcolor{gray}{$\pm$ 0.042} & 0.608 \textcolor{gray}{$\pm$ 0.038} &  0.074 \textcolor{gray}{$\pm$ 0.005} \\
              &   wilds-iwildcam & 0.333 \textcolor{gray}{$\pm$ 0.005} & 0.374 \textcolor{gray}{$\pm$ 0.009} &  0.041 \textcolor{gray}{$\pm$ 0.012} \\
              &      wilds-rxrx1 & 0.983 \textcolor{gray}{$\pm$ 0.006} & 0.985 \textcolor{gray}{$\pm$ 0.005} &  0.002 \textcolor{gray}{$\pm$ 0.001} \\
\midrule
       ResMLP &             PACS & 0.066 \textcolor{gray}{$\pm$ 0.006} & 0.465 \textcolor{gray}{$\pm$ 0.026} &  0.398 \textcolor{gray}{$\pm$ 0.026} \\
              &             VLCS & 0.145 \textcolor{gray}{$\pm$ 0.011} & 0.395 \textcolor{gray}{$\pm$ 0.017} &  0.250 \textcolor{gray}{$\pm$ 0.022} \\
              &      domain\_net & 0.398 \textcolor{gray}{$\pm$ 0.011} & 0.764 \textcolor{gray}{$\pm$ 0.012} &  0.367 \textcolor{gray}{$\pm$ 0.016} \\
              &     office\_home & 0.203 \textcolor{gray}{$\pm$ 0.010} & 0.453 \textcolor{gray}{$\pm$ 0.018} &  0.250 \textcolor{gray}{$\pm$ 0.021} \\
              &            sviro & 0.027 \textcolor{gray}{$\pm$ 0.001} & 0.245 \textcolor{gray}{$\pm$ 0.005} &  0.218 \textcolor{gray}{$\pm$ 0.005} \\
              & terra\_incognita & 0.164 \textcolor{gray}{$\pm$ 0.014} & 0.694 \textcolor{gray}{$\pm$ 0.013} &  0.530 \textcolor{gray}{$\pm$ 0.020} \\
              & wilds-camelyon17 & 0.032 \textcolor{gray}{$\pm$ 0.012} & 0.157 \textcolor{gray}{$\pm$ 0.015} &  0.125 \textcolor{gray}{$\pm$ 0.025} \\
              &       wilds-fmow & 0.544 \textcolor{gray}{$\pm$ 0.047} & 0.613 \textcolor{gray}{$\pm$ 0.039} &  0.069 \textcolor{gray}{$\pm$ 0.009} \\
              &   wilds-iwildcam & 0.317 \textcolor{gray}{$\pm$ 0.006} & 0.343 \textcolor{gray}{$\pm$ 0.009} &  0.026 \textcolor{gray}{$\pm$ 0.008} \\
              &      wilds-rxrx1 & 0.949 \textcolor{gray}{$\pm$ 0.010} & 0.959 \textcolor{gray}{$\pm$ 0.005} &  0.010 \textcolor{gray}{$\pm$ 0.005} \\
\midrule
     ResNet50 &             PACS & 0.054 \textcolor{gray}{$\pm$ 0.005} & 0.499 \textcolor{gray}{$\pm$ 0.032} &  0.445 \textcolor{gray}{$\pm$ 0.033} \\
              &             VLCS & 0.137 \textcolor{gray}{$\pm$ 0.011} & 0.325 \textcolor{gray}{$\pm$ 0.017} &  0.189 \textcolor{gray}{$\pm$ 0.022} \\
              &      domain\_net & 0.359 \textcolor{gray}{$\pm$ 0.011} & 0.749 \textcolor{gray}{$\pm$ 0.014} &  0.390 \textcolor{gray}{$\pm$ 0.016} \\
              &     office\_home & 0.180 \textcolor{gray}{$\pm$ 0.010} & 0.431 \textcolor{gray}{$\pm$ 0.017} &  0.251 \textcolor{gray}{$\pm$ 0.020} \\
              &            sviro & 0.025 \textcolor{gray}{$\pm$ 0.001} & 0.244 \textcolor{gray}{$\pm$ 0.005} &  0.220 \textcolor{gray}{$\pm$ 0.006} \\
              & terra\_incognita & 0.176 \textcolor{gray}{$\pm$ 0.014} & 0.698 \textcolor{gray}{$\pm$ 0.015} &  0.521 \textcolor{gray}{$\pm$ 0.019} \\
              & wilds-camelyon17 & 0.036 \textcolor{gray}{$\pm$ 0.013} & 0.195 \textcolor{gray}{$\pm$ 0.045} &  0.159 \textcolor{gray}{$\pm$ 0.052} \\
              &       wilds-fmow & 0.512 \textcolor{gray}{$\pm$ 0.045} & 0.579 \textcolor{gray}{$\pm$ 0.036} &  0.067 \textcolor{gray}{$\pm$ 0.010} \\
              &   wilds-iwildcam & 0.313 \textcolor{gray}{$\pm$ 0.015} & 0.328 \textcolor{gray}{$\pm$ 0.007} &  0.015 \textcolor{gray}{$\pm$ 0.010} \\
              &      wilds-rxrx1 & 0.894 \textcolor{gray}{$\pm$ 0.032} & 0.925 \textcolor{gray}{$\pm$ 0.021} &  0.030 \textcolor{gray}{$\pm$ 0.012} \\
\midrule
         Swin &             PACS & 0.048 \textcolor{gray}{$\pm$ 0.006} & 0.393 \textcolor{gray}{$\pm$ 0.027} &  0.345 \textcolor{gray}{$\pm$ 0.027} \\
              &             VLCS & 0.121 \textcolor{gray}{$\pm$ 0.010} & 0.350 \textcolor{gray}{$\pm$ 0.017} &  0.230 \textcolor{gray}{$\pm$ 0.021} \\
              &      domain\_net & 0.340 \textcolor{gray}{$\pm$ 0.010} & 0.718 \textcolor{gray}{$\pm$ 0.014} &  0.378 \textcolor{gray}{$\pm$ 0.017} \\
              &     office\_home & 0.152 \textcolor{gray}{$\pm$ 0.009} & 0.351 \textcolor{gray}{$\pm$ 0.018} &  0.199 \textcolor{gray}{$\pm$ 0.022} \\
              &            sviro & 0.019 \textcolor{gray}{$\pm$ 0.001} & 0.218 \textcolor{gray}{$\pm$ 0.005} &  0.199 \textcolor{gray}{$\pm$ 0.005} \\
              & terra\_incognita & 0.144 \textcolor{gray}{$\pm$ 0.012} & 0.643 \textcolor{gray}{$\pm$ 0.013} &  0.498 \textcolor{gray}{$\pm$ 0.019} \\
              & wilds-camelyon17 & 0.026 \textcolor{gray}{$\pm$ 0.010} & 0.137 \textcolor{gray}{$\pm$ 0.020} &  0.111 \textcolor{gray}{$\pm$ 0.028} \\
              &       wilds-fmow & 0.481 \textcolor{gray}{$\pm$ 0.053} & 0.548 \textcolor{gray}{$\pm$ 0.045} &  0.067 \textcolor{gray}{$\pm$ 0.008} \\
              &   wilds-iwildcam & 0.290 \textcolor{gray}{$\pm$ 0.004} & 0.282 \textcolor{gray}{$\pm$ 0.004} & -0.007 \textcolor{gray}{$\pm$ 0.004} \\
              &      wilds-rxrx1 & 0.875 \textcolor{gray}{$\pm$ 0.042} & 0.905 \textcolor{gray}{$\pm$ 0.027} &  0.029 \textcolor{gray}{$\pm$ 0.016} \\
\midrule
        ViT-B &             PACS & 0.063 \textcolor{gray}{$\pm$ 0.008} & 0.415 \textcolor{gray}{$\pm$ 0.027} &  0.352 \textcolor{gray}{$\pm$ 0.027} \\
              &             VLCS & 0.137 \textcolor{gray}{$\pm$ 0.011} & 0.341 \textcolor{gray}{$\pm$ 0.017} &  0.205 \textcolor{gray}{$\pm$ 0.020} \\
              &      domain\_net & 0.378 \textcolor{gray}{$\pm$ 0.013} & 0.736 \textcolor{gray}{$\pm$ 0.014} &  0.357 \textcolor{gray}{$\pm$ 0.017} \\
              &     office\_home & 0.163 \textcolor{gray}{$\pm$ 0.011} & 0.367 \textcolor{gray}{$\pm$ 0.019} &  0.204 \textcolor{gray}{$\pm$ 0.023} \\
              &            sviro & 0.023 \textcolor{gray}{$\pm$ 0.001} & 0.223 \textcolor{gray}{$\pm$ 0.006} &  0.200 \textcolor{gray}{$\pm$ 0.006} \\
              & terra\_incognita & 0.155 \textcolor{gray}{$\pm$ 0.012} & 0.697 \textcolor{gray}{$\pm$ 0.011} &  0.542 \textcolor{gray}{$\pm$ 0.018} \\
              & wilds-camelyon17 & 0.024 \textcolor{gray}{$\pm$ 0.007} & 0.148 \textcolor{gray}{$\pm$ 0.022} &  0.124 \textcolor{gray}{$\pm$ 0.029} \\
              &       wilds-fmow & 0.495 \textcolor{gray}{$\pm$ 0.038} & 0.565 \textcolor{gray}{$\pm$ 0.033} &  0.070 \textcolor{gray}{$\pm$ 0.006} \\
              &   wilds-iwildcam & 0.303 \textcolor{gray}{$\pm$ 0.006} & 0.334 \textcolor{gray}{$\pm$ 0.021} &  0.031 \textcolor{gray}{$\pm$ 0.017} \\
              &      wilds-rxrx1 & 0.951 \textcolor{gray}{$\pm$ 0.013} & 0.966 \textcolor{gray}{$\pm$ 0.008} &  0.015 \textcolor{gray}{$\pm$ 0.005} \\
\end{longtable}

}

\FloatBarrier

\section{Details on the metrics}
\label{app:metrics}
We choose a representative set of six metrics used in the robustness literature and describe the details below.

Some quantities are standard metrics, such as \emph{classification error} (top-1 classification error) and \emph{negative log-likelihood (NLL)}.
To account for outliers we capped the extreme values of the NLL in the factor analysis as described in \ref{app:factor-analysis}.

We also evaluate the \emph{expected calibration error} (ECE) \cite{guo2017calibration}. The ECE is zero for perfectly calibrated models, i.e.,
if the predicted probabilities by the model match their true probabilities. We calculate the ECE using $10$ bins.

Additionally, two variants of \emph{adversarial classification error} are evaluated by perturbing each test set image using the APGD (Automated Projected Gradient Descent) adversarial attack~\cite{croce2020reliable} with an $\ell_2$-attack of size $0.001$ and of size $0.02$.
If not noted otherwise, we only report the mean classification error resulting from both attack sizes.

Lastly we report \emph{Demographic disparity}, which can be interpreted as a measure of invariance. We first split the data into two environments using the method of~\cite{creager2021environment}, which maximizes the invariant risk minimization penalty~\cite{arjovsky2019invariant}. As the transfer data comes from a single distribution, we would not expect meaningful partitions of the data with systematic differences in the predictions, which we measure with the metric introduced in~\cite{locatello2019fairness}. Note that, while this metric was introduced to evaluate fairness, it should not be interpreted as such in this paper. The discovered groups may not have any semantic meaning nor fairness implication, so it should not be used to justify that a particular model is fairer than another.

\FloatBarrier

\section{Details on the degrees of freedom and hyperparameter selection}
\label{app:model_selection}
In our study, we evaluate each model listed in~\cref{tab:model_names} for all combinations of the hyperparameters listed in~\cref{tab:grid_search}.
In order to reduce the overall number of models to train, we first derived good candidates for the learning rate and number of epochs hyperparameters by a larger sweep on a subset of the datasets. For all models we run a large sweep on the the datasets \emph{VLCS-Caltech101}, \emph{OfficeHome-RealWorld} and \emph{DomainNet-Infograph} on the the extended grid of hyperparameters:
learning rate ($5\mathrm{e}{-5}$, $5\mathrm{e}{-4}$, $5\mathrm{e}{-3}$, $5\mathrm{e}{-2}$) and the number of training epochs ($3$, $10$, $100$, $1000$).
From this we derived the $2 \times 2$ grid of the parameters listed in \cref{tab:grid_search}. We chose the reduced grid of hyperparameter that lead to the best performance for all models (we made sure that for each model, the best performance is attained by at least one of the hyperparameter combinations in the selected grid).
A similar hyperparameter pre-selection strategy was used in \cite{vtab}.

\begin{table}[htb]
    \centering
    \caption{In our study, each model is trained for all combinations of hyperparameters listed in this table.}
    \label{tab:grid_search}
    \begin{tabular}{cccccc}

    \toprule
    \textbf{Training set size} & \textbf{Learning Rate} & \textbf{Train Epochs} & \textbf{Fine-tune} & \textbf{Augmentations} \\ \midrule
    \makecell{full \\ few-shot-10 \\ few-shot-100} & \makecell{$5\mathrm{e}{-4}$ \\ $5\mathrm{e}{-5}$} & \makecell{$10$ \\ $100$} & \makecell{Only head \\ Whole model} & \makecell{No augmentation \\ RandAugment \\ AugMix}\\
    \bottomrule
    \end{tabular}

\end{table}

\begin{table}[htb]
 \caption{The list of models used in our study. The pre-trained weights were taken from the PyTorch Image Models package~\cite{rw2019timm} using the displayed model names.}
    \label{tab:model_names}
    \centering
    \begin{tabular}{l l}
    \toprule
    \textbf{Model} & \textbf{Timm model name}\\
    \midrule
    Deit & \texttt{deit\_base\_distilled\_patch16\_224} \\
    DenseNet & \texttt{densenet169} \\
    EfficientNetV2 & \texttt{efficientnetv2\_rw\_s} \\
    gMLP & \texttt{gmlp\_s16\_224} \\
    MLP-Mixer & \texttt{mixer\_b16\_224} \\
    ResMLP & \texttt{resmlp\_24\_224} \\
    ResNet50d & \texttt{resnet50d} \\
    Swin Transformer & \texttt{swin\_small\_patch4\_window7\_224} \\
    Vision Transformer & \texttt{vit\_base\_patch16\_224} \\
    \bottomrule
    \end{tabular}
\end{table}

\FloatBarrier

\section{Societal impact, limitations and hardware overview}
\subsection{Limitations}
\label{sec:app:limitations}
Despite best efforts, a large scale experiment like this can never be fully extensive in terms of hyperparameter selection, the choice of model architectures, the evaluated metrics and the overall statistics.
We address the limitations by cautious interpretation of the experimental results and the conclusions, see \cref{sec:a-broad-look-at-out_of_distribution-generalization,sec:effect-of-augmentations-fine_tuning-stragety-and-few_shot-learning} and particularly the take-away messages therein.
\subsection{Societal impact}
\label{sec:app:societal-impact}
This work analyzes how in-distribution metrics relate to out-of-distribution performance.
Such questions are often highly relevant when deploying machine learning algorithms to real world systems, since those algorithms get typically trained in specific, possibly idealized environments, which differ from the application environment.
Understanding the in- to out-of-distribution generalization properties can therefore easily become relevant for the customer's satisfaction and security.
The challenge for improved in- and out-of-distribution generalization is also closely linked to algorithmic fairness \cite{zietlow2022leveling}, which has become a highly relevant societal topic.
\subsection{Computer overview}
All $31\mathrm{k}$ experiments were conducted on a cloud hosted cluster using Nvidia T4 GPUs. The aggregated compute time is 17 (GPU-)years.
\label{app:compute_overview}
\end{document}